\documentclass[12pt]{article}

\usepackage[english]{babel}
\usepackage[margin=1in]{geometry}
\usepackage{graphicx}
\usepackage[dvipsnames]{xcolor}
\usepackage{tikz}
\usetikzlibrary{positioning, arrows.meta}
\tikzstyle{arrow} = [thick,->,>=Stealth]
\tikzstyle{step} = [rectangle, draw=black, fill=blue!10, text width=5cm, text centered, rounded corners, minimum height=1cm]
\usepackage{pgfplots}
\pgfplotsset{compat=1.17}
\usepackage{colortbl}

\usepackage{caption}
\usepackage{subcaption}
\usepackage{longtable}
\usepackage{booktabs}
\usepackage{multirow}
\usepackage{threeparttable}
\usepackage{array}
\newcolumntype{x}[1]{>{\centering\hspace{0pt}}p{#1}}

\newcommand{\barCell}[3]{%
  \begin{tikzpicture}[baseline]
    \draw[fill=#3!70] (0,0) rectangle (#1/25,0.3);
    \node[anchor=west] at (#1/25 , 0.15) {\scriptsize #2};
  \end{tikzpicture}%
}

\usepackage{amsmath, amsthm, amssymb, bm, bbm}

\usepackage[numbers]{natbib}
\bibpunct[, ]{(}{)}{;}{a}{}{,}%

\usepackage{hyperref}
\definecolor{darkblue}{RGB}{0,0,168}
\definecolor{lightblue}{RGB}{173,216,230}
\hypersetup{
    colorlinks=true,
    linkcolor=darkblue,
    filecolor=darkblue,
    citecolor=darkblue,
    urlcolor=darkblue
}
\usepackage{rotating}
\usepackage{setspace}
\usepackage{indentfirst}
\usepackage{titlesec}
\usepackage{titletoc}
\titleformat{\section}{\normalfont\fontsize{13}{13}\bfseries\itshape}{}{0em}{\centering\MakeUppercase}
\vspace{0.5ex}
\titleformat{\subsection}{\normalfont\fontsize{13}{13}\bfseries\itshape}{}{-0.5em}{}
\titleformat{\subsubsection}
  {\normalfont\fontsize{12}{12}\bfseries} 
  {}                                              
  {-0.5em}                                        
  {}                                              
\titlespacing*{\subsubsection}
  {0pt}       
  {0.5ex plus 0.2ex minus 0.1ex}   
  {0.3ex}     

\titlespacing*{\subsection}
  {0pt}       
  {0.5ex plus 0.2ex minus 0.1ex}   
  {0.3ex}     

\titlespacing*{\section}
  {0pt}       
  {0.5ex plus 0.2ex minus 0.1ex}   
  {0.3ex}     
  
\usepackage{float} 
\newfloat{prompt}{htbp}{lop}
\floatname{prompt}{Prompt}

\usepackage{xcolor}
\definecolor{blue}{RGB}{25,90,160}
\definecolor{green}{RGB}{0,100,0}

\usepackage{algorithm}
\usepackage{algpseudocode}
\usepackage{tcolorbox}
\tcbuselibrary{listingsutf8}

\usepackage{bibunits}
\defaultbibliography{references}
\defaultbibliographystyle{jmr.bst}

\title{\large{\textbf{From Reviews to Actionable Insights:\\ An LLM-Based Approach for Attribute and Feature Extraction}}}

\author{
{\normalsize 
\textbf{Khaled Boughanmi, Kamel Jedidi, and Nour Jedidi}}\thanks{Khaled Boughanmi is an Assistant Professor of Marketing at the Samuel Curtis Johnson Graduate School of Management, Cornell University, Ithaca, NY (email: \texttt{kb746@cornell.edu}). He is the corresponding author. Kamel Jedidi is the Jerome A. Chazen Professor of Global Business at Columbia Business School, New York, NY (email: \texttt{kj7@gsb.columbia.edu}). Nour Jedidi is a PhD candidate in Computer Science at the University of Waterloo, Waterloo, Canada (email: \texttt{njedidi@uwaterloo.ca}). 
The authors gratefully acknowledge Malek Ben Slimane, Omar Besbes, and Robert J. Morais for their valuable feedback. They also thank Lucie Clifford, Yanjie Chen, and Yuxuan Wang from the Columbia Business School Behavioral Research Lab for their research assistance. The authors are grateful to the seminar participants at the universities of Jean Monnet Saint-Étienne, Lumière Lyon~2,  Paris Business School, Pompeu Fabra, and the COACTIS Research Center for their helpful comments and suggestions. This research was supported by a grant from the Cornell Center for Social Sciences.
}\\
}

\date{}

\begin{document}

    \thispagestyle{empty}
    \maketitle

    \singlespacing

\thispagestyle{empty}

\begin{abstract}
\noindent This research proposes a systematic, large language model (LLM) approach for extracting product and service attributes, features, and associated sentiments from customer reviews. Grounded in marketing theory, the framework distinguishes perceptual attributes from actionable features, producing interpretable and managerially actionable insights. We apply the methodology to 20,000 Yelp reviews of Starbucks stores and evaluate eight prompt variants on a random subset of reviews. Model performance is assessed through agreement with human annotations and predictive validity for customer ratings. Results show high consistency between LLMs and human coders and strong predictive validity, confirming the reliability of the approach. Human coders required a median of six minutes per review, whereas the LLM processed each in two seconds, delivering comparable insights at a scale unattainable through manual coding. Managerially, the analysis identifies attributes and features that most strongly influence customer satisfaction and their associated sentiments, enabling firms to pinpoint “joy points,” address “pain points,” and design targeted interventions. We demonstrate how structured review data can power an actionable marketing dashboard that tracks sentiment over time and across stores, benchmarks performance, and highlights high-leverage features for improvement. Simulations indicate that enhancing sentiment for key service features could yield 1–2\% average revenue gains per store.


\end{abstract}
\clearpage

\pagenumbering{arabic} 
\setcounter{page}{1}   

    \singlespacing

    Customer reviews heavily influence customer purchasing decisions \citep{elwalda2016impact}. According to \cite{forbes2024reviews}, 98\% of consumers read reviews before making a purchase, highlighting the rich information about products and services contained in these evaluations. Beyond guiding consumer choice, customer feedback serves as an important source of market intelligence for marketers on customer sentiments, perceptions, and preferences.

Features and attributes constitute the core elements that customers emphasize in their evaluations of products and services. 
Attributes indicate how customers feel about higher-level dimensions such as customer service, but they do not reveal which specific aspects/features such as waiting time are driving those evaluations. In this research, we define features as specific, tangible, and actionable characteristics of a product or service, while attributes are the benefits these features provide to customers. As in means–end chain theory \citep{gutman1982means}, features are characteristics and attributes are consequences. Extracting such information from customer reviews, along with associated sentiment, provides firms with a blueprint for understanding how consumers evaluate their experiences. 

Customer sentiment toward attributes and features enables businesses to detect ``pain points'' or areas needing improvement, and ``joy points'' or positive experiences that enhance customer loyalty. It is also a powerful predictor of overall satisfaction, which in turn shapes firm reputation and drives word-of-mouth, both of which are closely linked to long-term profitability and financial performance \citep{chevalier2006effect}. Such an analysis yields actionable insights by indicating which features should be prioritized for improvement to positively influence customer satisfaction.

Extracting such insights has long been challenging using traditional methods due to the complexity of textual data \citep{berger2019uniting}. Techniques such as topic modeling (e.g., LDA) and conventional NLP approaches struggle to reliably identify meaningful topics and associated sentiments in customer reviews due to linguistic and stylistic variation (e.g., “The staff was friendly and attentive” and “Employees treated me nicely and kept checking in” are semantically equivalent, but traditional methods may treat them as distinct topics) \citep{buschken2020improving}, semantic ambiguity (e.g., “The meal was hot” could mean spicy or overheated) \citep{jusoh2018study}, the multiplicity of themes and nuanced sentiments expressed within reviews (e.g., “The food was delicious, but the service was slow”) \citep{chakraborty2022attribute}, and the broader context in which words and phrases are used (e.g., “The place is small” could imply cramped and uncomfortable, or cozy and intimate, depending on the context) \citep{bojic2025evaluating}. 

Recent advances in large language models (LLMs) provide a powerful alternative by enabling the extraction of meaningful topics along with more accurate and context-aware interpretation of unstructured text. Notably, LLMs are approaching human-level understanding and interpretation of language \citep{halawi2024approaching}, creating new opportunities for scalable and cost-effective analysis of vast amounts of unstructured data. Since their inception and release, LLMs have received significant attention from the marketing research community \citep{blanchard2025express}. For example, \cite{brand2023using} use LLMs to conduct marketing research by generating multiple responses to survey questions, showing that they yield realistic willingness-to-pay estimates comparable to human data. \cite{goli2024frontiers} examine whether LLMs can capture human preferences, with a focus on intertemporal decision-making, and find substantive differences between human subjects and LLMs. \cite{li2024frontiers} investigate the potential for LLMs to substitute human participants in marketing research.
\cite{chakraborty2025can} develop an LLM-based model for salesforce hiring using text and audio data, predicting sales talent by analyzing candidate conversations.

This research proposes a systematic, LLM-based approach to extract product and service features, attributes, and associated sentiments from customer reviews. Distinguishing between features and attributes is critical, as it provides a theoretical structure for guiding LLMs to generate insights that are both theoretically meaningful and managerially actionable. Without this distinction, LLMs risk conflating features and attributes, producing topics that are either too abstract to guide decisions or too granular to offer strategic value. Unlike traditional methods, such as LDA, which often yield vague or thematically diffuse topics due to bag-of-words assumptions and limited semantic understanding \citep{Pham2023}, guided LLMs can generate coherent, context-aware insights aligned with marketing theory and managerial needs \citep{mu2024llm,arora2025ai}. Our approach operationalizes this guidance, ensuring outputs that not only produce meaningful attributes and features but also inform managerial decision-making.

Our approach proceeds in three steps. The first is an exploratory phase, in which we prompt and guide LLMs to generate a comprehensive set of attributes and their associated features from the corpus of reviews, ensuring that no important elements are overlooked. After removing semantically equivalent items, this step produces a concise list of distinct attributes and features. This list serves as the reference framework for systematically extracting and organizing information from all customer reviews in the corpus, ensuring consistency and reliability. Without it, the LLM risks generating ad hoc attributes and features that undermine the coherence and actionability of the analysis.

The second step is confirmatory where we present the LLM with one review at a time and prompt (and guide) it to (i) identify the attributes and features from the list that are explicitly mentioned in the review, and (ii) score them on sentiment. This process yields a structured dataset that specifies, for each review, the attributes and features mentioned along with their associated sentiment. 

To account for context and minimize hallucinations and spurious results, we first present the full review to the LLM for an overall sentiment evaluation, enabling it to capture the broader context. The review is then split into sentences \citep{buschken2020improving} which the LLM processes sequentially, assigning each to one or more attributes while considering the surrounding sentences for contextual accuracy (e.g., the sentence `What can a girl do.' is only meaningful when paired with the previous one `The coffee is great but expensive.'). For each attribute, the subset of associated sentences is then used as a “sub-review,” within which the LLM evaluates sentiment toward the attribute, assigns sentences to one or more of its features using the same process used for attributes, and evaluates sentiment toward each feature. To minimize order bias, attributes and features are randomly presented. This multi-step design enhances both the reliability and interpretability of the analysis compared to single-pass classification.

In the third step, we analyze the dataset to derive actionable managerial insights.

We demonstrate our approach using 20,000 Yelp reviews of Starbucks coffee shops \citep{yelp_open_dataset}, spanning 722 locations across 13 U.S. states and the Canadian province of Alberta from 2005 to 2022. As one of the most frequently reviewed brands on digital platforms, Starbucks constitutes an ideal research context, given the breadth of customer feedback encompassing product quality, service interactions, and store environment.

The first phase of our analysis yields a concise list of ten attributes, each linked to 3 to 6 features that span diverse domains of the customer experience, from coffee quality to customer service and store ambiance. The extracted features are actionable. For customer service, they include, among others, service efficiency, order accuracy, and staff professionalism. The ten attributes are highly interpretable, account for about 76\% of the review content, exhibit low correlation with one another, and are consistent with prior literature. The remaining 24\% reflects non-diagnostic aspects (e.g., comments such as ‘Starbucks is driving out small coffee shops’) or overall statements toward the Starbucks brand or local store. 

In the second phase, we test eight LLM prompt variants on a random subset of 300 reviews, assessing performance using agreement metrics with human annotations and predictive validity for customer ratings. The results indicate strong predictive validity and high levels of agreement between human coders and LLM outputs across all the variants, with our proposed approach achieving the highest agreement. Importantly, while human coders required a median of six minutes per review (with 90\% of reviews containing 2–10 sentences) and could not process more than five reviews in a session, the LLM completed each review in less than two seconds. This efficiency gain underscores the scalability of the approach, enabling the analysis of thousands of reviews that would be infeasible to code manually while maintaining accuracy comparable to human judgment.

Our descriptive analysis of the structured data shows that Customer Service, Coffee \& Beverage, and store-related attributes (Ambiance \& Atmosphere, Store Comfort \& Layout, and Facilities \& Accessibility) are the most frequently mentioned attributes of the Starbucks experience. Among these, Customer Service dominates how customers evaluate the brand. The most frequently mentioned features vary by attribute.
 
For Customer Service reviewers often emphasize Staff Professionalism and Service Efficiency/Wait time; for Coffee \& Beverage, Coffee Taste is the most salient feature; and for store-related attributes, Location Convenience and Seating Availability/Comfort dominate. 

Customer sentiment is generally polarized across most attributes and features, with sentiment toward customer service and its associated features nearly evenly split between positive and negative evaluations. These sentiments, both at the attribute- and feature-level are highly predictive of customer ratings ($R^2$ = .73 and .71, respectively), outperforming state-of-the-art NLP models. 

Finally, we demonstrate how marketers can leverage the structured review data to develop a marketing dashboard to monitor customer feedback and guide decisions. The dashboard displays attribute mentions and sentiments for each store and tracks their evolution over time. Across stores, the results reveal substantial variability in attribute and feature performance, highlighting store-level ``joy points" and ``pain points." This variability provides Starbucks with opportunities to implement targeted, localized actions to enhance customer satisfaction. Over time, we observe a steady decline in the share of positive sentiment and a corresponding rise in negative sentiment across most attributes. This is particularly true for Customer Service where the two curves intersect in 2016, a turning point in Starbucks’ employee relations marked by rising performance pressure, declining workplace reputation, growing employee dissatisfaction, and early unionization efforts.\footnote{\url{https://hbr.org/2024/06/how-starbucks-devalued-its-own-brand}}

The dashboard also simulates the impact of improving feature-level sentiment on store satisfaction. For example, a one point improvement in sentiment for Staff Professionalism (e.g., through training) is associated with an increases of .19 in average store rating. Based on prior evidence that a one-point increase in ratings can raise revenue by up to 9\% \citep{luca2016reviews}, this improvement could yield a 1 to 2\% gain in average store revenues. While not causal, such an analysis highlights actionable opportunities for targeted interventions and provides guidance for Starbucks in designing field experiments to assess expected ROI.

Textual data has attracted substantial attention in the marketing field due to the richness of information contained in such unstructured data. \cite{berger2019uniting} provide an extensive review of marketing research that leverages textual data to extract business insights, offering an excellent summary of studies that quantify consumer-generated text \citep[e.g.,][]{archak2011deriving,lee2011automated,anderson2014reviews, tirunillai2014mining, büschken2016sentence,liu2019large, jedidi2021r2m, boughanmiexpress}. We contribute to this literature in three ways. First, we introduce an attribute–feature framework to guide LLMs in analyzing customer reviews. Grounded in marketing theory, the framework separates the perceptual (attributes) from the actionable (features) and provides a structured approach for extracting information that is interpretable and managerially relevant. 

Second, we propose a multi-phase approach that provides marketers with an automatable tool that generates a robust list of attributes and features and uses it as a template for structuring unstructured review text. Our modular prompting design breaks the task into simpler subtasks  (e.g., attribute/feature identification, sentiment classification), leverages few-shot examples to guide the LLM through a structured reasoning process, and keeps intermediate outputs in context, enabling the LLM to use prior responses as additional signal for the current step. This design improves accuracy, minimizes hallucinations, and ensures coherent results~\citep{khotdecomposed}. Validation against human coders shows that the approach delivers human-level reliability and outperforms state-of-the-art NLP models in predicting customer ratings.

Finally, we develop a data analytics approach whose outputs can function as a marketing dashboard. Our approach provides fine-grained diagnostics by identifying attribute and feature-level ``pain points" and ``joy points," tracking how customer satisfaction evolves over time and varies across stores. Powered by automated LLM analysis, the dashboard can be applied in real time and at scale, enabling management to monitor performance, benchmark stores, and detect emerging issues or strengths. Importantly, it translates unstructured reviews into actionable insights that guide targeted interventions and A/B experiments to improve satisfaction and foster loyalty, both system-wide and at the store level.

The paper is organized as follows. We begin by describing the data and then present our LLM-based approach for extracting attributes and features from reviews. We next validate the approach against human annotations, followed by our empirical results and an illustration of how the structured data can be leveraged to build a marketing dashboard. We conclude by summarizing our contributions, discussing limitations, and suggesting directions for future research.

    \section*{Data}
\label{sec:data}

\begin{figure}[htbp]
    \centering
    \begin{subfigure}[t]{0.45\textwidth}
        \centering
        \includegraphics[width=\linewidth]{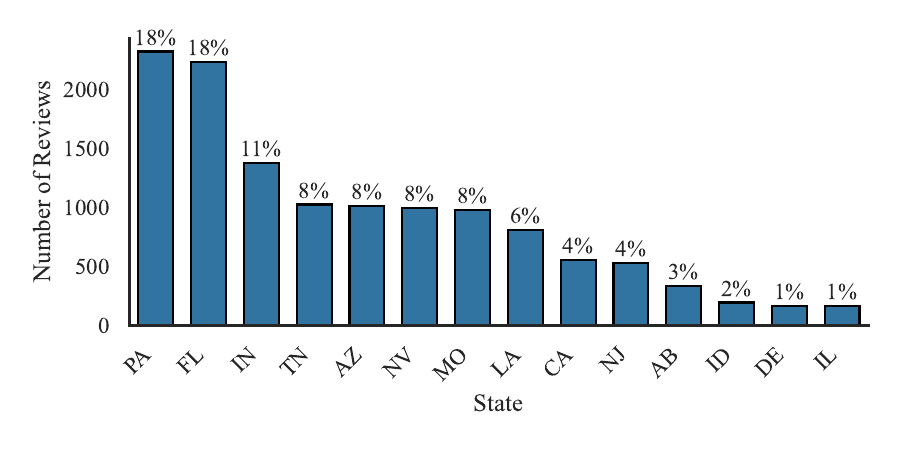}
        \caption{Distribution of Reviews by State}
        \label{fig:states_counts}
    \end{subfigure}
    \hfill
    \begin{subfigure}[t]{0.45\textwidth}
        \centering
        \includegraphics[width=\linewidth]{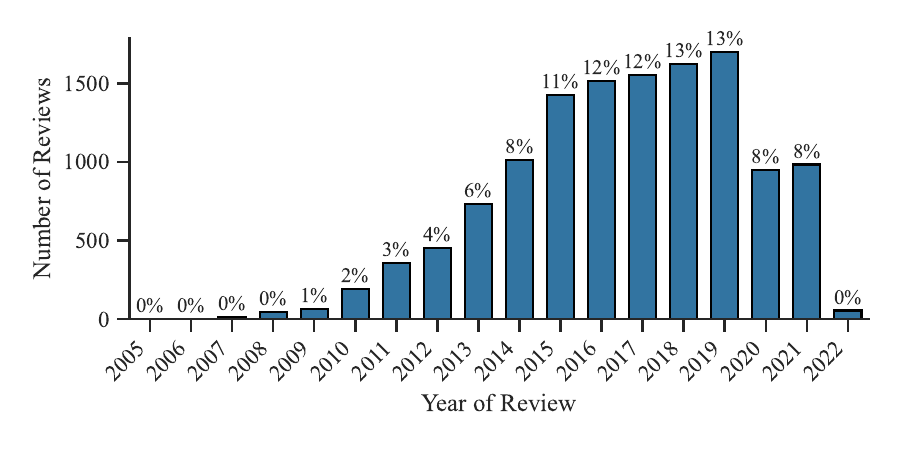}
        \caption{Distribution of Reviews over Time}
        \label{fig:review_years}
    \end{subfigure}
    \caption{Distribution of Reviews by State and over Time}
    \label{fig:reviews_combined}
\end{figure}

\begin{figure}[H]
    \centering
    \begin{subfigure}[t]{0.43\textwidth}
        \centering
        \includegraphics[width=\linewidth]{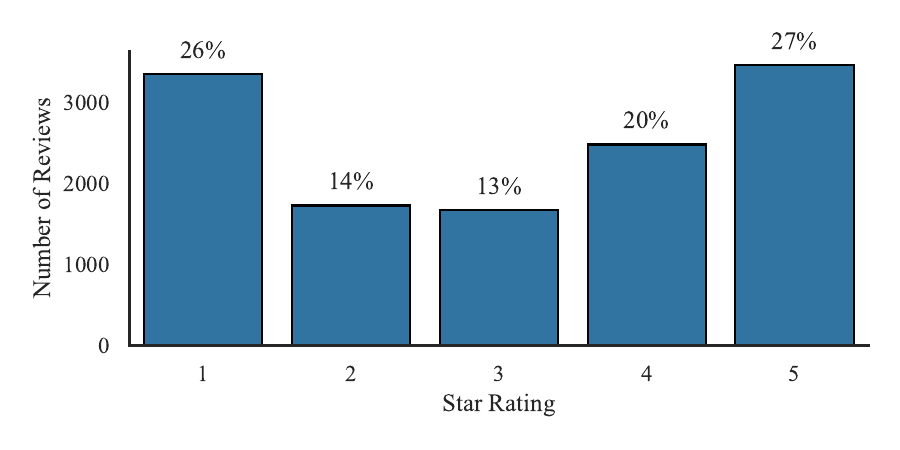}
        \caption{Review-level Ratings}
        \label{fig:review_level_ratings}
    \end{subfigure}
    \hfill
    \begin{subfigure}[t]{0.43\textwidth}
        \centering
        \includegraphics[width=\linewidth]{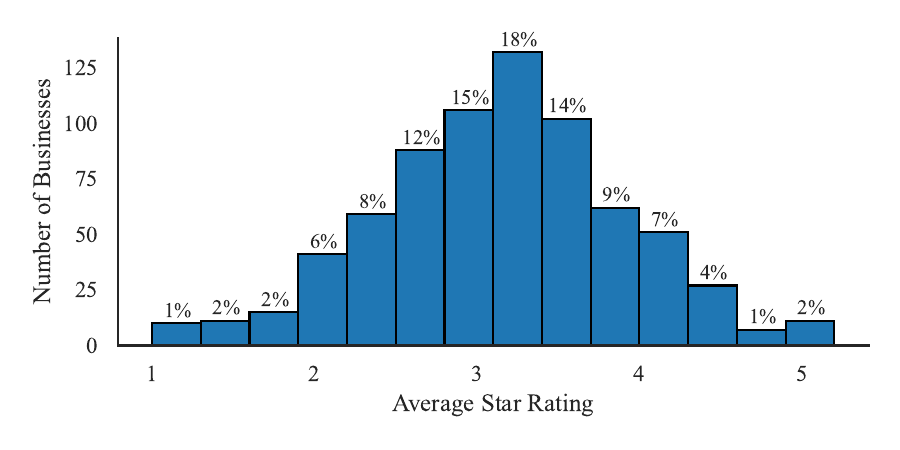}
        \caption{Store-level Ratings}
        \label{fig:store_level_ratings}
    \end{subfigure}
    \caption{Distribution of Customer Ratings}
    \label{fig:ratings}
\end{figure}

\begin{figure}[H]
    \centering
    \includegraphics[width=0.33\linewidth]{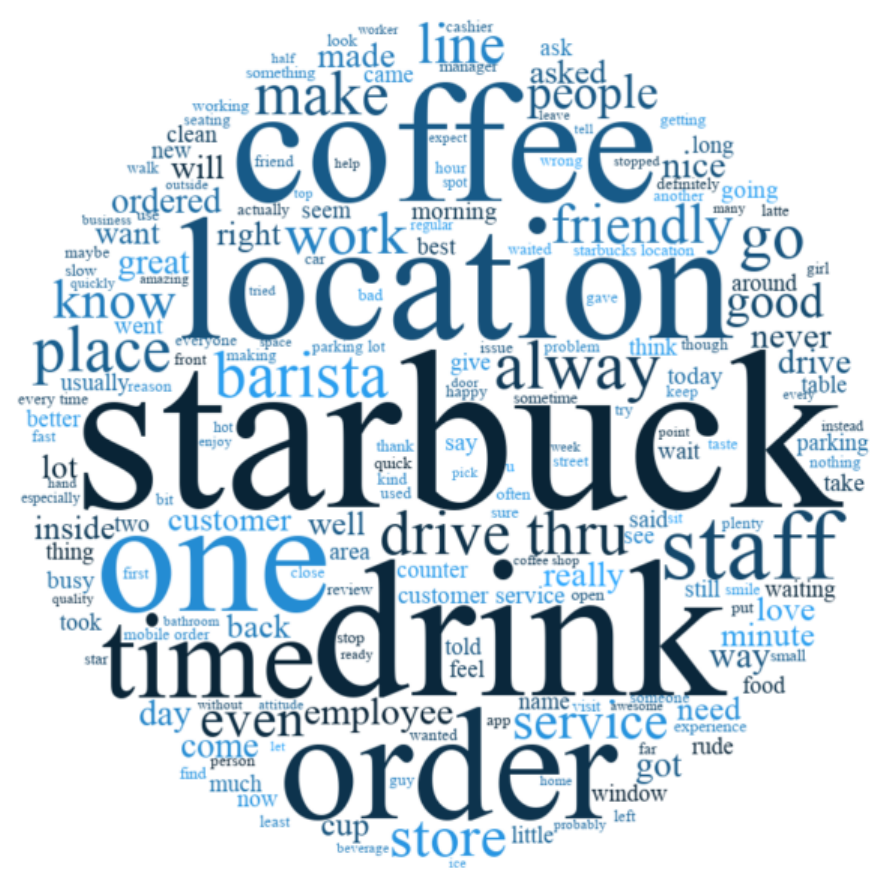}
    \caption{Word Cloud of Reviews}
    \label{fig:wordcloud}
\end{figure}

We apply our proposed approach to a publicly available Yelp dataset of customer reviews of Starbucks coffee shops, primarily in the U.S. \citep{yelp_open_dataset}. Our analysis is based on 12,682 randomly selected reviews from a corpus of about 20,000.\footnote{Due to budget limitations from LLM usage costs, we analyzed only 12,682 reviews. Given the large sample size and random selection, our findings should be robust.} The dataset includes 10,264 unique reviewers, who submitted an average of 1.24 reviews (median = 1). It spans 722 unique Starbucks locations across thirteen U.S. states and one Canadian province. On average, each store has about 18 reviews (median = 13), with the most reviewed location receiving over 100 reviews. Figure \ref{fig:states_counts} presents the distribution of reviews by state. Pennsylvania accounts for the highest number of reviews, followed by Florida and Indiana. Figure \ref{fig:review_years} shows the distribution of reviews over time. The reviews cover a 17-year period from 2005 to 2022, with 77\% of them recorded after 2015.
Our dataset includes customers’ overall ratings of their experience on a 5-point scale. As shown in Figure~\ref{fig:review_level_ratings}, the distribution of customer ratings is J-shaped with a disproportionate number of extreme scores, likely reflecting self-selection by customers with strong opinions, a pattern commonly observed in the review literature \citep{schoenmueller2020polarity}. Approximately 47\% of the reviews are positive, corresponding to ratings of 4 or 5 stars, while the remaining 53\% received weaker ratings of 1 to 3 stars. The average rating is 3.08 with a standard deviation of 1.27. Figure \ref{fig:store_level_ratings} presents the distribution of customer ratings across stores. The distribution is bell-shaped, consistent with the central limit theorem, which implies normality in the distribution of averages.
On average, a Starbucks store received a rating of 3.14, with a standard deviation of .76.
Our dataset also includes the review text, in which customers describe their experiences with the store. The reviews vary in length, style, and content. On average, a review contains five sentences, with a 90\% confidence interval of [2, 10]. Customers discuss a wide range of themes. The word cloud in Figure~\ref{fig:wordcloud} highlights the most prominent terms, including ``Starbucks," ``coffee," ``order," ``drinks," ``location," and ``staff." 

    \section*{Proposed LLM Approach}
Our approach relies on the marketing concepts of attributes and features to structure review information and guide LLMs in extracting insights that are both meaningful and managerially actionable. Consistent with means–end chain theory \citep{gutman1982means},\footnote{This distinction between attributes and features also parallels the Features–Advantages–Benefits (FAB) framework commonly used in marketing communications and sales \citep{kottler2009marketing}. For parsimony, we combine advantages and benefits under attributes.} attributes capture the benefits customers seek, while features represent the tangible characteristics that deliver those benefits.

\begin{figure}[htbp]
    \centering
    \includegraphics[width=0.8\linewidth]{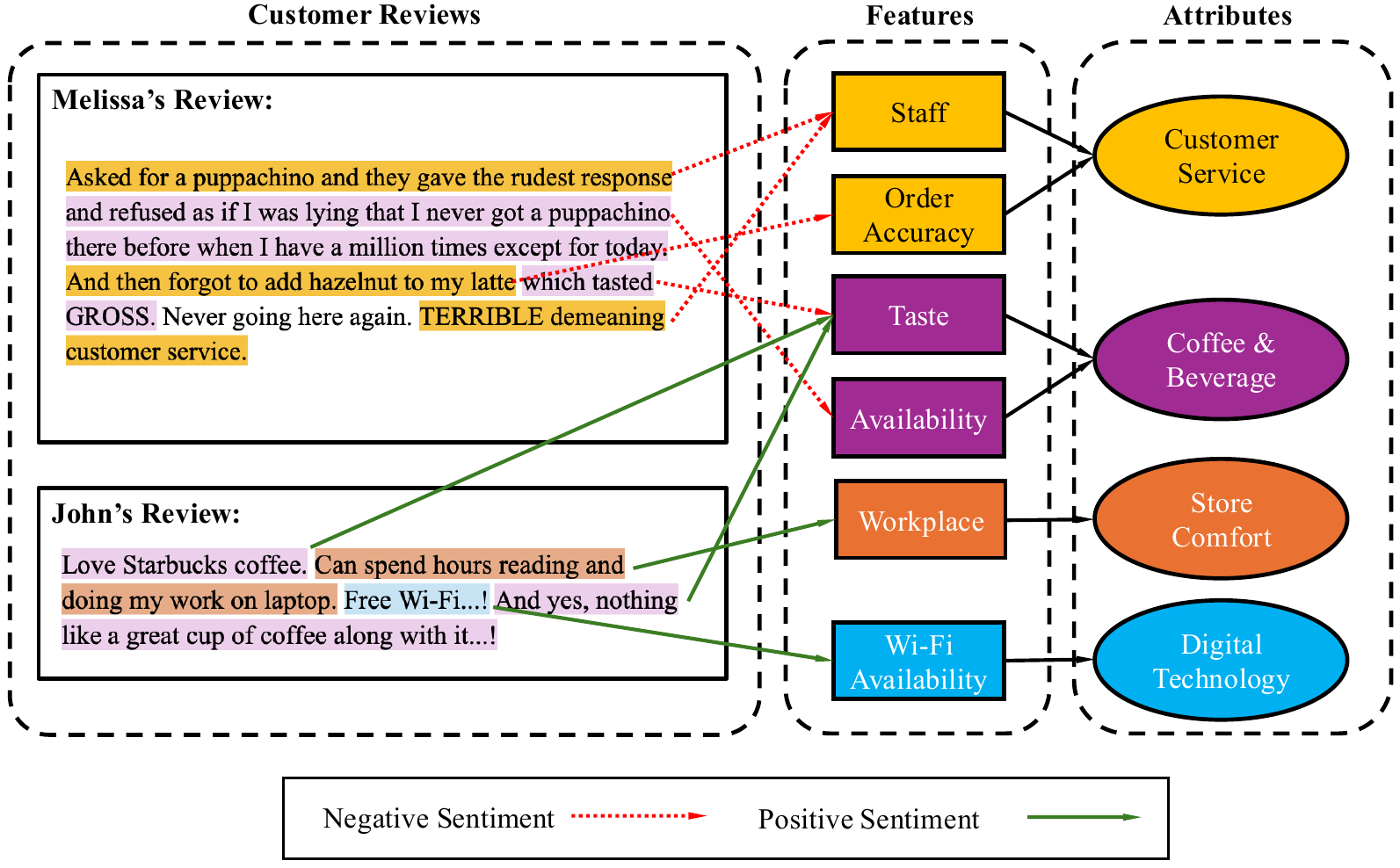}
    \caption{Illustration of Attribute–Feature–Sentiment Extraction from Customer Reviews}
    \label{fig:examples}
\end{figure}

Attributes and features, along with their associated sentiments, are the core elements customers emphasize when evaluating products and services in reviews. For example, consider Melissa's negative review shown in the top left of Figure 
\ref{fig:examples}. The opening phrase, `Asked for a Puppachino and they gave the rudest response' (highlighted in golden yellow), conveys dissatisfaction with staff, a feature of the Customer Service attribute. The continuation, `and refused as if I was lying …' (purple), points to the (un)availability of the Puppachino, tied to Coffee \& Beverage. The next sentence, `And then forgot to add hazelnut to my latte' (golden yellow), highlights order accuracy, another feature of Customer Service, while `which tasted GROSS' (purple) criticizes taste, a feature of Coffee \& Beverage. The statement `Never going here again' reflects an overall judgment and is classified under Other Attributes (not shown in the figure) in our framework, as it is neither diagnostic nor actionable. Finally, `TERRIBLE demeaning customer service' (golden yellow) refers to staff and the broader Customer Service attribute. Melissa’s review thus maps to two attributes: Customer Service and Coffee \& Beverage; each linked to actionable pairs of features (staff, order accuracy) and (taste, availability), respectively, with negative sentiment highlighted by dashed-red arrows. 

Using the same reasoning, John’s positive review (shown in the bottom left of Figure~\ref{fig:examples}) maps to three attributes: Coffee \& Beverage, Store Comfort, and Digital Technology. These are tied to actionable features: taste, workspace, and Wi-Fi availability, respectively, with positive sentiment highlighted by green arrows.

 These examples illustrate how our framework systematically maps review sentences to features and attributes, with sentiment indicating whether each is a pain point or a source of delight. In doing so, the framework separates attributes (perceptual dimensions) from features (actionable levers), yielding insights that are both interpretable and managerially relevant. This distinction provides the theoretical structure needed to guide LLMs and prevent conflating broad perceptions with actionable details. Below, we describe how we use LLMs to generate a comprehensive list of features and attributes and to extract this information from reviews along with their associated sentiments.
 
\begin{figure}[htbp]
\centering
\begin{tikzpicture}[node distance=1.0cm and 2cm, every node/.style={draw, rectangle, rounded corners, align=center, minimum width=4.5cm, minimum height=1.2cm, fill=blue!3, font=\sffamily}]

\node (step1) {\textbf{Step 1}: Exploratory Phase:\\ Generate a comprehensive list of attributes and features};
\node (step2) [below=of step1] {\textbf{Step 2}: Confirmatory Phase:\\ Extract attributes and features and score them on sentiment};
\node (step3) [below=of step2] {\textbf{Step 3}: Data Analysis:\\ Analyze data and generate insights};

\draw[->, thick] (step1) -- (step2);
\draw[->, thick] (step2) -- (step3);

\end{tikzpicture}
\caption{Pipeline of the Proposed Approach}
\label{fig:extraction_pipeline}
\end{figure}
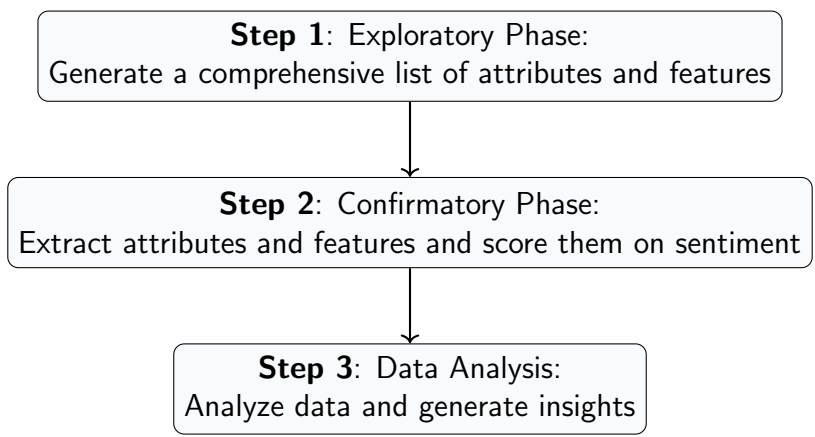

Our proposed approach, summarized in Figure~\ref{fig:extraction_pipeline}, is structured as a three-step pipeline in which the output of each step serves as the input for the next. 
\begin{enumerate}
    \itemsep0em
    \item \textbf{Step 1} is an exploratory phase designed to surface the full range of attributes and features present in the review corpus. Using guided prompts, the LLM generates candidate attributes and features from several random subsets of reviews, which are then consolidated by merging semantically similar items. We also examine the prevalence of attributes in the review corpus and exclude those with frequencies below 1\% of the sample. The outcome is a streamlined set of distinct attributes and features that managers can readily interpret and act upon. This structured and concise list becomes the foundation for the subsequent analysis, providing consistency across reviews and preventing the LLM from drifting into ad hoc or incoherent classifications.
        
    \item \textbf{Step 2} is a confirmatory phase in which each review is analyzed to detect the attributes and features identified in Step 1. Using guided prompts, the LLM first evaluates the full review to capture context and assign overall sentiment, then processes sentences sequentially, assigning them to one or more attributes while considering surrounding sentences for accuracy. For each attribute, the associated sentences are treated as a sub-review, within which the LLM assesses sentiment toward the attribute, assigns sentences to linked features, and evaluates sentiment at the feature level. This process produces a structured dataset that maps each review to the attributes and features it references, together with the sentiment expressed toward them.
    
    \item \textbf{Step 3} analyzes the structured dataset to generate actionable insights. 
    The output of this step can be used to power a marketing dashboard that pinpoints ‘pain points’ and ‘joy points,’ tracks satisfaction over time and across stores, and provides real-time guidance for targeted interventions and A/B experiments.
\end{enumerate}

We next present our prompting algorithms for the exploratory and confirmatory steps.


\subsection*{Generating a Concise List of Attributes and Features}
Algorithm \ref{alg:attribute_feature_extraction} provides the details of how we generate a comprehensive and non-redundant list of attributes and associated features from the reviews. We begin by randomly sampling $N$ batches of $n$ reviews each from the full corpus. Batching is critical because it partitions the corpus (20,000 reviews in our case) into smaller subsets (1,000 reviews here), enabling the LLM to focus more effectively on information extraction. Smaller batches improve reliability by reducing the likelihood of missed information and enhancing detail retention \citep{flemings2024characterizing}. Empirically, we find that batching yields a more comprehensive and richer set of attributes and features than a one-shot extraction from the full corpus.

For each batch, we first provide the LLM with explicit definitions of both attributes and features, along with illustrative examples for guidance, consistent with evidence that examples improve LLM performance via few-shot prompting \citep{brown2020language, min2022rethinking}. We then prompt it to independently extract (i) all features and (ii) all attributes mentioned across the reviews in the batch. Running the two processes separately ensures that the identification of features does not bias the identification of attributes, and vice versa.

\begin{algorithm}[htbp]
\caption{Attribute and Feature Generation from Customer Reviews}
\label{alg:attribute_feature_extraction}
\scriptsize
\begin{algorithmic}[1]
\State \textbf{Input:} A corpus of $M$ customer reviews (e.g., $M=20,000$)
\State \textbf{Goal:} Generate a comprehensive, non-redundant list of attributes and associated features

\Statex
\State \textbf{Task 1: Sampling Phase}
\State Randomly sample $N$ batches of $n$ reviews each from the corpus (e.g., $N=20$, $n=1,000$)

\Statex
\State \textbf{Task 2: Attribute and Feature Identification}
\For{each batch}
    \State Define attributes and features
    \State Prompt the LLM to extract all \textbf{features} mentioned in the reviews \hfill \Comment{Task 2.1}
    \State Independently prompt the LLM to extract all \textbf{attributes} mentioned in the same reviews \hfill \Comment{Task 2.2}
\EndFor
\State \textit{Note: Tasks 2.1 and 2.2 are conducted independently to avoid mutual bias.}

\Statex
\State \textbf{Task 3: Consolidation Phase}
\State Aggregate all extracted attributes and features from the $N$ batches
\State Clean the results by:
\begin{itemize}
    \itemsep0em
    \item[-] Removing duplicates
    \item[-] Merging semantically similar items
    \item[-] Standardizing terminology
\end{itemize}

\State \Return Comprehensive, non-redundant list of attributes and associated features
\end{algorithmic}
\label{alg:step1}
\end{algorithm}

After iterating through all N=20 batches, we proceed to a consolidation step. This step aggregates the extracted attributes and features into a comprehensive master list. It involves manual intervention to remove duplicates and merge semantically similar items, whether the similarity is lexical or conceptual. Next, we standardize the terminology to ensure consistency across the final list of attributes and features. Finally, we examine the prevalence of attributes in the review corpus and exclude those with frequencies below 1\% of the sample. The  details of the procedure are detailed in Prompts \ref{prompt:feature_discovery} and \ref{prompt:att_discovery} in  Web Appendix \ref{sec:att_disc_prompts}. 

\begin{sidewaystable}[htbp]
\caption{List of Attributes and their Associated Features}
\scriptsize
\label{tab:attribtues_features}
\begin{tabular}{lllll}
\toprule
\textbf{Store   Ambiance \& Atmosphere (Sensory Experience \& Mood)} &  & \textbf{Store Comfort \& Layout   (Physical Comfort \& Functionality)}        &  &  \\
\midrule
Interior Design \& Décor                                             &  & Indoor/Outdoor Seating                                                        &  &  \\
Music, Lighting, Noise                                               &  & Seating Availability \& Comfort                                               &  &  \\
Pet-Friendly Coffee Shop                                             &  & Tables Arrangement                                                            &  &  \\
Sense of Community/Inclusivity                                       &  & Temperature Control                                                           &  &  \\
                                                                     &  & Workspace Quality                                                             &  &  \\
                                                                     &  &                                                                               &  &  \\
\midrule
\textbf{Store Cleanliness \&   Hygiene (Sanitation \& Maintenance)}  &  & \textbf{Facilities \& Accessibility (Convenience \& Inclusivity)}             &  &  \\
\midrule
Air Quality \& Odors                                                 &  & Drive-Through Availability \& Quality                                         &  &  \\
Restroom Cleanliness                                                 &  & Parking Accessibility                                                         &  &  \\
Store Cleanliness/Trash Disposal                                     &  & Restroom Access                                                               &  &  \\
                                                                     &  & Store \& Online Operating Hours                                               &  &  \\
                                                                     &  & Store Location Convenience                                                    &  &  \\
                                                                     &  &                                                                               &  &  \\
\midrule
\textbf{Customer Service (Interaction   \& Efficiency)}              &  & \textbf{Coffee \& Beverage (Taste \& Consistency)}                            &  &  \\
\midrule
Complaints \& Conflict Resolution                                    &  & Coffee \& Beverage Customization \& Personalization                           &  &  \\
Customer Service Consistency                                         &  & Coffee \& Beverage Ingredient Quality                                         &  &  \\
Drive-Through Service Quality                                        &  & Coffee \& Beverage Selection                                                  &  &  \\
Management, Staff Friendliness, Expertise   \& Professionalism       &  & Coffee \& Beverage Flavor Consistency                                         &  &  \\
Order Accuracy                                                       &  & Coffee \& Beverage Taste                                                      &  &  \\
Service Efficiency \& Speed/Wait Time                                &  & Coffee Preparation \& Brewing Quality                                         &  &  \\
                                                                     &  &                                                                               &  &  \\
\midrule
\textbf{Food \& Pastry (Freshness   \& Variety)}                     &  & \textbf{Digital Services \& Technology (Connectivity \&   Innovation)}        &  &  \\
\midrule
Food \& Pastry Flavor Consistency                                    &  & Digital Payment Methods                                                       &  &  \\
Food \& Pastry Ingredient Quality                                    &  & Mobile \& Online Ordering                                                     &  &  \\
Food \& Pastry Taste                                                 &  & Wifi Connectivity \& Power Outlets                                            &  &  \\
Food \& Pastry Selection                                             &  &                                                                               &  &  \\
                                                                     &  &                                                                               &  &  \\
\midrule
\textbf{Price/Value \& Promotions   (Value \& Affordability)}        &  & \textbf{Environment \& Sustainability (Eco‐Friendliness, Ethical   Sourcing)} &  &  \\
\midrule
Discounts \& Refills                                                 &  & Energy \& Water Use Efficiency                                                &  &  \\
Loyalty, Rewards \& Membership   Benefits                            &  & Ethical Coffee Sourcing/Fair Trade                                            &  &  \\
Value for Money                                                      &  & Waste Reduction \& Recycling \\
\bottomrule
\end{tabular}
\end{sidewaystable}

Our discovery approach identified ten main attributes and their corresponding features, listed in Table \ref{tab:attribtues_features}. 
These attributes capture the key dimensions of how customers evaluate their coffee shop experience. They span service interactions and operational efficiency, the quality and consistency of coffee, beverages, and food, and multiple aspects of the store environment, including ambiance \& atmosphere, comfort, cleanliness, and accessibility. They also reflect digital services and technology, price and value perceptions, and broader concerns such as sustainability and ethical sourcing. These attributes provide a comprehensive framework for understanding customer evaluations of Starbucks and similar coffeehouse experiences.

The features associated with each attribute in Table~\ref{tab:attribtues_features} capture distinct, actionable aspects of the customer experience. For instance, under Customer Service, features span conflict resolution, consistency, order accuracy, efficiency, professionalism, and drive-through quality. These features provide managers with clear levers to improve perceptions of service. Similarly, Coffee \& Beverage features cover taste, preparation quality, consistency, selection, and customization, while store-related attributes such as ambiance, comfort, and accessibility include features tied to aesthetics, seating, layout, parking, and operating hours. Other attributes also map to tangible levers, including food freshness and variety, digital amenities such as WiFi and mobile ordering, price/value perceptions, and sustainability practices. Collectively, these features provide the actionable detail behind higher-level attributes, allowing firms to identify where to intervene to strengthen customer satisfaction.

The discovered attributes and features align closely with the five SERVQUAL service quality dimensions \citep{parasuraman1988servqual}:
\begin{itemize}
    \itemsep0em
    \item \textbf{Tangibles} map to ambiance, comfort, and digital services;
    \item \textbf{Reliability} maps to order accuracy, service consistency, and beverage taste;
    \item \textbf{Responsiveness} maps to conflict resolution, service speed, and drive-through quality;
    \item \textbf{Assurance} maps to staff professionalism and accessibility;
    \item \textbf{Empathy} maps to sustainability, value perceptions, and inclusivity.
\end{itemize}

Our list is also consistent with the broader service literature, which has demonstrated the importance of ambiance and cleanliness \citep{wakefield1999customer}, order accuracy, timeliness, and product consistency \citep{sulek2004relative}, and pricing, loyalty programs, and sustainability initiatives \citep{konuk2019influence, keh2006reward}. Prior studies also emphasize staff professionalism and empathy \citep{alhelalat2017impact}, store layout and comfort \citep{almohaimmeed2017restaurant}, and technology-enabled services such as mobile ordering and WiFi \citep{dixon2009customer} as critical drivers of satisfaction and loyalty.

In sum, the alignment with SERVQUAL and the broader literature provides evidence of face validity for our approach. In addition, as we show later, we obtain low-to-moderate correlations between the attribute sentiments, indicating they capture distinct dimensions of the customer experience and providing evidence of discriminant validity.

Next, we describe the confirmatory step for extracting attributes and features.

\subsection*{Attribute and Feature Identification \& Sentiment Scoring}

In this step, each review is analyzed to detect the presence of attributes and features identified in Step 1 and to score their sentiment, using guided prompts that incorporate contextual cues to ensure accurate interpretation, minimize hallucinations and omissions, and maintain consistency across reviews. Algorithm~\ref{alg:attribute_feature_sentiment} outlines our procedure, which iterates over the corpus one review at a time. First, the full review is passed to the LLM to assess the overall sentiment of the customer on a scale from 1 to 5 (1= strongly negative and 5=strongly positive) and to enable it to capture the broader context of the review. The detailed prompt is available in Prompt~\ref{prompt:Review_sentiment} in Web Appendix~\ref{sec:att_disc_prompts}.

\begin{algorithm}[htbp]
\caption{Structured Attribute and Feature Sentiment Extraction}
\scriptsize
\label{alg:attribute_feature_sentiment}
\begin{algorithmic}[1]
\State \textbf{Input:} Predefined list of attributes and associated features, corpus of $M$ reviews
\State \textbf{Goal:} Generate a structured dataset indicating, for each review, the mentioned attributes and features along with their sentiment scores

\For{each review in the corpus}
    \State \textbf{Task 1: Overall Sentiment}
    \State \hspace{1em} Present the full review to the LLM to assess overall sentiment

    \Statex
    \State \textbf{Task 2: Attribute-Level Processing}
    \For{each sentence in the review}
        \State Assign the sentence to one or more predefined attributes
    \EndFor
    \For{each attribute assigned}
        \State Assess sentiment toward the attribute based on its associated sentences
    \EndFor

    \Statex
    \State \textbf{Task 3: Feature-Level Processing}
    \For{each sentence associated with an attribute}
        \State Assign the sentence to one or more of the attribute's features
    \EndFor
    \For{each identified feature}
        \State Assess sentiment toward the feature based on its associated sentences
    \EndFor

    \Statex
    \State \textit{Note:} If a sentence is not associated with any attribute or feature, classify it as \texttt{“Other Attributes”} or \texttt{“Other Features”}
\EndFor

\State \Return Structured dataset with attribute and feature mentions and associated sentiment scores for each review
\end{algorithmic}
\end{algorithm}

The review is then split into sentences. Such splitting is important because it enables the LLM to focus on specific attributes or features rather than parsing multiple ideas in longer texts, thereby improving granularity \citep{buschken2020improving}. Shorter inputs also reduce hallucinations and omissions by limiting the model’s context load \citep{liu2025towards}, while standardizing sentences as the unit of analysis enhances classification consistency. This approach further allows the capture of fine-grained sentiment shifts within the same review (e.g., positive toward Coffee \& Beverage but negative toward Customer Service) \citep{chakraborty2022attribute}. Finally, mapping sentences directly to attributes and features supports interpretability by providing an audit trail from structured outputs back to the review text.

For each sentence, the LLM is tasked with assigning the sentence to {\it one or more} of the pre-defined attributes identified in Step 1 while considering the surrounding sentences for contextual accuracy and, when necessary, revisiting the full review to ensure contextual accuracy. Sentences that cannot be matched to any attribute are classified as ``Other Attributes.''  For example, the sentence `drinks are decent, but expensive' refers to two attributes, Coffee \& Beverage and Price/Value, and should be assigned to both. By contrast, a sentence like `never going here again' does not map onto any of the attributes listed in Table~\ref{tab:attribtues_features} and is therefore classified as ``Other Attributes.'' See Prompt~\ref{prompt:sentence_attribute} in Web Appendix~\ref{sec:att_disc_prompts} for details. 

For each attribute, the LLM is then presented with all sentences associated with it and instructed to (1) evaluate customer sentiment on a 1–5 scale, allowing attribute sentiment to be captured in context rather than at the individual sentence level (See Prompt~\ref{prompt:att_sentiment} in Web Appendix~\ref{sec:att_disc_prompts} for more details) (2) assign sentences to {\it one or more} features of the attribute (Table~\ref{tab:attribtues_features}) and evaluate sentiment for each feature on a 1–5 scale.
Sentences that cannot be matched to any feature are classified as ``Other Features.''  
See Prompts~\ref{prompt:feat_sentiment} and \ref{prompt:feat_sentiment_class} in Web Appendix~\ref{sec:att_disc_prompts} for more details. 

We measure sentiment on a 5-point scale to represent the full range of emotions from strongly negative to strongly positive. For example, a sentence representing strong positive sentiment is: `I love Starbucks coffee.' A neutral or mixed sentiment sentence is: `The coffee is decent.' An example of strong negative sentiment is `TERRIBLE demeaning customer service. By using this scale, we aim to test the LLM ability to accurately perceive and rate the sentiments expressed throughout the review. 

Our multi-step prompting design enhances robustness and validity by breaking the task into smaller, guided steps rather than asking the LLM to extract everything in a single pass, a point we examine further in our prompt engineering experiment.
To minimize order bias, attributes and features are randomly presented. To improve accuracy, robustness, and interpretability, each prompt instructs the LLM to generate a chain-of-thought reasoning process~\citep{wei2022chain} before providing its final output. We use phrases like ``think step-by step''~\citep{kojima2022large} or by enforcing that it first fills out a ``reasoning'' field in its final output. This allows the model to leverage its reasoning process to help makes its final output, rather than, say, justifying its reasoning \textit{after} making its decision. Additionally, this reasoning process can provide potential understanding into how the LLM made its final output decision~\citep{wei2022chain}. Lastly, to enhance reproducibility and reduce randomness in outputs, we set the LLM temperature to 0, ensuring deterministic responses across runs.

\begin{table}[htbp]
  \centering
  \caption{ Attribute and Feature-Level Sentiment Extraction from an Illustrative Review}
    \scriptsize
    \begin{tabular}{llll}
    \toprule
    Coffee Shop Attribute & Sentiment & Feature & Sentiment \\
    \midrule
    Store Ambiance \& Atmosphere  & NA    & NA    & NA \\
    Customer Service & NA    & NA    & NA \\
    Coffee \& Beverage & 5     & Coffee \& Beverage Taste & 5 \\
    Food \& Pastry & NA    & NA    & NA \\
    Store Comfort \& Layout  & 5     & Workspace Quality & 5 \\
    Store Cleanliness \& Hygiene & NA    & NA    & NA \\
    Pricing \& Promotions & NA    & NA    & NA \\
    Facilities \& Accessibility & NA    & NA    & NA \\
    Digital Services \& Technology & 5     & Wifi Connectivity \& Power Outlets & 5 \\
    Sustainability \& Eco-Friendliness & NA    & NA    & NA \\
    \midrule
    \multicolumn{4}{l}{Note: Entries marked NA indicate that the reviewer did not mention such an information.}  \\
    \end{tabular}%
  \label{tab:example_review}%
\end{table}%

The final outcome of Step 2 is a structured dataset that transforms unstructured review data into structured outputs, specifying for each review the attributes and features mentioned and their associated sentiment. Table \ref{tab:example_review} illustrates this transformation using John’s review in Figure \ref{fig:examples}. In this example, the reviewer mentioned (i) Coffee \& Beverage, focusing on Taste; (ii) Store Comfort \& Layout, focusing on Workspace; and (iii) Digital Services, focusing on Free WiFi. Sentiment scores are provided at both the attribute and feature levels. 
    \section*{Validation with Human Annotators}
Human coders have long been the gold standard in content analysis due to their ability to capture context, cultural references, and subtle language cues \citep{bojic2025evaluating}. However, manually coding large volumes of unstructured reviews is slow, resource-intensive, and difficult to scale, as fatigue and inconsistency set in even for trained annotators \citep{culotta2016mining}. In contrast, LLMs can process reviews consistently and efficiently at scale, offering the potential to deliver timely, structured insights for managerial action \citep{brown2020language}. To ensure accuracy and contextual reliability, we validate our LLM outputs against human-coded benchmarks, which remain the practical reference for evaluating automated text analysis.

\subsection*{Method}
We evaluate eight LLM prompts on a random subset of 300 reviews, assessing performance through agreement with human annotations and predictive validity for customer ratings. Our goals are to measure alignment with human judgment, evaluate the extent of LLM hallucination, and examine the benefits of structured, context-aware prompting.  
This validation ensures that the LLM outputs are accurate, robust, and trustworthy for practical use. 

We employ a $2 \times 2 \times 2$ experimental design, resulting in eight sets of prompting strategies. These strategies vary in the inclusion of chain-of-thought~\citep{wei2022chain} reasoning (with vs. without), the LLM model used (GPT-4o mini vs. GPT-4.1 mini), and the level of analysis (sentence vs. review). In the review-level analysis, the LLM identifies attributes and features directly from the full review text without first splitting it into sentences. We accessed ChatGPT via its application programming interface (API).

We apply each strategy to the same set of 300 customer reviews, and the resulting outputs were evaluated against human-coded annotations. The sample of 300 reviews is statistically sufficient for validation while balancing content coverage, the feasibility of human annotation, and research budget constraints. Coders were compensated \$15 for annotating five reviews to encourage attentiveness during this demanding task.

In this experiment, human coders followed the same sentence-level coding guide used in our confirmatory phase, ensuring a consistent reference standard across all conditions. Because manual review-level coding is cognitively demanding and prone to fatigue, asking human coders to annotate reviews at this level would be highly challenging and unreliable.

This study was reviewed by the Institutional Review Board and determined to be exempt from IRB oversight (Protocol Number: IRB-AAAV6626, approval date: February 24, 2025). The exemption was granted given the minimal risk design, which involved human annotators coding textual reviews. 

We recruited ten human annotators (4 MBA, 4 MS, 2 PhD; average age 28; 7 female, 3 male; 9 fluent and 1 advanced in English; average of eight marketing courses completed). We used exactly the same process and language employed in training the LLM during the confirmatory phase to familiarize coders with the attributes and features and to train them on how to code the reviews. Details of the survey are provided in Web Appendix~\ref{sec:human_survey} and the training video can be requested from the authors. Before proceeding to the annotation phase, coders were required to complete a 10-question quiz (see Figure~\ref{fig:quiz} in Web Appendix~\ref{sec:human_survey}) assessing their familiarity with the attributes and features listed in Table \ref{tab:attribtues_features}, with a minimum score of 9 out of 10 required to qualify for the next task. 

As with LLMs, we asked each coder to carefully read the full review and assign an overall sentiment score on a 5-point scale ranging from 1 (strongly negative) to 5 (strongly positive). Next, each review was split into individual sentences, and coders were instructed to assign each sentence to one or more relevant attributes. They then evaluated the reviewer’s sentiment toward each attribute based on the assigned sentences. Finally, coders identified specific features mentioned within the sentences that related to the assigned attributes and rated the sentiment toward each identified feature. The order of attributes was randomized to minimize bias. See  Figures~\ref{fig:human_survey2} through \ref{fig:human_survey6} in Web Appendix~\ref{sec:human_survey} for more details.

Each coder annotated five reviews per session. We determined this number based on a pilot test conducted in our behavioral lab involving nine research assistants, which suggested that coder fatigue set in beyond this point. On average, each coder annotated 30 reviews. 

The pilot test showed that coding each review took about six minutes, varying by review length, and that the process became easier after the first review. Coders also suggested improvements such as sorting the attributes alphabetically, warning annotators about the potential presence of vulgar language, and strengthening the training with videos. Based on this feedback, we refined our instruments and created two video tutorials: one introducing the attributes and features, and another providing step-by-step instructions for completing the coding task.

Finally, coders took a median of six minutes to code one review. Their survey feedback indicated that the survey instructions were clear (10/10) and the task difficulty averaged 3.3 on a 5-point scale (1 = extremely easy, 5 = extremely difficult). Four coders out of 10 reported challenges in attribute/feature coding, particularly in scoring sentiment on a 5-point scale for certain reviews. Most coders (8/10) considered the survey time reasonable, though two felt it was too long. Finally, two coders suggested missing attributes/features, such as drink temperature. 


\subsection*{Validation Results}
Our analysis begins by examining the effects of three experimental factors: chain-of-thought reasoning (with vs. without), LLM model (GPT-4o mini vs. GPT-4.1 mini), and level of analysis (sentence vs. review). We evaluate their impact on two metrics: (i) raw agreement, which measures the extent to which LLMs and human coders identify the same set of attributes and features in a review, and (ii) Krippendorff’s $\alpha$ \citep{hayes2007answering}, which assesses the same consistency while adjusting for chance agreement. This analysis allows us to identify which prompting strategies yield the most reliable results and to guide the choice of the best-performing prompt for subsequent analyses.

Let Agreement = 1 if the LLM and human coders identify the same attribute or feature, and 0 otherwise. Define three dummy variables: GPT-4.1 = 1 if the LLM model is GPT-4.1 mini (0 if GPT-4o mini), Sentence = 1 if the analysis is at the sentence level (0 if review level), and Reasoning = 1 if chain-of-thought reasoning is included (0 otherwise). We estimate the following logistic regression ($\chi^2_{3}= 55.83, p < .001$; coefficient p-values are in parentheses):\footnote{All two-way and three-way interaction terms yield p-values above .05.}
{\setlength{\abovedisplayskip}{5pt}
 \setlength{\belowdisplayskip}{5pt}
$$
\textrm{Logit[Probability(Agreement=1)]}
= \underset{(p<.001)}{2.66}
+ \underset{(p<.001)}{.12}\textrm{GPT4.1}
+ \underset{(p=.012)}{.05}\textrm{Sentence}
+ \underset{(p<.001)}{.09}\textrm{Reasoning}.
$$
}

This analysis indicates that raw agreement is generally high across all prompting strategies, with each factor contributing positively to performance. The largest and statistically significant improvement comes from using GPT-4.1 mini, followed by modest but significant gains from the inclusion of reasoning and sentence-level analysis. These results suggest that GPT-4.1 mini with sentence-level reasoning provides the most reliable prompt for aligning LLM outputs with human coders.

\begin{figure}[htbp]
    \centering
    \includegraphics[width=\textwidth]{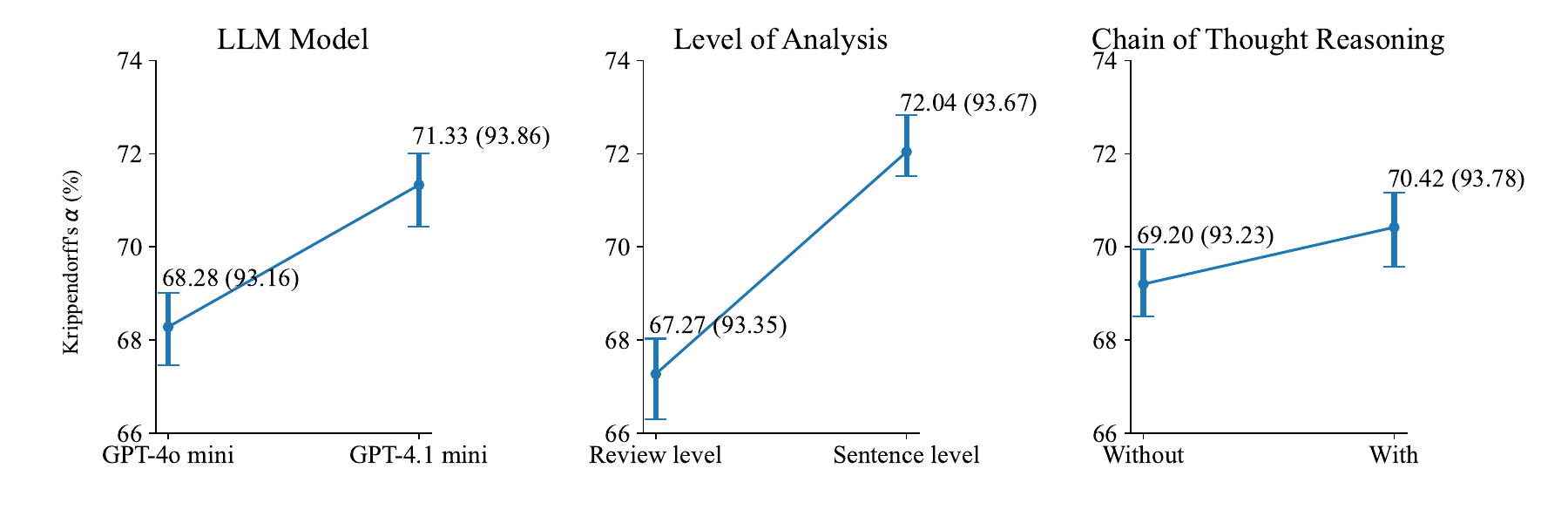}
    \textit{Note.} Numbers in parentheses are raw agreement in percentage.
    \caption{Krippendorff’s $\alpha$ by Experimental Condition}
    \label{fig:main_alpha}
\end{figure}

We arrive at similar conclusions when correcting for chance using Krippendorff’s $\alpha.$ Figure~\ref{fig:main_alpha} reports both raw and corrected agreement levels by condition. Raw agreement is consistently high (93\%–94\%), providing strong evidence that LLMs can reach human-level annotation. After correcting for chance, we observe significantly higher reliability when using GPT-4.1 mini compared to GPT-4o mini ($\alpha$ = 71.33\% vs. 68.28\%, an improvement of 3 points) and when analyzing at the sentence level rather than the review level ($\alpha$  = 72.04\% vs. 67.27\%, an improvement of 4.77 points). By contrast, the inclusion of chain-of-thought reasoning yields only a modest increase ($\alpha$ = 70.42\% vs. 69.20\%), with overlapping confidence intervals indicating no statistically significant effect. 
Overall, these results reinforce that the GPT-4.1 sentence-level prompting provides the most robust alignment with human coders.

These findings help explain why GPT-4.1 mini and sentence-level analysis perform best. GPT-4.1 mini benefits from improved model architecture and training, which enhance its alignment with human judgment and reduce hallucinations. Sentence-level prompting, in turn, compels the model to process smaller, more structured units of text, making it easier to capture context accurately, minimize spurious outputs, and avoid omissions. This advantage is evident in the extraction patterns: review-level prompting yields significantly fewer attributes and features than human coders, reflecting systematic omissions. On average, humans extracted 3.53 attributes (95\% CI: 3.38–3.67) and 4.43 features (95\% CI: 4.28–4.78) per review, compared with only 2.61 attributes (95\% CI: 2.54–2.67) and 3.80 features (95\% CI: 3.69–3.91) for review-level prompting. In contrast, sentence-level prompting closely mirrors human output, extracting 3.45 attributes (95\% CI: 3.37–3.52) and 4.91 features (95\% CI: 4.77–5.06). Together, these results highlight the value of structured, context-aware prompting for producing reliable, human-aligned annotations.

Based on these results, we identify the GPT-4.1 mini, sentence-level with reasoning configuration as the best-performing prompting strategy. We now validate it in detail against human annotations. This configuration achieves excellent agreement on attribute and feature mentions, with raw agreement of .95 (95\% CI: .94–.95) and Krippendorff’s $\alpha$ of .75 (95\% CI: .73–.76), approaching the conventional .80 threshold for strong reliability. It also mirrors human judgments of overall review sentiment, with a correlation of .94 (95\% CI: .93–.95). Next, we report the detailed validation results at the attribute and feature levels.

\paragraph{Validation at the Attribute Level}

\begin{figure}[htbp]
    \centering
    \begin{subfigure}[t]{0.45\textwidth}
        \centering
        \includegraphics[width=\linewidth]{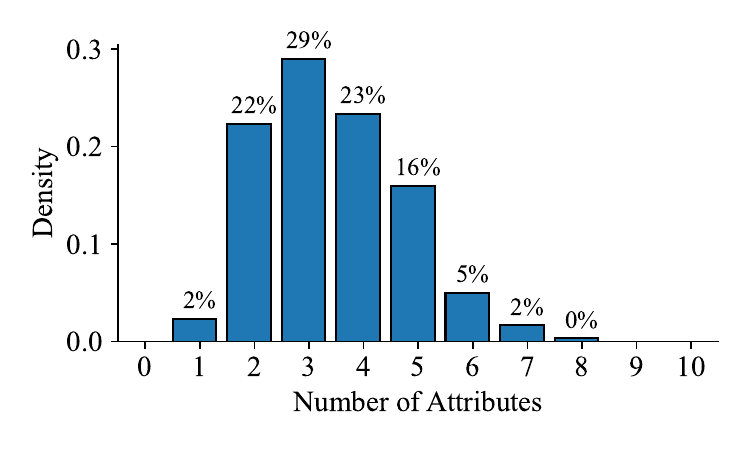}
        \caption{Human}
        \label{fig:hist_att_men_human_300}
    \end{subfigure}
    \hfill
    \begin{subfigure}[t]{0.45\textwidth}
        \centering
        \includegraphics[width=\linewidth]{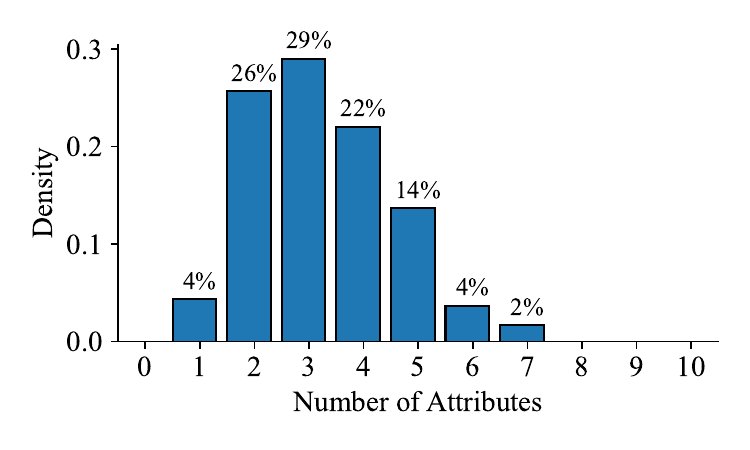}
        \caption{LLM}
        \label{fig:hist_att_men_gpt_300}
    \end{subfigure}
    \caption{Distributions of Attribute Mentions at the Review Level by Human and LLM}
    \label{fig:hist_att_men_300}
\end{figure}

For attribute mentions, we obtain raw agreement of .93 (95\% CI: .92-.94) between humans and LLM and Krippendorff’s $\alpha$ of .84 (95\% CI: .82-.86), indicating strong reliability. Figure~\ref{fig:hist_att_men_300} compares the distributions of the number of attributes mentioned per review by GPT-4.1 mini and human coders. The histograms are highly similar and statistically indistinguishable (Kolmogorov–Smirnov test=.05, p= .788).

When evaluating sentiment scoring, the strongest LLM–human agreement occurs under the 3-point scale (negative, neutral, positive) compared to the 5-point scale. In this case, raw agreement significantly improves from .81 (95\% CI: .80–.83) to .88 (95\% CI: .87–.89), and Krippendorff’s $\alpha$ rises significantly from .63 (95\% CI: .61–.65) to .76 (95\% CI: .74–.78), indicating strong reliability. This result is consistent with feedback from our human respondents in the pilot test, who reported difficulty applying fine-grained distinctions on the 5-point scale. 

Table~\ref{tab:comp_human_gpt_distr_ment} compares the distributions of attribute-level mentions and sentiment between GPT-4.1 mini and human coders on the 3-point sentiment scale. The two distributions are highly congruent. For example, human coders indicate that 89\% of reviews mention customer service (46\% positive, 43\% negative, and the remainder neutral), whereas the LLM yields nearly identical results, with 90\% mentions (43\% positive, 42\% negative).

\begin{table}[htbp]
  \centering
  \caption{Attribute-Level Mention and Sentiment Distributions by Humans and LLM}
  \scriptsize
    \begin{tabular}{llll}
    \toprule
          & \multicolumn{3}{c}{Human}       \\
\cmidrule{2-4}    Attribute & Mention (\%) & Positive (\%) & Negative (\%)    \\
    \midrule
    Customer Service	&	\barCell{89}{89}{blue}	&	\barCell{46}{46}{green}	&	\barCell{43}{43}{red}		\\
Coffee \& Beverage	&	\barCell{55}{55}{blue}	&	\barCell{25}{25}{green}	&	\barCell{17}{17}{red}		\\
Facilities \& Accessibility	&	\barCell{33}{33}{blue}	&	\barCell{16}{16}{green}	&	\barCell{13}{13}{red}	\\
Store Ambiance \& Atmosphere	&	\barCell{22}{22}{blue}	&	\barCell{17}{17}{green}	&	\barCell{5}{5}{red}\\
Store Comfort \& Layout	&	\barCell{19}{19}{blue}	&	\barCell{10}{10}{green}	&	\barCell{6}{6}{red}		\\
Store Cleanliness \& Hygiene	&	\barCell{13}{13}{blue}	&	\barCell{7}{7}{green}	&	\barCell{8}{8}{red}		\\
Digital Services \& Technology	&	\barCell{13}{13}{blue}	&	\barCell{7}{7}{green}	&	\barCell{4}{4}{red}	\\
Price/Value \& Promotions	&	\barCell{14}{14}{blue}	&	\barCell{4}{4}{green}	&	\barCell{7}{7}{red}	\\
Food \& Pastry	&	\barCell{11}{11}{blue}	&	\barCell{6}{6}{green}	&	\barCell{3}{3}{red}		\\
Environment \& Sustainability	&	\barCell{1}{1}{blue}	&	\barCell{0}{0}{green}	&	\barCell{1}{1}{red}	\\
    \midrule
    \midrule
                 & \multicolumn{3}{c}{GPT 4.1-mini} \\
\cmidrule{2-4}    Attribute & Mention (\%) & Positive (\%) & Negative (\%) \\
    \midrule
    Customer Service		&	\barCell{90}{90}{blue}	&	\barCell{43}{43}{green}	&	\barCell{42}{42}{red}	\\
Coffee \& Beverage		&	\barCell{45}{45}{blue}	&	\barCell{25}{25}{green}	&	\barCell{17}{17}{red}	\\
Facilities \& Accessibility			&	\barCell{31}{31}{blue}	&	\barCell{16}{16}{green}	&	\barCell{13}{13}{red}	\\
Store Ambiance \& Atmosphere		&	\barCell{23}{23}{blue}	&	\barCell{14}{14}{green}	&	\barCell{7}{7}{red}	\\
Store Comfort \& Layout			&	\barCell{18}{18}{blue}	&	\barCell{12}{12}{green}	&	\barCell{6}{6}{red}	\\
Store Cleanliness \& Hygiene		&	\barCell{15}{15}{blue}	&	\barCell{7}{7}{green}	&	\barCell{7}{7}{red}	\\
Digital Services \& Technology	&	\barCell{12}{12}{blue}	&	\barCell{6}{6}{green}	&	\barCell{3}{3}{red}	\\
Price/Value \& Promotions	&	\barCell{12}{12}{blue}	&	\barCell{4}{4}{green}	&	\barCell{9}{9}{red}	\\
Food \& Pastry			&	\barCell{9}{9}{blue}	&	\barCell{6}{6}{green}	&	\barCell{3}{3}{red}	\\
Environment \& Sustainability	&	\barCell{1}{1}{blue}	&	\barCell{1}{1}{green}	&	\barCell{1}{1}{red}	\\
    \bottomrule
    \end{tabular}%
  \label{tab:comp_human_gpt_distr_ment}%
\end{table}%

\begin{figure}[htbp]
    \centering
    \begin{subfigure}[t]{0.48\textwidth}
        \centering
        \includegraphics[width=\linewidth]{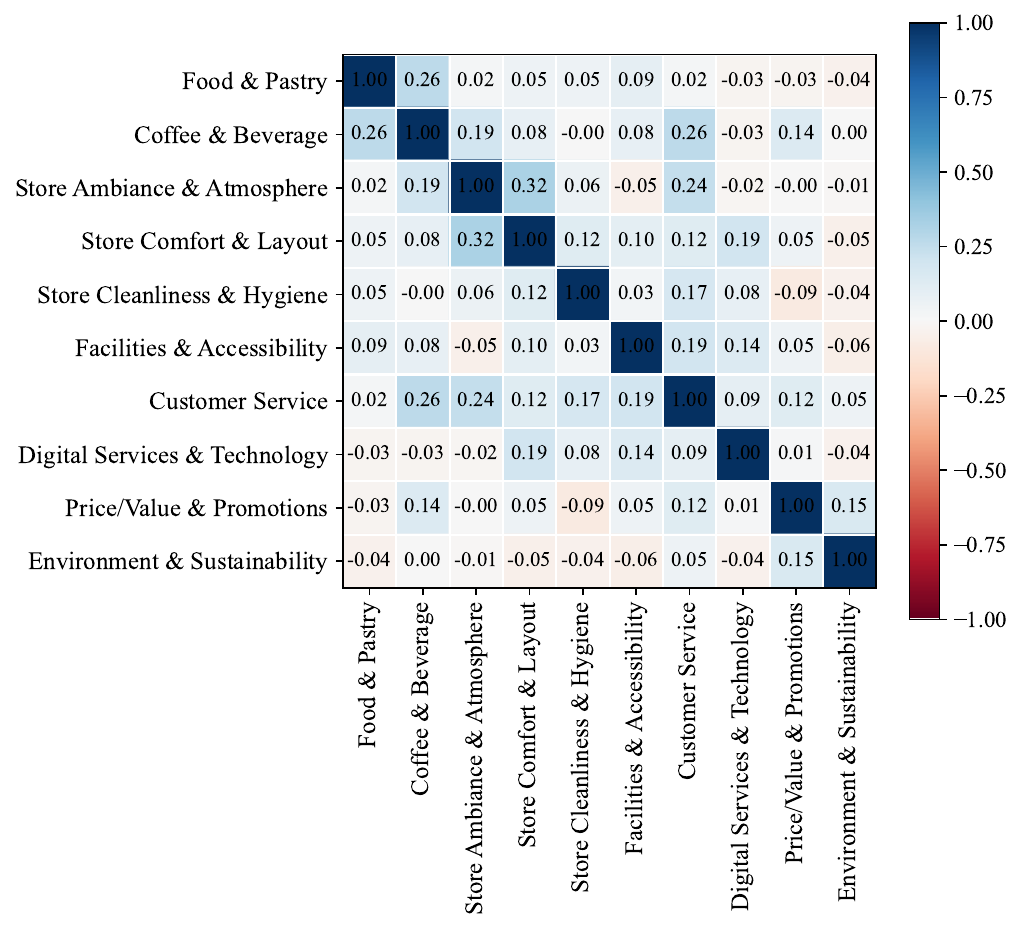}
        \caption{Human}
        \label{fig:factors_human}
    \end{subfigure}
    \hfill
    \begin{subfigure}[t]{0.48\textwidth}
        \centering
        \includegraphics[width=\linewidth]{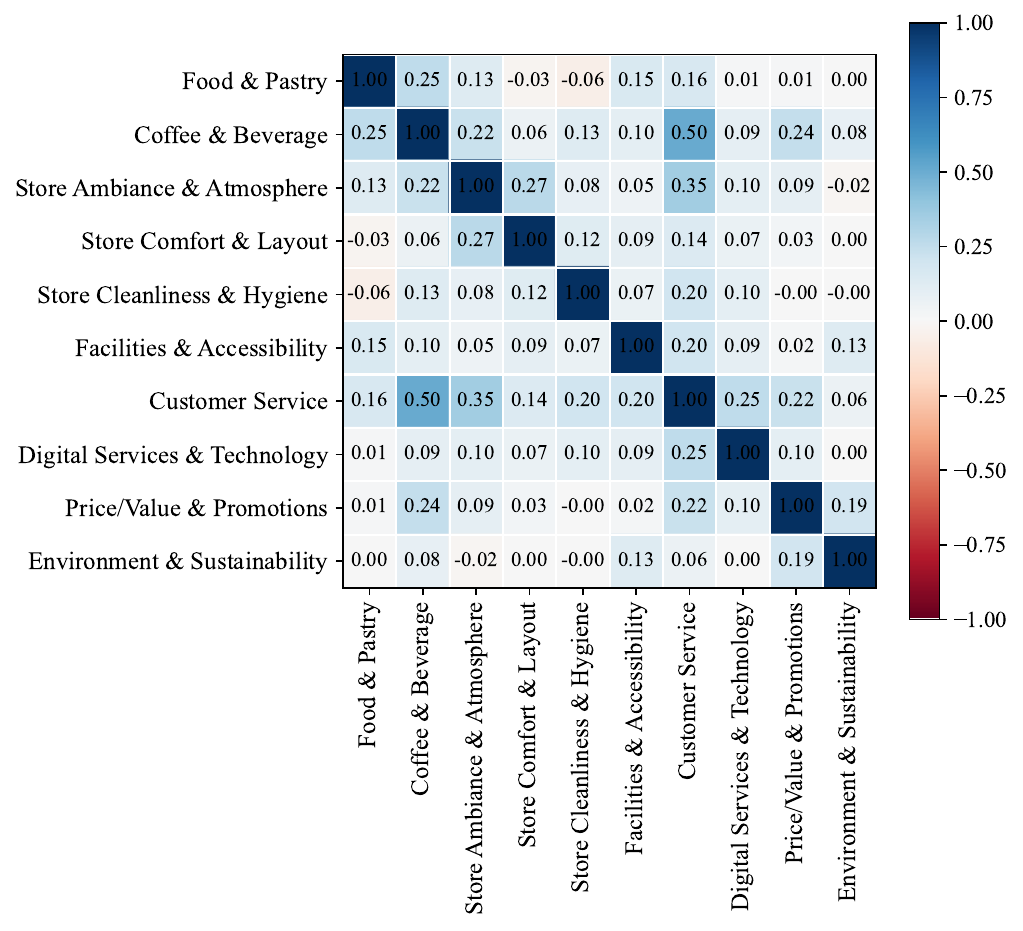}
        \caption{LLM}
        \label{fig:factors_gpt}
    \end{subfigure}
    \caption{Correlations of Attribute Sentiments by Human and LLM}
    \label{fig:factors}
\end{figure}

We also examine the correlations among the sentiment scores of the 10 attributes across the 300 reviews to assess whether the attributes capture distinct dimensions of the customer experience or overlap substantially. Figure~\ref{fig:factors} compares the correlation matrices for human coders and GPT-4.1 mini. In both cases, correlations are generally low to moderate, indicating that the ten attributes capture distinct aspects of the customer experience and that the results provide evidence of discriminant validity. A Jennrich test of equality of the two correlation matrices \citep{jennrich1970asymptotic} is insignificant ($\chi^2_{45} = 48.99$, $p = .316$). This similarity of patterns across GPT-4.1 mini and human coders suggests that the LLM preserves the structure of inter-attribute relationships, supporting the validity of sentence-level coding.

Finally, we test whether attribute sentiments identified by human annotators and LLMs are predictive of customers’ overall ratings. Both models achieve a high $R^2$ of .74, and the correlation between their regression coefficients is .96. This indicates that the attribute sentiments extracted by LLMs closely mirror those identified by humans and are equally predictive of overall satisfaction. This provides confidence that firms can rely on LLM-based extraction to generate predictive, human-comparable insights at scale.
See Web Appendix~\ref{sec:human_regression}.

\begin{figure}[htbp]
    \centering
    \begin{subfigure}[t]{0.45\textwidth}
        \centering
        \includegraphics[width=\linewidth]{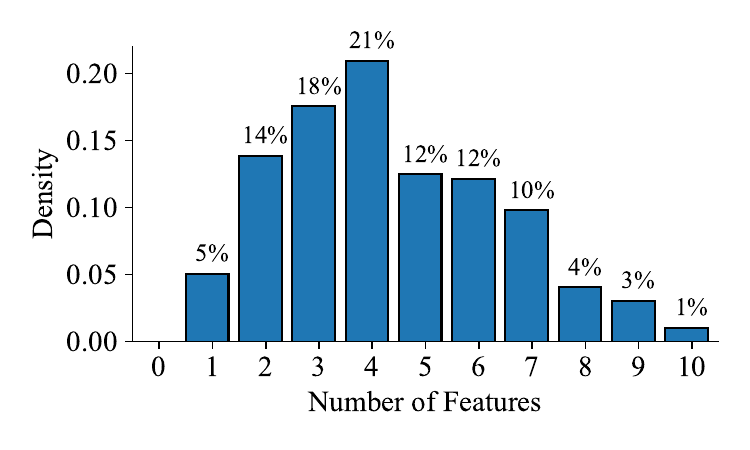}
        \caption{Human}
        \label{fig:hist_feat_men_human_300}
    \end{subfigure}
    \hfill
    \begin{subfigure}[t]{0.45\textwidth}
        \centering
        \includegraphics[width=\linewidth]{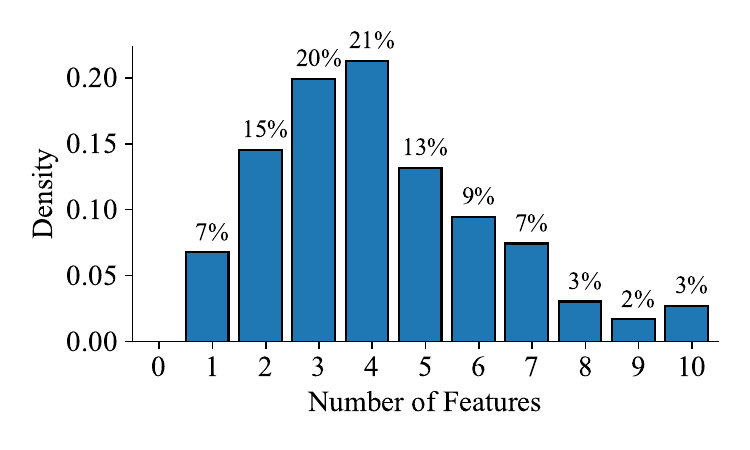}
        \caption{LLM}
        \label{fig:hist_feat_men_gpt_300}
    \end{subfigure}
    \caption{Distributions of Feature Mentions at the Review Level by Human and LLM}
    \label{fig:hist_feat_men_300}
\end{figure}

\paragraph{Validation at the Feature Level}
For feature mentions, we obtain raw agreement of .95 (95\% CI: .94-.95) between humans and LLM and Krippendorff’s $\alpha$ of .66 (95\% CI: .64-.68). Figure~\ref{fig:hist_feat_men_300} compares the distributions of the number of features mentioned per review by GPT-4.1 mini and human coders. The two histograms are highly similar and statistically indistinguishable (Kolmogorov–Smirnov test=.06, p= .722). 

Web Appendix~\ref{sec:feature_validation} (Table~\ref{tab:features_human_llm}) compares the distributions of feature-level mentions and sentiment between GPT-4.1 mini and human coders on the 3-point sentiment scale. As for attributes, the two distributions are highly congruent. 
For example, human coders indicate that 70\% of reviews mention staff friendliness (42\% positive, 26\% negative, and the remainder neutral), while the LLM produces nearly identical figures, with 72\% mentions (45\% positive, 26\% negative).

Overall, the findings indicate that our proposed sentence-level LLM approach provides a reliable and valid approximation of human-level performance in attribute/feature extraction and sentiment scoring. The inclusion of reasoning did not yield measurable performance gains but may add value for interpretability and diagnostic purposes. Between models, GPT-4.1 mini performed better than GPT-4o mini. Importantly, sentence-level extraction outperformed review-level extraction. This is consistent with prior research showing that review-level classification is prone to omission and misclassification because reviews often contain multiple sentiments and topics, making sentence-level analysis a more reliable prompting approach \citep{buschken2020improving, chakraborty2022attribute}.

Finally, sentiment agreement was captured more reliably on the 3-point scale than on the 5-point scale. This is consistent with feedback from our pilot study and exit survey, where coders reported difficulty in consistently scoring reviews on a 5-point scale. They noted challenges in distinguishing between adjacent categories (e.g., somewhat positive vs. positive), especially when reviews contained ambiguous or mixed sentiments. Similarly, LLMs struggled to mirror fine-grained distinctions on the 5-point scale, often collapsing sentiment into broader categories or showing lower agreement with human annotations. These results reinforce the utility of the 3-point scale as a more robust and interpretable measure for attribute- and feature-level sentiment analysis.

\section*{Empirical Results}
We present the empirical results from analyzing 12,682 reviews using our LLM approach to extract attributes, features, and sentiments. We then discuss the managerial implications.
\begin{figure}[H]
    \centering
    \begin{subfigure}[t]{0.45\textwidth}
        \centering
        \includegraphics[width=\linewidth]{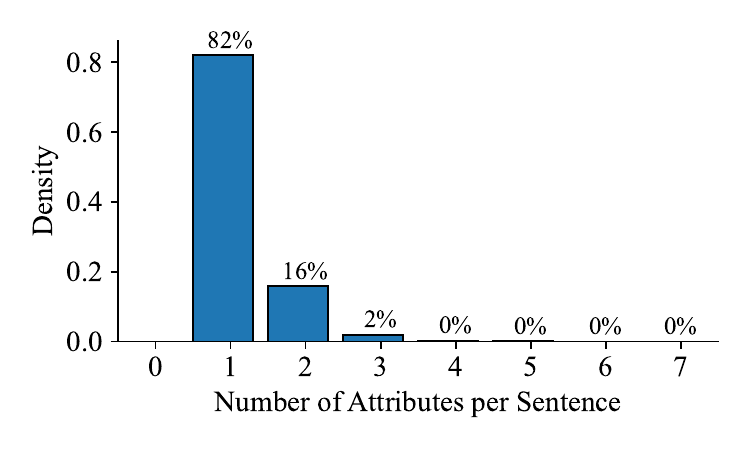}
        \caption{Sentence Level}
        \label{fig:num_att_ment_gpt}
    \end{subfigure}
    \hfill
    \begin{subfigure}[t]{0.45\textwidth}
        \centering
        \includegraphics[width=\linewidth]{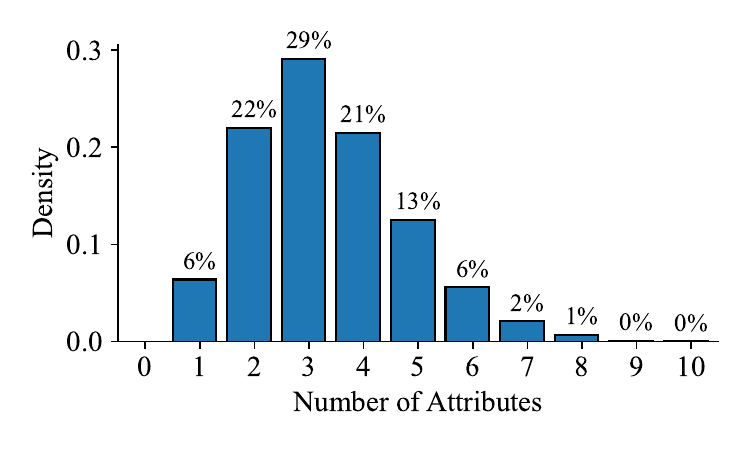}
        \caption{Review Level}
        \label{fig:hist_att_men}
    \end{subfigure}
    \caption{Distributions of Attribute Mentions at the Sentence and Review Levels}
    \label{fig:hist_att}
\end{figure}

On average, ChatGPT 4.1-mini took two seconds to code one review. We find that the LLM’s assessment of overall review sentiment correlates strongly with actual ratings ($r = .90$), indicating that it reliably infers sentiment from review text. Attribute mentions are highly concentrated at the sentence level: 82\% of sentences reference at most one attribute, with an average of 1.50 (see Figure~\ref{fig:num_att_ment_gpt}). This finding is consistent with \cite{sudhir2015peter}, who report that review sentences typically focus on a single topic. At the review level, the average number of attributes mentioned is three; few reviews discuss more than six, and almost none mention all ten attributes listed in Table~\ref{tab:attribtues_features} (see Figure~\ref{fig:hist_att_men}).

\begin{table}[htbp]
\centering
\caption{Attribute-Level Mention and Sentiment Distributions}
\scriptsize
\begin{tabular}{llll}
\toprule
{Attribute} & {Mention (\%)} & {Positive (\%)} & {Negative (\%)} \\
\midrule
Customer Service & \barCell{88}{88}{blue} & \barCell{46}{46}{green} & \barCell{42}{42}{red} \\
Coffee \& Beverage & \barCell{49}{49}{blue} & \barCell{27}{27}{green} & \barCell{18}{18}{red} \\
Facilities \& Accessibility & \barCell{37}{37}{blue} & \barCell{20}{20}{green} & \barCell{14}{14}{red} \\
Store Ambiance \& Atmosphere & \barCell{23}{23}{blue} & \barCell{17}{17}{green} & \barCell{5}{5}{red} \\
Store Comfort \& Layout & \barCell{22}{22}{blue} & \barCell{14}{14}{green} & \barCell{6}{6}{red} \\
Store Cleanliness \& Hygiene & \barCell{14}{14}{blue} & \barCell{8}{8}{green} & \barCell{6}{6}{red} \\
Food \& Pastry & \barCell{13}{13}{blue} & \barCell{6}{6}{green} & \barCell{5}{5}{red} \\
Digital Services \& Technology & \barCell{10}{10}{blue} & \barCell{5}{5}{green} & \barCell{4}{4}{red} \\
Price/Value \& Promotions & \barCell{10}{10}{blue} & \barCell{3}{3}{green} & \barCell{6}{6}{red} \\
Environment \& Sustainability & \barCell{0}{0}{blue} & \barCell{0}{0}{green} & \barCell{0}{0}{red} \\
\bottomrule
\end{tabular}
\label{tab:attribute_sentiment}
\end{table}

\subsection*{Attribute Mention and Sentiment}

Table~\ref{tab:attribute_sentiment} reports the distribution of mentions across the ten attributes, along with their associated positive and negative sentiment counts. These distributions closely mirror those in Table~\ref{tab:comp_human_gpt_distr_ment}, based on the random sample of 300 reviews coded by humans and GPT-4.1 mini, providing further validation of our approach.

Customer Service is the most frequently mentioned attribute, followed by Coffee \& Beverage and Facilities \& Accessibility. Mentions of Store Ambiance (23\%), Store Comfort \& Layout (22\%), and Store Cleanliness \& Hygiene (14\%) are less common individually, but together account for 59\% of mentions, underscoring the importance of the store environment in shaping the customer experience. Surprisingly, Environment \& Sustainability is  mentioned less than 1\% in reviews, neither positively nor negatively, raising questions about whether Starbucks’ initiatives in this area resonate with customers.

The right side of the table shows the distribution of positive and negative sentiment across attributes (neutral omitted since percentages sum to one). Customer Service, the most salient attribute, is also the most polarizing, with sentiment nearly evenly split—underscoring its centrality to the Starbucks experience but also its inconsistency. Coffee \& Beverage is evaluated more favorably, though 18\% negative sentiment indicates that product quality is not uniformly reliable; a similar pattern holds for Facilities \& Accessibility. Store-related attributes, including ambiance and comfort, are generally viewed positively. The remaining attributes received mixed evaluations. Overall, these results paint a mixed overall sentiment: customers value Starbucks’ beverages and store environment, but inconsistent service quality remains a critical vulnerability in the customer experience.

\subsection*{Feature Mention and Sentiment}
Table~\ref{tab:features_sentiment} reports the distribution of mentions across features, along with their associated positive and negative sentiment percentages. These distributions provide a detailed diagnostic of the specific, concrete aspects driving customer sentiment toward the broader attributes. Note that features with less than 3\% mentions are not reported in the Table.

As for the attribute level, features related to Customer Service and Coffee \& Beverage dominate. Within Customer Service, the most frequently mentioned feature in reviews is Staff Friendliness, Expertise, and Professionalism, followed by Service Efficiency \& Speed/Wait time and Order Accuracy. Within Coffee \& Beverage, Taste and Preparation \& Brewing Quality are the most salient. For Facilities \& Accessibility, Store Location Convenience is most frequently mentioned, while for Store Comfort \& Layout, Seating Availability \& Comfort stands out. These results underscore that customer evaluations focus most heavily on service interactions, beverage quality, and the store environment.

The sentiment distributions across features provide further insight. Within Customer Service, Staff Friendliness \& Professionalism attract more praise than criticism, whereas Service Efficiency \& Speed/Wait time draws more negative mentions than positive ones. For Coffee \& Beverage, sentiment is generally favorable but not uniformly so: coffee taste is viewed more positively than negatively, while coffee preparation and brewing quality emerge as a pain point. Store-related features present a mixed picture. Store Location Convenience performs strongly while Drive-Through Availability \& Quality reveal weaknesses. Store comfort features, such as seating and workspace quality, are evaluated positively but appear less frequently than other issues.

Overall, these results indicate that while Starbucks earns mixed sentiment for staff professionalism, coffee taste, and location convenience, recurring frustrations with service speed, order accuracy and drive-through access remain critical vulnerabilities. By pinpointing these pain points, our framework highlights concrete, actionable levers that Starbucks can target to reduce dissatisfaction and strengthen customer satisfaction.
 
\begin{table}[htbp]
\centering
\caption{Feature-level Mentions and Sentiment Distribution}
\label{tab:features_sentiment}
\scriptsize
\begin{tabular}{p{2.5cm} p{5.5cm}lll}
\toprule
Attribute & Feature & Mention (\%) & Positive (\%) & Negative (\%) \\
\midrule
Customer Service 
 & Management, Staff Friendliness, Expertise & \barCell{71}{71}{blue} & \barCell{43}{43}{green} & \barCell{28}{28}{red} \\
 & Service Efficiency \& Speed/Wait Time & \barCell{48}{48}{blue} & \barCell{20}{20}{green} & \barCell{26}{26}{red} \\
 & Order Accuracy & \barCell{26}{26}{blue} & \barCell{9}{9}{green} & \barCell{16}{16}{red} \\
 & Complaints \& Conflict Resolution & \barCell{16}{16}{blue} & \barCell{4}{4}{green} & \barCell{11}{11}{red} \\
 & Customer Service Consistency & \barCell{14}{14}{blue} & \barCell{6}{6}{green} & \barCell{8}{8}{red} \\
 & Drive-Through Service Quality & \barCell{13}{13}{blue} & \barCell{5}{5}{green} & \barCell{8}{8}{red} \\

 \midrule
 Coffee \& Beverage 
 & Taste & \barCell{25}{25}{blue} & \barCell{17}{17}{green} & \barCell{7}{7}{red} \\
 & Preparation \& Brewing Quality & \barCell{20}{20}{blue} & \barCell{9}{9}{green} & \barCell{11}{11}{red} \\
 & Selection & \barCell{16}{16}{blue} & \barCell{9}{9}{green} & \barCell{3}{3}{red} \\
 & Customization \& Personalization & \barCell{12}{12}{blue} & \barCell{6}{6}{green} & \barCell{5}{5}{red} \\
 & Flavor Consistency & \barCell{9}{9}{blue} & \barCell{5}{5}{green} & \barCell{3}{3}{red} \\
 & Ingredient Quality & \barCell{7}{7}{blue} & \barCell{3}{3}{green} & \barCell{4}{4}{red} \\
 \midrule
 \multirow{2}{*}{\parbox[c]{\linewidth}{\raggedright
Facilities \& Accessibility}}  
 & Store Location Convenience & \barCell{21}{21}{blue} & \barCell{16}{16}{green} & \barCell{3}{3}{red} \\
 & Drive-Through Availability \& Quality & \barCell{14}{14}{blue} & \barCell{5}{5}{green} & \barCell{7}{7}{red} \\
 & Parking Accessibility & \barCell{9}{9}{blue} & \barCell{3}{3}{green} & \barCell{6}{6}{red} \\
 & Store \& Online Operating Hours & \barCell{4}{4}{blue} & \barCell{2}{2}{green} & \barCell{2}{2}{red} \\
 \midrule
 \multirow{2}{*}{\parbox[c]{\linewidth}{\raggedright
Store Ambiance \& Atmosphere}} 
 & Sense of Community/Inclusivity & \barCell{7}{7}{blue} & \barCell{6}{6}{green} & \barCell{1}{1}{red} \\
& Interior Design \& Décor & \barCell{6}{6}{blue} & \barCell{5}{5}{green} & \barCell{1}{1}{red} \\
 & Music, Lighting, Noise & \barCell{6}{6}{blue} & \barCell{3}{3}{green} & \barCell{2}{2}{red} \\

 \midrule
\multirow{2}{*}{\parbox[c]{\linewidth}{\raggedright
Store Comfort \& Layout}} 
 & Seating Availability \& Comfort & \barCell{16}{16}{blue} & \barCell{10}{10}{green} & \barCell{5}{5}{red} \\
 & Indoor/Outdoor Seating & \barCell{8}{8}{blue} & \barCell{7}{7}{green} & \barCell{1}{1}{red} \\
 & Tables Arrangement & \barCell{5}{5}{blue} & \barCell{3}{3}{green} & \barCell{2}{2}{red} \\
  & Workspace Quality & \barCell{5}{5}{blue} & \barCell{4}{4}{green} & \barCell{1}{1}{red}  \\
 \midrule
\multirow{2}{*}{\parbox[c]{\linewidth}{\raggedright
Store Cleanliness \& Hygiene}}  
 & Store Cleanliness/Trash Disposal & \barCell{11}{11}{blue} & \barCell{7}{7}{green} & \barCell{4}{4}{red} \\
 \\
 \midrule
Food \& Pastry 
 & Selection & \barCell{7}{7}{blue} & \barCell{3}{3}{green} & \barCell{3}{3}{red} \\
 & Taste & \barCell{6}{6}{blue} & \barCell{4}{4}{green} & \barCell{2}{2}{red} \\
 \midrule
\multirow{2}{*}{\parbox[c]{\linewidth}{\raggedright
Digital Services \& Technology}}  
 & Mobile \& Online Ordering & \barCell{6}{6}{blue} & \barCell{3}{3}{green} & \barCell{3}{3}{red} \\
 & Wifi Connectivity \& Power Outlets & \barCell{4}{4}{blue} & \barCell{3}{3}{green} & \barCell{1}{1}{red} \\
 \midrule
\multirow{2}{*}{\parbox[c]{\linewidth}{\raggedright
Price/Value \& Promotions}}  
& Value for Money & \barCell{7}{7}{blue} & \barCell{1}{1}{green} & \barCell{5}{5}{red} \\
\\
\bottomrule
\multicolumn{5}{l}{{Note. Features with less than 3\% mentions are not reported.}}
\end{tabular}
\end{table}

Finally, we compared the predictive validity of our LLM-based approach with 25 state-of-the-art NLP methods, including bag-of-words, deep neural networks, and transformer-based models. Our approach outperforms these benchmarks in predictive accuracy while preserving interpretability. Detailed results are reported in Web Appendix \ref{sec:comp}.
    \section*{Generating Actionable Insights}
Our structured dataset of attribute- and feature-level sentiments enables granular analyses to identify issues and guide targeted actions to improve customer satisfaction. We highlight three dashboard applications: (1) tracking attribute sentiment dynamics over time, (2) visualizing variation in attribute and feature sentiments across stores nationwide, and (3) identifying high-leverage features for enhancing satisfaction at the aggregate and store levels.

\subsection*{Evolution of Attribute Sentiments over Time }
\begin{figure}[htbp]
    \centering
    \includegraphics[width=0.9\linewidth]{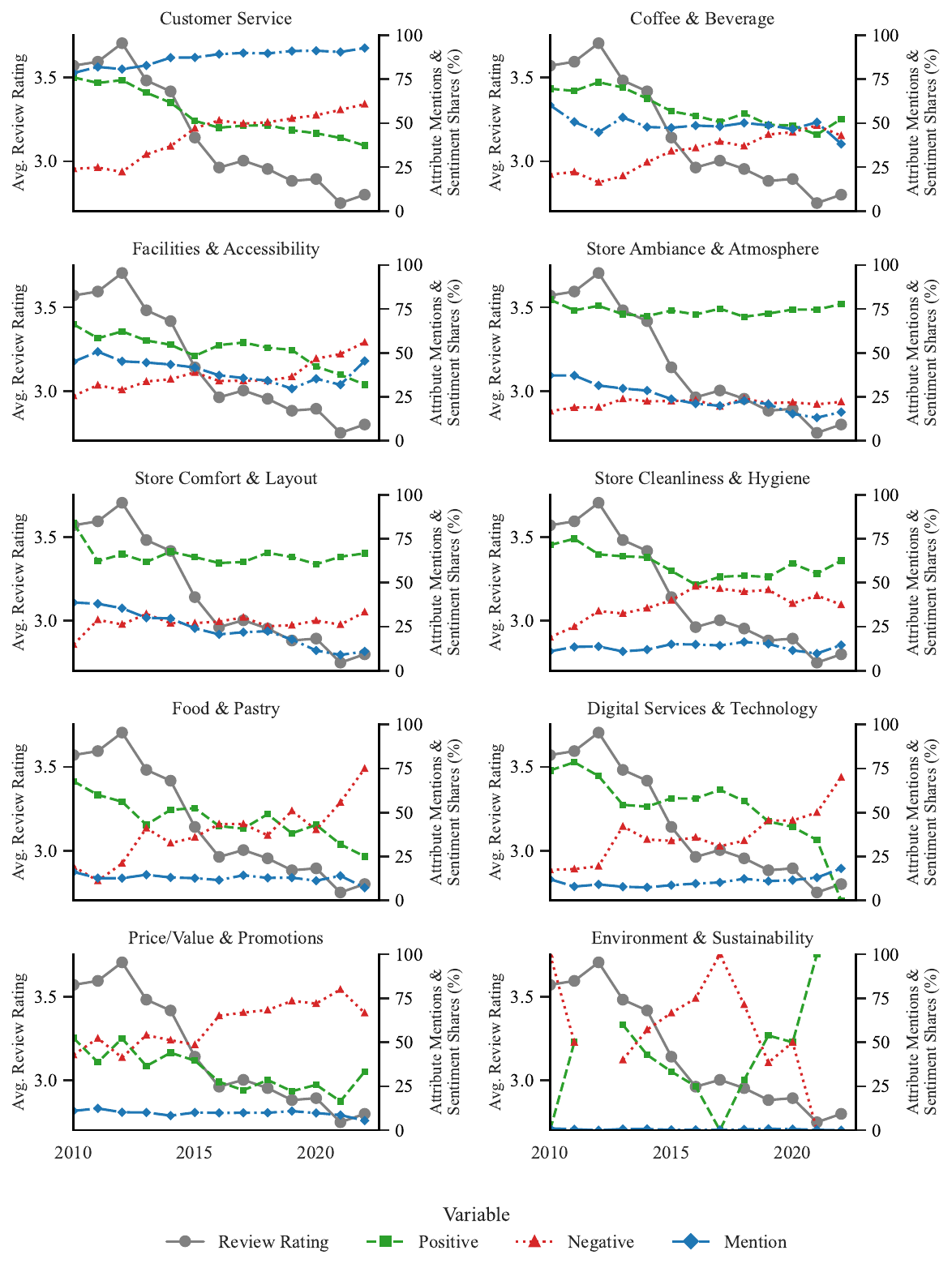}
    \caption{Time Evolution of Starbucks Ratings, Attribute Mentions, and Sentiment}
    \label{fig:attribute_sentiment_trends}
\end{figure}

Figure~\ref{fig:attribute_sentiment_trends} displays the trends in Starbucks’ average (i) customer ratings, (ii) attribute mentions, and (iii) the shares of positive and negative sentiment (defined as the proportion of positive or negative responses among all non-neutral responses) for each of the ten attributes in Table~\ref{tab:attribtues_features} over the 15-year span of our dataset. Similar to the Net Promoter Score (NPS), which contrasts promoters and detractors, we use the shares of positive and negative sentiment to capture the balance of favorable versus unfavorable evaluations.
Attribute mentions exhibit heterogeneous dynamics over time, indicating that customers’ topics of interest have shifted. Mentions of {Store Ambiance \& Atmosphere} and {Store Comfort \& Layout} have declined, potentially reflecting Starbucks’ shift from a ``third place'' concept (i.e., a social space between home and work for community and connection) to a ``grab-and-go'' coffee shop model \citep{oldenburg1997our}. By contrast, attributes such as {Coffee \& Beverage}, {Food \& Pastry}, and {Price/Value} remain relatively stable. Most notably, Customer Service has gained prominence, with mentions rising from 79\% of reviews in 2010 to 93\% in 2022. This trend underscores that service interactions have become an increasingly critical driver of the customer experience and a key area for managerial attention.

Across attributes, we observe a broad decline in the balance of positive versus negative sentiment over time. In the early years of the dataset, the share of positive sentiment (green) was consistently higher than the share of negative sentiment (red) across most attributes, reflecting a predominance of favorable evaluations. Over time, however, these trends converge, with positive sentiment steadily losing ground while negative sentiment becomes more prominent. By the end of the period, the red line surpasses the green for several attributes signaling a shift toward more critical customer evaluations.

Customer Service is a case in point. In 2010, the odds of positive-to-negative sentiment were roughly 3:1, reflecting a clear predominance of favorable evaluations. By 2022, this pattern had reversed: these fell below 1, while the odds of negative-to-positive climbed above 1.5. The two trends intersected around 2016, marking the point when negative sentiment began to dominate. This intersection year coincides with a turning point in Starbucks’ employee relations. According to Harvard Business Review (2024),\footnote{\url{https://hbr.org/2024/06/how-starbucks-devalued-its-own-brand}} 2016 marked the start of a cultural shift as leadership prioritized speed, efficiency, and digital transactions over personal connection with customers. These changes heightened performance pressures on employees and eroded the company’s `third place' ethos, creating widespread dissatisfaction and fueling unionization efforts. Such organizational tensions provide external validation for the deterioration in customer service sentiment revealed by our analysis. Thus, these findings highlight the tight link between employee experience and customer experience, underscoring that sustaining service quality will require investment in both.

\subsection*{Attribute Sentiments Across Stores}

\begin{figure}[htbp]
    \hspace{1cm}
    \includegraphics[width=\linewidth]{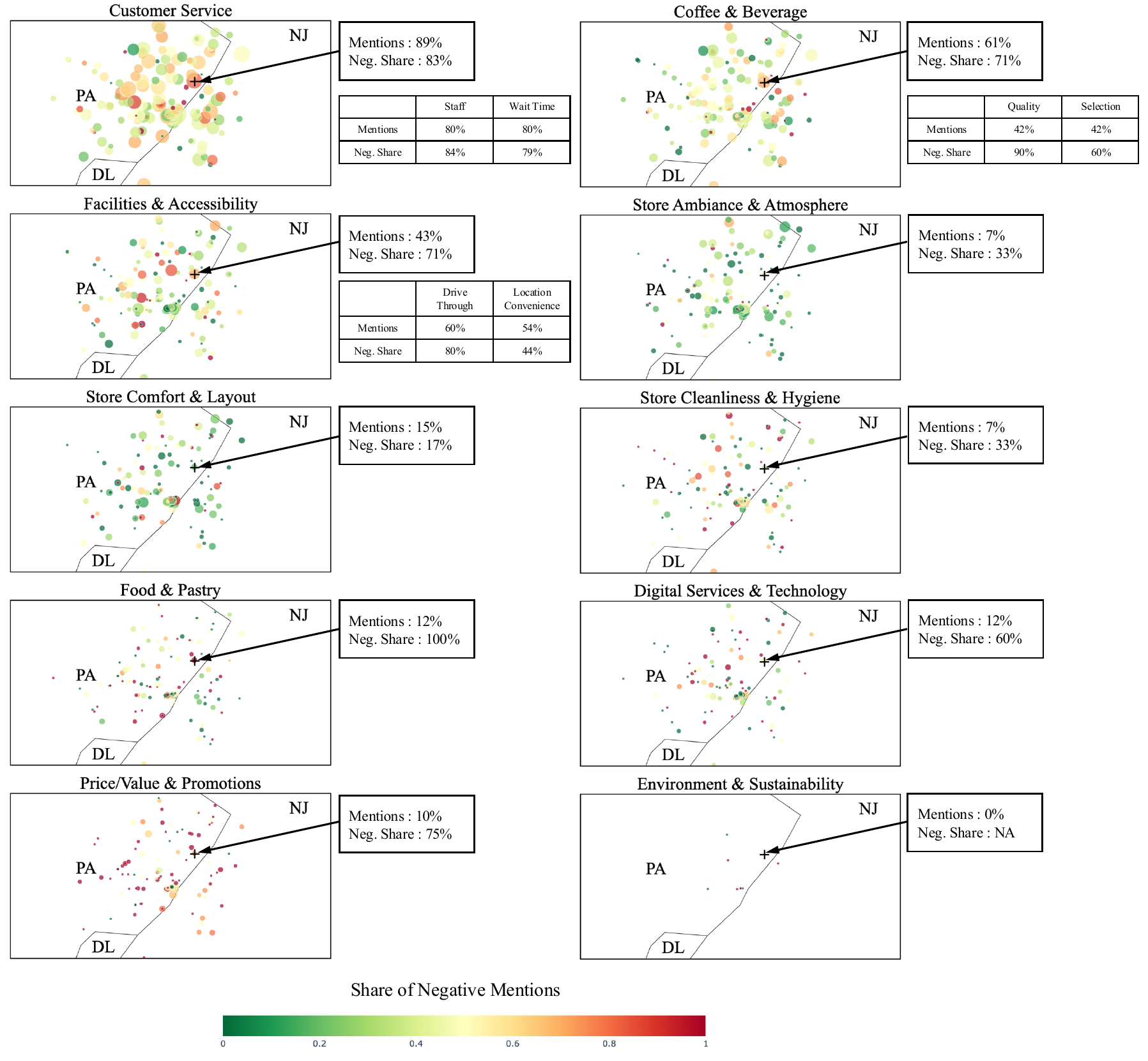}
    \captionsetup{justification=centering}
    \caption{Store-Level Variations in Attribute Sentiment}
    \label{fig:spacial_results}
\end{figure}

Figure~\ref{fig:spacial_results} presents store-level attribute mentions and associated sentiments for Starbucks locations in New Jersey and Pennsylvania across the ten attributes in heat map format.\footnote{We focus on these two states because displaying all 722 stores in our sample within a single figure is infeasible.} In the figure, larger bubbles represent attributes with a higher percentage of mentions in reviews, while bubble color reflects sentiment valence. Greener bubbles indicate a higher share of positive sentiment and redder bubbles indicate a higher share of negative sentiment. As in Figure~\ref{fig:attribute_sentiment_trends}, these shares are computed among all non-neutral responses. 
Consider the store highlighted in the figure
under Customer Service (top-left side of figure), represented by a large, reddish bubble. For this store, 89\% of reviews mention customer service, with 83\% negative (measured among non-neutral mentions). Managers can drill down further to see the drivers of this dissatisfaction. Among the reviews citing customer service, 80\% specifically mention Staff Professionalism and/or Service Efficiency/Wait Time negatively 84\% and 79\%, respectively (again percentages are computed among non-neutral mentions). If desired, the dashboard can also surface representative reviews mentioning customer service to provide managers with a more concrete picture of the issues. Here is an example of such reviews of this store: “One of the worst Starbucks I've been to. I've had to wait 25–35 minutes for one drink I ordered on the app. Slow service, rude staff."

For the same store, Coffee \& Beverage is the second most mentioned attribute (61\% of reviews), with 71\% of those mentions negative. The dashboard further shows that this dissatisfaction is mainly driven by Coffee Taste and Preparation \& Brewing Quality, both recurring sources of negative feedback.

The third most mentioned attribute is Facilities \& Accessibility (43\% of reviews), with 71\% of mentions negative. Within this attribute, features such as Drive-Through Quality and Parking Accessibility account for much of the discontent, signaling that convenience and access are pain points for this store.

The attribute- and feature-level diagnostics in Figure~\ref{fig:spacial_results} provide managers with a clear picture of where and why this store underperforms, allowing them to prioritize targeted improvements.
Such a dashboard provides managers with granular, location-specific insights that go beyond average ratings. By showing which attributes and features drive customer satisfaction at each store, it enables managers to separate systemic issues (e.g., widespread complaints about service speed or beverage quality) from location-specific problems (e.g., access or parking at a particular store). In doing so, the dashboard transforms unstructured review data into actionable intelligence that supports day-to-day decision-making and helps prioritize targeted improvements across the store network.

Overall, the figure reveals high variability in customer sentiment across stores and attributes. Starbucks can leverage this variability by sharing best practices from high-performing locations, while recognizing that some differences reflect local customer expectations and demand-side factors rather than store operations alone. These insights provide valuable diagnostics for benchmarking and improving attribute-specific performance. More broadly, they help managers distinguish between systemic issues and location-specific problems, turning unstructured review data into actionable intelligence for day-to-day decision-making.


\subsection*{Identifying High-Leverage Attributes and Features for Enhancing Customer Satisfaction}

We now use our structured data to identify attributes and features with high impact on customer satisfaction. Specifically, we assess how changes in the sentiment of a given attribute or feature is associated with changes in a review's rating.  Ideally, such an assessment should be conducted through an experiment in which sentiment is exogenously manipulated. However, such procedure is challenging. First, sentiment is inherently subjective. Second, even if sentiment manipulation were feasible, it might alter the customer’s reviewing behavior. Customers tend to self-select when leaving reviews: extremely dissatisfied customers are more likely to leave negative reviews, while highly satisfied customers tend to leave very positive ones \citep{chen2021reviews,schoenmueller2020polarity}. As a result, drawing causal conclusions is not feasible in our current analysis. Nonetheless, our analysis characterizes the association between actionable attributes/features and review ratings. Although this approach does not yield pure counterfactual estimates in the causal inference tradition, it provides firms with a practical diagnosis to identify where to begin and what to expect (correlationally) in terms of features' impact on customer satisfaction.

Our analysis separately regresses customer ratings on (i) attribute-level and (ii) feature-level sentiments. For each attribute or feature, we define four dummy variables (positive, neutral, negative, and not mentioned, with the latter indicating that the attribute/feature does not appear in the review) and use negative sentiment as the reference category. In addition, we incorporate meta-data from Yelp, including fixed effects for Starbucks store location, review year, and the year the reviewer joined Yelp. We also control for the number of years the reviewer held elite status prior to the review, with elite status granted by Yelp to reviewers who consistently produce helpful content. These meta-data variables help mitigate endogeneity from selection and account for observed reviewer heterogeneity. Standard errors are clustered at the store level to adjust for potential correlation due to store-level selection.

For each attribute- and feature-level regression, we estimate two models. The first includes only the corresponding sentiment variables extracted from our structured data. The second augments this specification by adding all metadata fixed effects. The aim is to compare predictive performance with and without contextual controls and to assess the robustness of our estimates across specifications. Because it is rarely mentioned, Environment \& Sustainability is excluded from the attribute-level regression along with its associated features from the feature-level regression.

\subsubsection*{Identifying High-Leverage Attributes}
Table~\ref{tab:attr_ff} reports the attribute-level regression results. Both models yield identical adjusted $R^2$ values of .71, indicating that adding metadata controls provides no meaningful incremental explanatory power beyond attribute-level sentiments. The regression coefficients and their significance levels are nearly identical across the two specifications. All attributes are statistically significant, underscoring their role as key drivers of customer satisfaction.
\begin{table}[htbp]
  \centering
  \caption{Attribute Regressions With and Without Control Variables}
  \scriptsize
  \begin{tabular}{lcccccc@{\hskip .5cm}cccccc}
    \toprule
    & \multicolumn{6}{c}{\textbf{Model 1 (Without Controls)}} & \multicolumn{6}{c}{\textbf{Model 2 (With Controls)}} \\
    \cmidrule(lr){2-7} \cmidrule(lr){8-13}
    & \multicolumn{3}{c}{Neutral} & \multicolumn{3}{c}{Positive} 
    & \multicolumn{3}{c}{Neutral} & \multicolumn{3}{c}{Positive} \\
    \cmidrule(lr){2-4} \cmidrule(lr){5-7} \cmidrule(lr){8-10} \cmidrule(lr){11-13}
    \textbf{Feature} & Coef. & SE & $p$-value & Coef. & SE & $p$-value 
                     & Coef. & SE & $p$-value & Coef. & SE & $p$-value \\
    \midrule
    Customer Service & 1.27 & .07 & $< .001$ & 2.24 & .02 & $< .001$ 
                     & 1.26 & .08 & $< .001$ & 2.18 & .03 & $< .001$ \\
    Coffee \& Beverage & .45 & .04 & $< .001$ & .78 & .03 & $< .001$ 
                       & .43 & .04 & $< .001$ & .76 & .03 & $< .001$ \\
    Facilities \& Accessibility & .29 & .04 & $< .001$ & .44 & .03 & $< .001$
                                & .30 & .05 & $< .001$ & .44 & .03 & $< .001$ \\
    Store Ambiance \& Atmosphere & .27 & .08 & $< .001$ & .49 & .04 & $< .001$
                                 & .26 & .09 & .005 & .48 & .05 & $< .001$ \\
    Store Comfort \& Layout & -.02 & .07 & .793 & .18 & .04 & $< .001$
                            & -.02 & .07 & .741 & .16 & .04 & $< .001$ \\
    Store Cleanliness \& Hygiene & .33 & .17 & .050 & .49 & .04 & $< .001$
                                 & .26 & .17 & .133 & .47 & .05 & $< .001$ \\
    Food \& Pastry & .18 & .07 & .005 & .31 & .05 & $< .001$
                   & .16 & .06 & .011 & .28 & .05 & $< .001$ \\
    Digital Services \& Technology & .16 & .09 & .059 & .23 & .05 & $< .001$
                                   & .17 & .10 & .084 & .24 & .06 & $< .001$ \\
    Price/Value \& Promotions & .13 & .10 & .191 & .47 & .05 & $< .001$
                              & .14 & .10 & .173 & .46 & .06 & $< .001$ \\
    \midrule
    Business FE & \multicolumn{6}{c}{No} & \multicolumn{6}{c}{Yes} \\
    Year FE & \multicolumn{6}{c}{No} & \multicolumn{6}{c}{Yes} \\
    Reviewer Controls & \multicolumn{6}{c}{No} & \multicolumn{6}{c}{Yes} \\
    Missing Features & \multicolumn{6}{c}{Yes} & \multicolumn{6}{c}{Yes} \\
    Nb. Obs. & \multicolumn{6}{c}{12,682} & \multicolumn{6}{c}{12,682} \\
    $R^2$ & \multicolumn{6}{c}{.71} & \multicolumn{6}{c}{.73} \\
    Adj. $R^2$ & \multicolumn{6}{c}{.71} & \multicolumn{6}{c}{.71} \\
    \bottomrule
    \multicolumn{13}{l}{\textit{Note.} Environment \& Sustainability was excluded from the analysis.} \\
  \end{tabular}
  \label{tab:attr_ff}
\end{table}

\begin{figure}[htbp]
    \centering
    \begin{subfigure}[t]{0.43\textwidth}
        \centering
        \includegraphics[width=\linewidth]{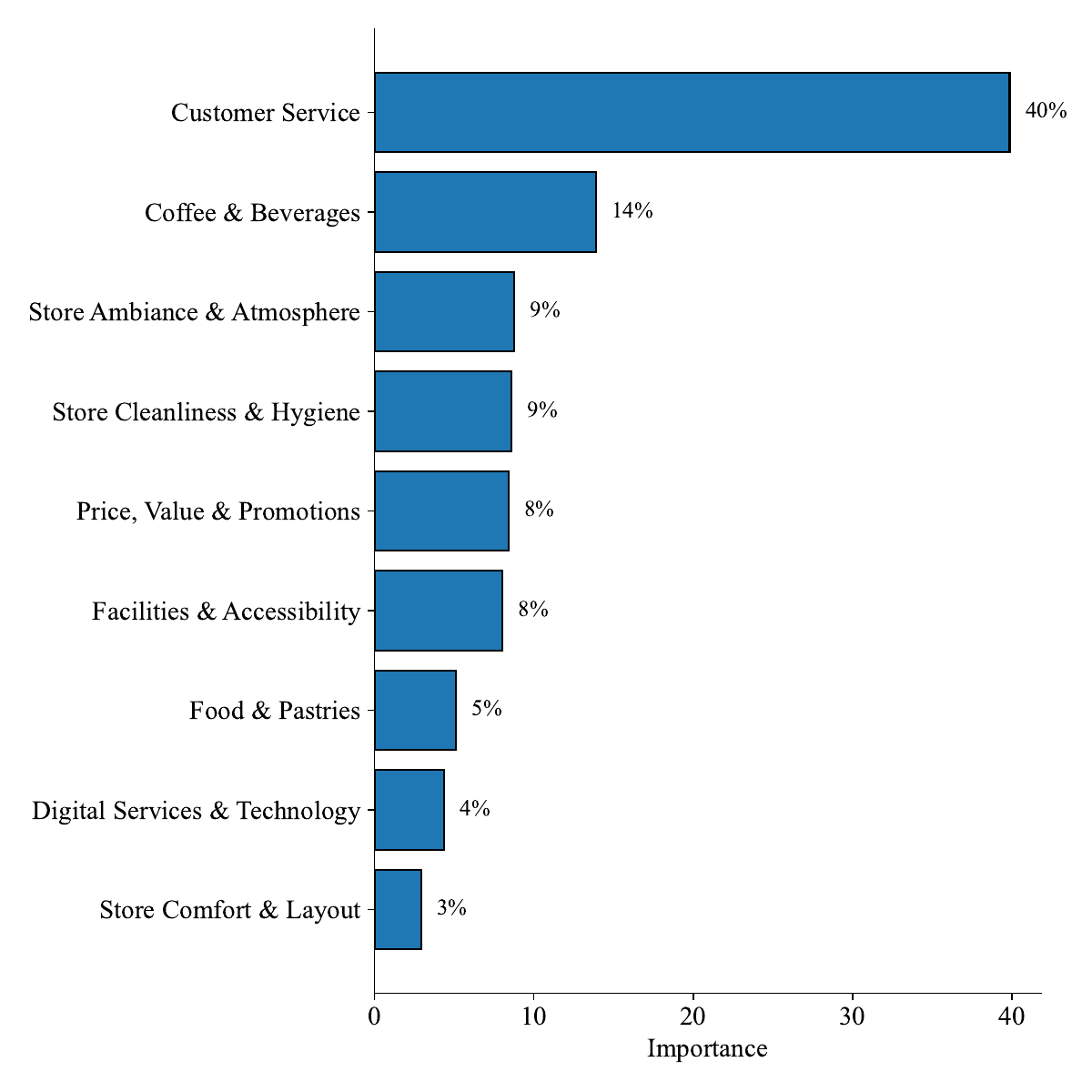}
        \caption{Relative Attribute Importance}
        \label{fig:attribute_importance}
    \end{subfigure}
    \hfill
    \begin{subfigure}[t]{0.43\textwidth}
        \centering
        \includegraphics[width=\linewidth]{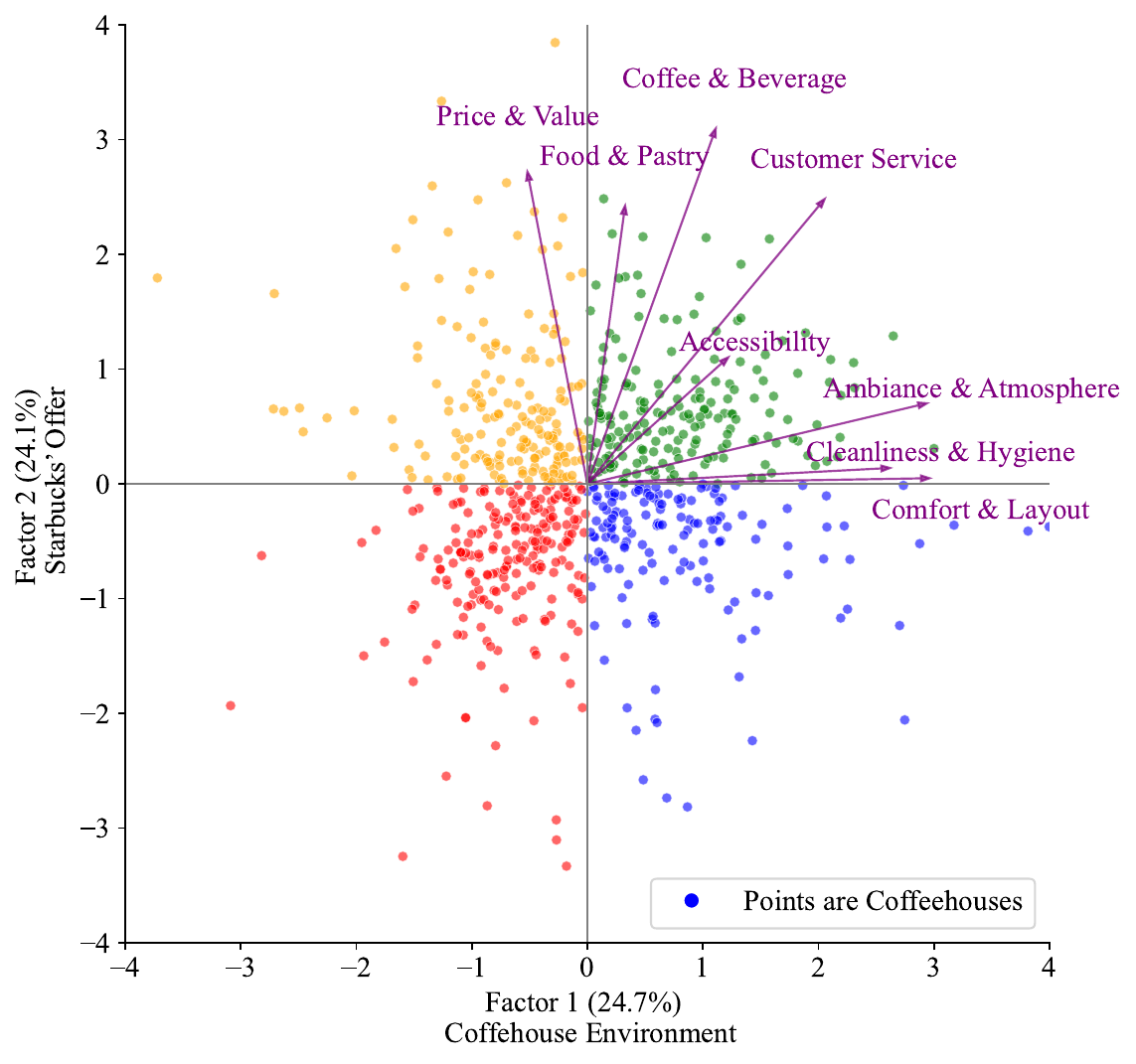}
        \caption{Perceptual Map of Coffeehouses}
        \label{fig:factor_analysis}
    \end{subfigure}
    \caption{Attribute Importance and Perceptual Map of Coffeehouses}
    \label{fig:importance_factor}
\end{figure}

As in conjoint analysis, Figure~\ref{fig:attribute_importance} depicts the relative importance of these attributes in predicting ratings. The results indicate that Customer Service is by far the most important driver of customer ratings (40\% importance) followed by Coffee \& Beverage (14\% importance). Together the store-level attributes (Ambiance \& Atmosphere, Comfort \& Layout, Cleanliness \& Hygiene) contribute 21\% importance. 
These findings highlight the importance of service, product quality, and the store environment in shaping customer satisfaction.

These results fit well with the perceptual map in Figure~\ref{fig:factor_analysis}, derived from a factor analysis of average sentiment across 722 stores and eight attributes (with two attributes omitted due to sparse observations). Together, they provide a succinct picture of how customers evaluate the Starbucks experience. The two factors capture 48.8\% of the variance in the data, split nearly evenly across them (all remaining factors have eigenvalues less than 1). The first factor (Coffeehouse Environment) loads on Ambiance \& Atmosphere, Cleanliness \& Hygiene, and Comfort \& Layout; the second factor (Starbucks’ Offer) loads on Coffee \& Beverage, Food \& Pastry, and Price, Value \& Promotions. By contrast, Customer Service loads almost equally on both factors, reflecting its cross-cutting role in shaping perceptions of both the environment and the core offer. This result highlights that service is not only the single most important driver of satisfaction but also a unifying dimension that spans all facets of the Starbucks experience. It is also consistent with Starbucks CEO Brian Niccol’s recently announced turnaround plan: “The goal is to provide customers in the United States — across more than 17,000 stores — with premium-priced, unique beverages in a welcoming coffeehouse environment, but at a fast-food pace” (New York Times, September 12, 2025).

Figure~\ref{fig:factor_analysis} provides a roadmap for strategic actions. As in quadrant analysis, coffee shops plotted as green points are performing well on both factors, reflecting favorable sentiment toward both the coffeehouse environment, the core offer, and service. By contrast, red points represent locations performing poorly on both dimensions, signaling the need for comprehensive improvement. Yellow points indicate stores doing relatively better on the coffeehouse environment dimension, while blue points show those performing relatively better on the offer dimension. This map thus allows managers to benchmark locations, identify strengths and weaknesses, and prioritize interventions tailored to local performance patterns.

\subsubsection*{Identifying High-Leverage Features}
While informative, measuring the impact of attribute-level sentiments on satisfaction is not directly actionable. Actionability requires analysis at the feature level, where specific drivers of satisfaction (e.g., wait time) can be identified and addressed. We now present the results of this feature-level analysis and demonstrate how it can generate actionable insights.
\begin{sidewaystable}[htbp]
\caption{Feature Regressions With and Without Control Variables}
\tiny
\begin{tabular}{lcccccc@{\hskip 0.5cm}cccccc}
\toprule
& \multicolumn{6}{c}{\textbf{Model 1 (Without Controls)}} & \multicolumn{6}{c}{\textbf{Model 2 (With Controls)}} \\
\cmidrule(lr){2-7} \cmidrule(lr){8-13}
& \multicolumn{3}{c}{Neutral} & \multicolumn{3}{c}{Positive}
& \multicolumn{3}{c}{Neutral} & \multicolumn{3}{c}{Positive} \\
\cmidrule(lr){2-4} \cmidrule(lr){5-7} \cmidrule(lr){8-10} \cmidrule(lr){11-13}
\textbf{Feature} & Coef. & SE & $p$-value & Coef. & SE & $p$-value & Coef. & SE & $p$-value & Coef. & SE & $p$-value \\
\midrule
\multicolumn{13}{l}{\textbf{Customer Service}} \\
~~~Complaints \& Conflict Resolution & .17 & .12 & .152 & .40 & .05 & $< .001$ & .17 & .14 & .244 & .41 & .05 & $< .001$ \\
~~~Customer Service Consistency & .26 & .10 & .009 & .28 & .05 & $< .001$ & .23 & .12 & .050 & .25 & .05 & $< .001$ \\
~~~Drive-Through Service Quality & .07 & .10 & .492 & .31 & .05 & $< .001$ & .07 & .09 & .470 & .33 & .05 & $< .001$ \\
~~~Management, Staff Friendliness, Expertise \& Professionalism & .70 & .09 & $< .001$ & 1.71 & .03 & $< .001$ & .65 & .09 & $< .001$ & 1.65 & .03 & $< .001$ \\
~~~Order Accuracy & .25 & .11 & .022 & .48 & .04 & $< .001$ & .24 & .11 & .029 & .48 & .04 & $< .001$ \\
~~~Service Efficiency \& Speed/Wait Time & .65 & .07 & $< .001$ & .84 & .03 & $< .001$ & .63 & .07 & $< .001$ & .79 & .03 & $< .001$ \\
\midrule
\multicolumn{13}{l}{\textbf{Coffee \& Beverage}} \\
~~~Coffee \& Beverage Customization \& Personalization & .18 & .10 & .084 & .23 & .05 & $< .001$ & .19 & .11 & .097 & .22 & .05 & $< .001$ \\
~~~Coffee \& Beverage Flavor Consistency & .10 & .13 & .475 & .04 & .06 & .486 & .09 & .16 & .584 & .05 & .07 & .487 \\
~~~Coffee \& Beverage Ingredient Quality & .22 & .13 & .093 & .28 & .07 & $< .001$ & .22 & .13 & .084 & .28 & .07 & $< .001$ \\
~~~Coffee \& Beverage Selection & .23 & .06 & $< .001$ & .40 & .06 & $< .001$ & .22 & .07 & $< .001$ & .41 & .06 & $< .001$ \\
~~~Coffee \& Beverage Taste & .20 & .09 & .020 & .63 & .04 & $< .001$ & .21 & .09 & .013 & .59 & .04 & $< .001$ \\
~~~Coffee Preparation \& Brewing Quality & .17 & .11 & .128 & .39 & .04 & $< .001$ & .11 & .11 & .314 & .39 & .05 & $< .001$ \\
\midrule
\multicolumn{13}{l}{\textbf{Facilities \& Accessibility}} \\
~~~Drive-Through Availability \& Quality & .19 & .07 & .007 & .28 & .05 & $< .001$ & .18 & .07 & .010 & .29 & .06 & $< .001$ \\
~~~Parking Accessibility & .22 & .11 & .038 & .18 & .06 & .003 & .24 & .10 & .025 & .16 & .07 & .021 \\
~~~Store \& Online Operating Hours & .67 & .12 & $< .001$ & .83 & .08 & $< .001$ & .63 & .12 & $< .001$ & .81 & .09 & $< .001$ \\
~~~Store Location Convenience & .16 & .07 & .023 & .34 & .05 & $< .001$ & .18 & .07 & .012 & .32 & .05 & $< .001$ \\
\midrule
\multicolumn{13}{l}{\textbf{Store Ambiance \& Atmosphere}} \\
~~~Interior Design \& Décor & .19 & .17 & .262 & .41 & .10 & $< .001$ & .26 & .20 & .187 & .38 & .12 & .001 \\
~~~Music, Lighting, Noise & .26 & .13 & .043 & .29 & .07 & $< .001$ & .31 & .12 & .010 & .29 & .09 & $< .001$ \\
~~~Sense of Community/Inclusivity & .51 & .17 & .002 & .66 & .08 & $< .001$ & .52 & .19 & .007 & .68 & .09 & $< .001$ \\
\midrule
\multicolumn{13}{l}{\textbf{Store Comfort \& Layout}} \\
~~~Indoor/Outdoor Seating & .34 & .12 & .005 & .28 & .09 & .002 & .45 & .13 & $< .001$ & .32 & .09 & $< .001$ \\
~~~Seating Availability \& Comfort & -.16 & .09 & .085 & .07 & .05 & .195 & -.12 & .09 & .174 & .06 & .05 & .222 \\
~~~Tables Arrangement & .00 & .14 & .978 & .17 & .08 & .042 & -.08 & .13 & .561 & .14 & .08 & .082 \\
~~~Workspace Quality & -.39 & .22 & .074 & .30 & .10 & .002 & -.43 & .22 & .053 & .31 & .10 & .003 \\
\midrule
\multicolumn{13}{l}{\textbf{Store Cleanliness \& Hygiene}} \\
~~~Store Cleanliness/Trash Disposal & .09 & .25 & .727 & .53 & .05 & $< .001$ & .05 & .21 & .826 & .52 & .06 & $< .001$ \\
\midrule
\multicolumn{13}{l}{\textbf{Food \& Pastry}} \\
~~~Food \& Pastry Selection & .07 & .08 & .368 & .25 & .07 & $< .001$ & .07 & .08 & .431 & .24 & .08 & .002 \\
~~~Food \& Pastry Taste & .44 & .16 & .006 & .49 & .08 & $< .001$ & .34 & .15 & .029 & .42 & .09 & $< .001$ \\
\midrule
\multicolumn{13}{l}{\textbf{Digital Services \& Technology}} \\
~~~Mobile \& Online Ordering & .36 & .11 & .002 & .36 & .07 & $< .001$ & .33 & .11 & .003 & .39 & .07 & $< .001$ \\
~~~Wifi Connectivity \& Power Outlets & .35 & .17 & .041 & .40 & .10 & $< .001$ & .36 & .18 & .047 & .39 & .11 & $< .001$ \\
\midrule
\multicolumn{13}{l}{\textbf{Price/Value \& Promotions}} \\
~~~Value for Money & .06 & .13 & .616 & .63 & .09 & $< .001$ & .09 & .12 & .440 & .62 & .10 & $< .001$ \\
\midrule
\multicolumn{13}{l}{\textbf{Fit Statistics}} \\
~~~Business FE & \multicolumn{6}{c}{No} & \multicolumn{6}{c}{Yes} \\
~~~Year FE & \multicolumn{6}{c}{No} & \multicolumn{6}{c}{Yes} \\
~~~Reviewer Controls & \multicolumn{6}{c}{No} & \multicolumn{6}{c}{Yes} \\
~~~Missing Features & \multicolumn{6}{c}{Yes} & \multicolumn{6}{c}{Yes} \\
~~~Nb. Obs. & \multicolumn{6}{c}{12,682} & \multicolumn{6}{c}{12,682} \\
~~~$R^2$ & \multicolumn{6}{c}{.68} & \multicolumn{6}{c}{.71} \\
~~~Adj. $R^2$ & \multicolumn{6}{c}{.68} & \multicolumn{6}{c}{.69} \\
\bottomrule
\multicolumn{13}{l}{\textit{Note.} Features with fewer than 3\% mentions were excluded from the analysis.}
\end{tabular}
\label{tab:actionable}
\end{sidewaystable}

Table~\ref{tab:actionable} reports the results. The two models yield comparable adjusted $R^2$ values of .68 and .69, respectively, suggesting that adding metadata controls provides little incremental explanatory power beyond feature-level sentiments. Importantly, the regression coefficients from both models are generally of similar magnitudes and statistical significance levels. Model 2 appears to be more conservative, likely due to larger number of parameters estimated (smaller degrees of freedom). For example, the positive sentiment of the feature ``Tables Arrangement,'' under the attribute Store Comfort \& Layout, becomes statistically insignificant at the 95\% confidence level in Model 2.

The results of Model 2 reveal several interesting patterns. All significant coefficients for neutral and positive sentiment are positive, indicating that improvements in sentiment relative to the negative baseline are associated with higher star ratings, offering face validity of our findings. Moreover, every attribute has at least one feature that is statistically significant, suggesting that all attributes are meaningfully associated with review ratings.

The strongest effect on positive sentiment is observed for Management, Staff Friendliness, Expertise \& Professionalism (1.65), underscoring the centrality of customer–employee interactions in the coffee shop experience. Other large effects include Store \& Online Operating Hours (.81), Service Efficiency \& Waiting Time (.79), Sense of Community/Inclusivity (.68), and Coffee \& Beverage Taste (0.59).

As is in the attribute-level analysis, these findings highlight the importance of service, product quality, and the store environment in shaping customer satisfaction. Service-related features—particularly staff professionalism and efficiency—have the largest effects on ratings, followed by aspects of the store experience and core beverage quality. 


\paragraph{Store-Level Impact}
Following the tradition in conjoint analysis, we use the parameter estimates from Model 2 to simulate the impact of improving feature sentiment by one level (e.g., from negative to neutral or from neutral to positive) on customer ratings. If a feature is already positive, its value is left unchanged. The impact is measured as the difference in the predicted rating before and after the sentiment change. Averaging these differences across all reviews for a given store yields the simulated effect of improving sentiment by one level for that store. Repeating this procedure across all features and stores generates the distribution of feature-improvement impacts on store ratings. This approach enables managers to quantify the expected gains in satisfaction from addressing specific features and to prioritize improvements with the highest potential impact.

\begin{figure}[htbp]
    \centering
    \includegraphics[width=0.75\linewidth]{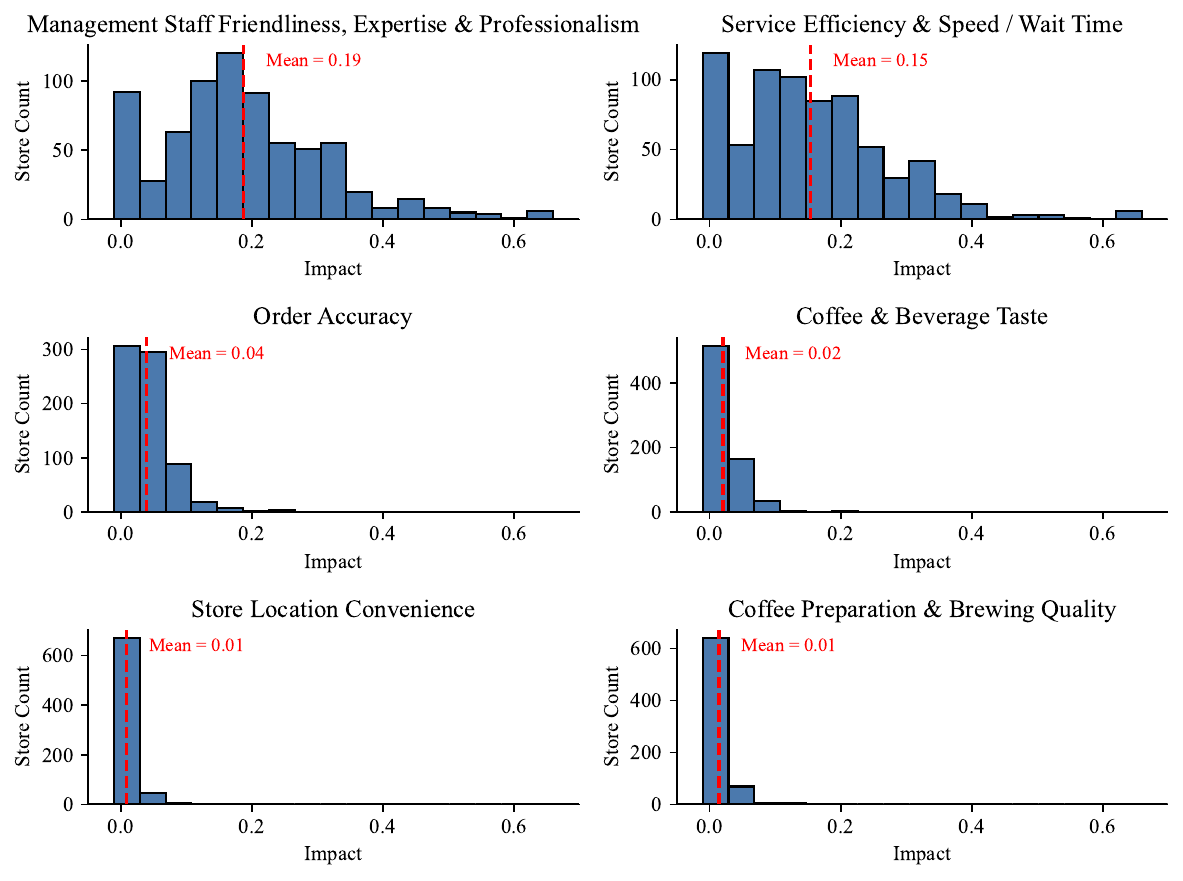}
    \caption{Distribution of Store-Level Effects from Improvements in Selected Features}
    \label{fig:impact_dist}
\end{figure}

Figure~\ref{fig:impact_dist} presents the distribution of the six most impactful features across the 722 Starbucks locations in our dataset. Management, Staff Friendliness, Expertise \& Professionalism and Service Efficiency \& Speed/Wait Time exhibit the highest average impacts and the greatest heterogeneity across stores, with mean effects of .19 and .16 rating points and standard deviations of .13 and .12, respectively. These are sizable effects. Prior research shows that a one-star increase in Yelp ratings can raise revenues by 5–9\% for independent restaurants, with the strongest effects observed among non-chain establishments \citep{luca2016reviews}. This result implies that improvements in these two features could translate, on average, into revenue gains of approximately .95–1.71\% and .80–1.44\% per store, respectively. The large range of impact of both features (0 to upwards of .6) suggests substantial inconsistency in how customers experience Starbucks at different locations and highlights an opportunity for the company to pursue its turnaround through targeted store-level actions. 

\begin{figure}[htbp]
    \centering
    \includegraphics[width=0.45\linewidth]{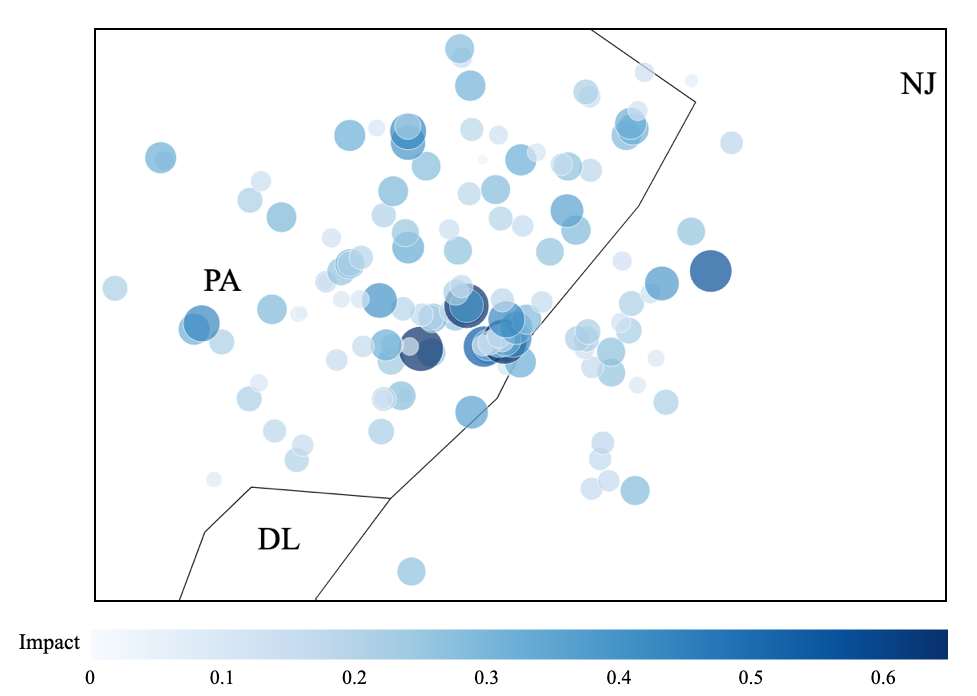}
    \caption{Impact of Improving Management, Staff Friendliness, Expertise \& Professionalism  on Individual Starbucks Locations}
    \label{fig:stroe_leve_impact}
\end{figure}

Figure~\ref{fig:stroe_leve_impact} illustrates such targeting for the Management, Staff Friendliness, Expertise \& Professionalism feature. Using actual location data from the Yelp dataset, the figure visualizes the predicted impact on store ratings for New Jersey and Pennsylvania locations from a one-level improvement in sentiment toward this feature. Lighter shades indicate stores with no incremental effect, while darker shades represent those expected to benefit most from the intervention. These results highlight high-leverage opportunities for enhancing customer satisfaction and store reputation, and they provide direct guidance for store-level targeted managerial actions.
The other four features in Figure~\ref{fig:impact_dist} are relatively less impactful on average, but their effects can be sizable for certain stores. For example, the impact of Order Accuracy ranges from 0 to .24. Thus, targeting stores where the impact exceeds, say, .10 may be worthwhile, as improvements in this feature could yield meaningful gains in customer satisfaction and store performance.

Our impact analysis demonstrates how improving feature sentiment by one level (e.g., enhancing perceptions of Service Efficiency, Speed \& Wait Time) can yield meaningful gains in customer ratings. While the analysis does not prescribe the exact means of improvement, it points to specific interventions Starbucks might consider, such as adding baristas during peak hours, optimizing mobile order workflows, or redesigning store layouts to ease congestion. Importantly, our results highlight where the problems lie at the store level and thus provide actionable guidance on which features to target. Rather than running experiments in a broad or uninformed way, our study helps Starbucks focus its resources on high-leverage areas. Although our estimates are correlational rather than causal, they offer a valuable starting point for prioritization, with follow-up A/B experiments needed to assess the causal effects of specific interventions and refine decision-making.

In sum, our structured dataset of attribute- and feature-level sentiments can be transformed into a marketing dashboard that equips managers with actionable diagnostics. By tracking sentiment dynamics over time, benchmarking performance across stores, and identifying high-leverage features for intervention, the dashboard turns unstructured reviews into a practical decision-support tool. Instead of relying on broad metrics or ad hoc experimentation, managers can use the dashboard to focus resources where they matter most, improving both customer satisfaction and store performance.

    \section*{Conclusion}

This study develops and validates a systematic, LLM-based approach for extracting attributes, features, and sentiments from customer reviews in a way that is both theoretically grounded and managerially actionable. The exploratory phase identified ten attributes and their associated features with strong face and discriminant validity, while the confirmatory phase produced a structured dataset capturing attribute- and feature-level mentions and sentiments.

A central strength of our approach is scalability. Human coders required a median of six minutes per review, while LLMs processed the same reviews in just two seconds with comparable reliability. This efficiency enables firms to analyze tens of thousands of reviews in real time, generating insights at a scale that is infeasible with manual coding. 

Our prompt-engineering experiments show that, relative to sentence-level prompting, review-level prompting yields significantly fewer attributes and features than human coders, reflecting systematic omissions. We also find significantly higher reliability with GPT-4.1 mini compared to GPT-4o mini, suggesting that LLM annotation accuracy is improving over time. By contrast, incorporating chain-of-thought reasoning provides only a modest improvement. Overall, the sentence-level with reasoning configuration of GPT-4.1 mini delivers the most accurate, context-aware outputs, achieving the highest agreement with human coders and sentiment distributions that are statistically indistinguishable from human benchmarks.

Managerially, our approach enables marketers to identify the key attributes and features highlighted in customer reviews, assess their associated sentiment, and quantify their impact on satisfaction. For Starbucks, customer service overwhelmingly shapes the experience yet remains highly polarized; service-related features, especially staff professionalism and efficiency, exert the strongest influence on ratings, followed by features related to the store environment (layout, comfort, accessibility) and beverage quality (coffee taste).


Our approach also guides marketers in leveraging structured review data to power an actionable dashboard that tracks sentiment across segments and over time, and identifies high-leverage attributes and features to enhance satisfaction. For Starbucks, dynamic analysis reveals a pivotal shift in 2016, when negative sentiment toward service overtook positive, coinciding with deteriorating employee relations documented in Harvard Business Review. At the store level, the dashboard highlights substantial heterogeneity in customer “pain” and “joy” points across locations, and simulations indicate that improving sentiment for high-impact features such as staff professionalism and service efficiency could yield 1–2\% average revenue gains per store. These diagnostics equip Starbucks to prioritize interventions both system-wide and locally, supporting its ongoing turnaround efforts.

Although we demonstrate our framework in the Starbucks context, it is readily applicable across industries where textual feedback is abundant, such as hospitality, healthcare, food service, and entertainment. By surfacing attribute-level “joy” and “pain” points,  benchmarking performance across segments and over time, and simulating the likely impact of interventions, the approach offers marketers a prescriptive, scalable decision-support tool.

We acknowledge limitations in this research. The initial refinement of attributes and features requires human oversight, and prompts were tailored to the coffee shop domain. Future research should automate this step, develop domain-agnostic prompts, and extend the framework to multimodal data. Moreover, while our results are predictive and robust, they remain correlational. Importantly, our analysis can guide marketers on which A/B experiments to run, by showing where problems lie and which features to prioritize. This helps firms focus resources on high-leverage areas, with follow-up experiments needed to establish causal effects and refine decision-making.

In sum, this study demonstrates how LLMs can scale human-like interpretation of reviews into actionable and prescriptive insights that guide targeted interventions. By combining prompt engineering innovations with a marketing dashboard that translates unstructured feedback into diagnostics, we show how firms can track dynamic sentiment shifts, identify systemic and local issues, and target interventions with precision, providing a foundation for effective turnaround strategies and long-term performance improvement.

    \singlespacing
    \bibliographystyle{jmr}
    \bibliography{references}

    \clearpage
\appendix
\titlecontents{section}[0em]{\bfseries}{}{}{\hfill\contentspage}
\titlecontents{subsection}[2em]{\itshape}{}{}{\hfill\contentspage}
\renewcommand{\thesection}{\Alph{section}}
\setcounter{section}{0}
\setcounter{page}{1}
\begin{center}
	{\Large \textbf{Web Appendix}}
\end{center}

These materials have been supplied by the authors to aid in the understanding of their paper. 
\vspace{1cm}
\tableofcontents
\clearpage

\setcounter{table}{1}
\renewcommand{\thetable}{\ref{sec:att_disc_prompts}\arabic{table}}
\setcounter{figure}{0}
\renewcommand{\thefigure}{\ref{sec:att_disc_prompts}\arabic{figure}}
\setcounter{figure}{0}
\renewcommand{\theprompt}{\ref{sec:att_disc_prompts}\arabic{prompt}}

\section{Web Appendix \ref{sec:att_disc_prompts}: Prompts for Attribute and Feature Extraction}
\label{sec:att_disc_prompts}
We provide details of the prompting procedures used the exploratory and confirmatory steps (1 and 2) of our proposed approach. Prompt~\ref{prompt:feature_discovery} outlines the instructions for Step 1, where the LLM extracts features from customer reviews. Similarly, Prompt~\ref{prompt:att_discovery} presents the instructions for identifying attributes in the review text as part of Step 1. For Step 2, Prompt~\ref{prompt:Review_sentiment} specifies the instructions for classifying the overall sentiment of each review. Prompt~\ref{prompt:sentence_attribute} presents the instructions for the assignment of sentences to attributes, while Prompt~\ref{prompt:att_sentiment} details the extraction of attribute-level sentiment at the sentence level. Finally, Prompts~\ref{prompt:feat_sentiment} and \ref{prompt:feat_sentiment_class} describe the procedures for identifying features and extracting their corresponding sentiment at the sentence level.

\singlespacing
\begin{prompt}[H]
\centering
\caption{Prompt for Feature Discovery}
\scriptsize
\begin{tcolorbox}[title=Feature Discovery, colback=gray!5, colframe=darkblue, fonttitle=\bfseries]
Reviews:\\
\{\}\\
\\
Instruction:\\
You are an AI assistant designed to help identify and extract key features mentioned in reviews.\\

We are conducting a marketing research study to extract attributes and features from customer reviews of coffee shops like Starbucks.\\

We define "features" as the specific, tangible characteristics of a product or service. They are the concrete, factual elements that describe\\
what something is or does. For a coffee shop, a features might be "Coffee brewing methods"\\

We define "attributes" as the benefits or emotional qualities that those features provide to the customer. They represent the value and\\
experience that customers perceive. Continuing with our coffee shop example, an attributes might be "Artisanal, personalized coffee experience" (from various brewing methods)\\

Thus, attributes represent the key aspects of a coffee shop whereas features represent the details associated with each attribute.\\

From the given reviews, identify the features that are being discussed by the reviewers. Please parsimoniously list the features, making sure\\
that you use unambiguous wording and avoiding duplicates. Order the list alphabetically. Report the results in a JSON object and do not explain yourself.\\
Return only the JSON in the following format with 'features' as the key and the values being a list of features.\\

Features =
\end{tcolorbox}
\vspace{0.2cm}
\label{prompt:feature_discovery} 
\end{prompt}

\clearpage

\begin{prompt}[H]
\centering
\caption{Prompt for Attribute Discovery}
\scriptsize
\label{prompt:att_discovery}
\begin{tcolorbox}[title=Attribute Discovery, colback=gray!5, colframe=darkblue, fonttitle=\bfseries]
Reviews:\\
\{\}\\
\\
Instruction:\\
You are an AI assistant designed to help identify and extract key attributes mentioned in reviews.\\

We are conducting a marketing research study to extract attributes and features from customer reviews of coffee shops like Starbucks.\\

We define "features" as the specific, tangible characteristics of a product or service. They are the concrete, factual elements that describe\\
what something is or does. For a coffee shop, a features might be "Coffee brewing methods"\\

We define "attributes" as the benefits or emotional qualities that those features provide to the customer. They represent the value and\\
experience that customers perceive. Continuing with our coffee shop example, an attributes might be "Artisanal, personalized coffee experience" (from various brewing methods)\\

Thus, attributes represent the key aspects of a coffee shop whereas features represent the details associated with each attribute.\\

From the given reviews, identify attributes that are being discussed by reviewers. Please parsimoniously list the attributes, making sure that you use unambiguous\\
wording and avoiding duplicates. Order the list alphabetically. Report the results in a JSON object. Return only the JSON in the following format with 'attributes' as the key and the values being a list of attributes.\\

Attributes =
\end{tcolorbox}
\end{prompt}

\clearpage

\begin{prompt}[H]
\centering
\caption{Prompt for Review Sentiment Classification}
\scriptsize
\label{prompt:Review_sentiment}
\begin{tcolorbox}[title=Review Sentiment Classification Prompt, colback=gray!5, colframe=darkblue, fonttitle=\bfseries]
---------------------\\
Task: Review Sentiment Classification \\
---------------------\\
\\
---------------------\\
Review\\
---------------------\\
\{\}\\
---------------------\\
Instructions: \\
---------------------\\
Please read the full customer review carefully, focusing on the key coffee shop attributes and features the customer highlights and their associated sentiments. The goal is to gain a clear understanding of the customer's feedback and sentiment about the coffee shop. 
\\
\\
*Question*: What is the overall sentiment the customer has towards the coffee shop? 
\\
\\
Please think step-by-step prior to providing your sentiment prediction. When classifying sentiment, please strictly use the below scale:
\\
\\
\hspace{1em}-- Strongly Negative \\
\hspace{1em}-- Negative \\
\hspace{1em}-- Neutral \\
\hspace{1em}-- Positive \\
\hspace{1em}-- Strongly Positive \\
\\
---------------------\\
Output Format: \\
---------------------\\
Please output your response in the following JSON format:
\\
\\
\texttt{ \{ } \\
\quad \texttt{"reasoning": "<Your step-by-step reasoning here>",} \\
\quad \texttt{"sentiment": "<Your sentiment prediction here>"} \\
\texttt{ \} }
\end{tcolorbox}
\end{prompt}

\begin{prompt}[H]
    \centering
    \caption{Sentence Attribute Assignment}
    \scriptsize
    \label{prompt:sentence_attribute}

\begin{tcolorbox}[title=Sentence Attribute Assignment Prompt, colback=gray!5, colframe=darkblue, fonttitle=\bfseries]
---------------------\\
Task: Sentence Attribute Assignment\\
---------------------\\
---------------------\\
Review Sentence\\
---------------------\\
\{\} \\
---------------------\\
Task Details:\\
---------------------\\
\textbf{Main Task:}\\
-- Above is a sentence or sentence(s) from the customer review. Please \textbf{match the sentence(s)} to the attribute(s) it mentions or describes.\\
-- If a sentence does not align with any of the pre-defined attributes, \textbf{assign it to “Other Attributes"}.\\

\textbf{Examples of Sentence Assignments:}\\
-- “I'm a Starbucks junkie.”\\
\hspace*{2em}- Assign this to \textbf{Other Attributes} because it expresses brand loyalty but does not refer to any specific coffee shop attribute.\\
-- “Our orders were very simple: one hot tea, one drip coffee, and two breakfast sandwiches.”\\
\hspace*{2em}- Assign this to \textbf{Coffee \& Beverage} and \textbf{Food \& Pastries} because it explicitly mentions drinks and food.\\

\textit{Make sure to select all applicable attributes for each sentence. This is the most critical part of the survey and heavily influences your bonus.}\\
---------------------\\
Instructions:\\
---------------------\\
1. Before assigning a sentence to a particular attribute, consider its relationship to the preceding and following sentence(s) for better context and accuracy. For example, the sentence “What can a girl do?” may seem ambiguous on its own. However, when it follows the sentence “Expensive, but good coffee,” it conveys a sense of dissatisfaction with the price. In this context, it should be assigned to the \textbf{Price/Value \& Promotions} attribute. \\
2. Ensure that \textbf{each review sentence is assigned to at least one attribute}. If a sentence fits multiple attributes, extract all relevant ones accordingly.\\
3. \textbf{"Other Attributes"} include aspects not tied to specific attributes, such as:\\
\hspace*{2em}- Local Starbucks Store Attitude (mentions a specific location)\\
\hspace*{2em}- Attitude Toward the Brand (references Starbucks as a whole)\\
\hspace*{2em}- Overall Customer Experience (general impressions of the visit)\\
\hspace*{2em}- Recommendation \& Future Behavior (intentions, recommendations, repeat visits)\\
\hspace*{2em}- If you encounter blank sentences (“...”) or unusual symbols that don’t convey meaningful content, assign them to “Other Attributes."\\
4. Follow this approach for each sentence above, ensuring it is mapped to the most appropriate attribute(s).\\
---------------------\\
Output Format:\\
---------------------\\
Please output your response in the following JSON format:\\
\vspace{-0.5cm}
\begin{verbatim}
{
    Sentence ID: {
        "sentence": "<Write sentence here>",
        "reasoning": "<Your step-by-step reasoning here>",
        "attributes": [<List of matching attributes here>],
    }
}
\end{verbatim}
To do so, please complete the following output word for word (do not adapt any of it!):\\
\begin{verbatim}
{
    "Sentence {fill-sentence-index-here}": {
        "sentence": "{fill-review-sentence-here}",
        "reasoning": "<Your step-by-step reasoning here>",
    }
}        
\end{verbatim}
\end{tcolorbox}
\end{prompt}

\begin{prompt}
    \centering
    \caption{Sentence Attribute Sentiment Classification}
    \scriptsize
    \label{prompt:att_sentiment}

\begin{tcolorbox}[title=Sentence Attribute Sentiment Classification Prompt, colback=gray!5, colframe=darkblue, fonttitle=\bfseries]
------------\\
Task: Sentence Attribute Sentiment Classification\\
------------\\

In the following questions, we will present the attribute \textbf{``\{fill-attribute-here\}''} you identified in the previous question and ask you to describe the customer's sentiment toward \textbf{``\{fill-attribute-here\}''}. 
To help refresh your memory, we will display the review sentences you assigned to \textbf{``\{fill-attribute-here\}''}.\\

------------------------------\\
Task Details and Instructions:\\
------------------------------\\

Below are the sentences you associated with the attribute \textbf{``\{fill-attribute-here\}''}:\\

\begin{verbatim}
{}
\end{verbatim}

How would you rate the customer's overall sentiment towards \textbf{``\{fill-attribute-here\}''} at this coffee shop? Prior to classifying the sentiment, please think step-by-step and consider: \\
\hspace*{2em}-- \textit{The specific sentences} you previously associated with \textbf{``\{fill-attribute-here\}''}\\
\hspace*{2em}-- \textit{All relevant information} related to \textbf{``\{fill-attribute-here\}''} within the overall context of the customer review. Feel free to read the customer review again.\\

When classifying sentiment, please strictly use the below scale:\\
\hspace*{2em}Strongly Negative\\
\hspace*{2em}Negative\\
\hspace*{2em}Neutral\\
\hspace*{2em}Positive\\
\hspace*{2em}Strongly Positive\\

------------\\
Output Format:\\
------------\\
Please output your response in the following JSON format:\\
\begin{verbatim}
{
"{fill-attribute-here}": {
    "reasoning_sentiment": "<Your step-by-step reasoning here about the sentiment towards {fill-attribute-here}>",
    "sentiment": "<Your sentiment prediction here for {fill-attribute-here}>"
}
}
\end{verbatim}

\end{tcolorbox}
\end{prompt}

\clearpage

\begin{prompt}
    \centering
    \caption{Sentence Feature Assignment Prompt}
    \scriptsize
    \label{prompt:feat_sentiment}
\begin{tcolorbox}[title=Sentence Feature Assignment Prompt, colback=gray!5, colframe=darkblue, fonttitle=\bfseries]
---------------------\\
Task: Sentence Feature Assignment\\
---------------------\\
---------------------\\
Review (repeated for reference)\\
---------------------\\
\{\}\\
---------------------\\
Review Sentence\\
---------------------\\
\{\}\\
---------------------\\
Task Details:\\
---------------------\\
\textbf{Main Task:}\\
-- Above is a sentence or sentence(s) from the customer review. Please \textbf{match the sentence(s)} to the feature(s) of \textbf{\{fill-attribute-here\}} that it mentions or describes.\\
-- If a sentence does not align with any of the pre-defined features, \textbf{assign it to "Other Features"}.\\
\textbf{Examples:}\\
If the attribute is \textit{Coffee \& Beverage}:\\
\hspace*{2em}-- “Our orders were very simple: one hot tea, one drip coffee, and two breakfast sandwiches.”\\
\hspace*{3em}- Assign this only to \textbf{Coffee \& Beverage Selection} as it explicitly mentions the coffee selection.\\
\hspace*{3em}- Even though \textbf{Food \& Pastry Selection} is valid, it is not a feature of \textbf{Coffee \& Beverage}.\\

If the attribute is \textit{Store Comfort \& Layout}:\\
\hspace*{2em}-- “I found a cozy corner with plenty of natural light, a nearby outlet, and strong Wi-Fi, which made it a great place to get some work done.”\\
\hspace*{3em}- Assign this to \textbf{Indoor/Outdoor Seating} and \textbf{Workspace Quality}.\\
\hspace*{3em}- Even though \textbf{Wifi Connectivity \& Power Outlets} is mentioned, it is not a feature of \textbf{Store Comfort \& Layout}.\\

\textit{Make sure to select all applicable features for each sentence. This is the most critical part of the survey and heavily influences your bonus.}\\
---------------------\\
Features\\
---------------------\\
Below are the features associated with the attribute \textbf{\{fill-attribute-here\}}. Constrain your selection only to these: \textbf{\{fill-all-attribute-features-here\}}\\
---------------------\\
Instructions:\\
---------------------\\
1. Before assigning a sentence to a particular feature, consider its relationship to nearby sentences for context.\\
2. Ensure that \textbf{each sentence is assigned to at least one feature}. Extract multiple features if relevant.\\
3. "Other Features" include any aspect not explicitly covered by listed features or blank/incoherent content.\\
4. Always map the sentence to the most appropriate feature(s).\\
------------\\
Output Format:\\
------------\\
Please output your response in the following JSON format:
\vspace{-0.4cm}
\begin{verbatim}
{
    Sentence ID: {
        "sentence": "<Write sentence here>""
        "reasoning": "<Your step-by-step reasoning here>",
        "features": [<List of matching features here>],
    }
 }
\end{verbatim}
To do so, please complete the following output word for word (do not adapt any of it!):
\vspace{-0.4cm}
\begin{verbatim}
{
    "Sentence {fill-sentence-index-here}": {
        "sentence": "{fill-review-sentence-here}",
        "reasoning": "<Your step-by-step reasoning here>",
    } 
}
\end{verbatim}
\end{tcolorbox}

\end{prompt}

\clearpage

\begin{prompt}[H]
    \centering
    \caption{Sentence Feature Sentiment Classification}
    \scriptsize
    \label{prompt:feat_sentiment_class}

\begin{tcolorbox}[title= Sentence Feature Sentiment Classification Prompt, colback=gray!5, colframe=darkblue, fonttitle=\bfseries]
------------\\
Task: Sentence Feature Sentiment Classification\\
------------\\

In the following questions, we will present the feature \textbf{``\{fill-feature-here\}''} you identified in the previous question and ask you to describe the customer's sentiment toward \textbf{``\{fill-feature-here\}''}. 
To help refresh your memory, we will display the review sentences you assigned to \textbf{``\{fill-feature-here\}''}.\\

------------------------------\\
Task Details and Instructions:\\
------------------------------\\

Below are the sentences you associated with the feature \textbf{``\{fill-feature-here\}''}:\\

\begin{verbatim}
\{\}
\end{verbatim}

How would you rate the customer's overall sentiment towards \textbf{``\{fill-feature-here\}''} at this coffee shop? Prior to classifying the sentiment, please think step-by-step and consider: \\
\hspace*{2em}-- \textit{The specific sentences} you previously associated with \textbf{``\{fill-feature-here\}''}\\
\hspace*{2em}-- \textit{All relevant information} related to \textbf{``\{fill-feature-here\}''} within the overall context of the customer review. Feel free to re-read the review.\\

When classifying sentiment, please strictly use the below scale:\\
\hspace*{2em}Strongly Negative\\
\hspace*{2em}Negative\\
\hspace*{2em}Neutral\\
\hspace*{2em}Positive\\
\hspace*{2em}Strongly Positive\\

------------\\
Output Format:\\
------------\\
Please output your response in the following JSON format:\\

\begin{verbatim}
{
    "{fill-feature-here}": {
        "reasoning_sentiment": "<Your step-by-step reasoning here about the sentiment towards {fill-feature-here}>",
        "sentiment": <Your sentiment prediction here for {fill-feature-here}>"
    }
}
\end{verbatim}
\end{tcolorbox}
\end{prompt}

\clearpage
\setcounter{table}{0}
\renewcommand{\thetable}{\ref{sec:human_survey}\arabic{table}}
\setcounter{figure}{0}
\renewcommand{\thefigure}{\ref{sec:human_survey}\arabic{figure}}

\section{Web Appendix \ref{sec:human_survey}: Human Validation}
\label{sec:human_survey}
We provide the details of the survey administered to human annotators to validate the results of our proposed approach for extracting attributes and features from reviews and identifying their corresponding sentiments.

\begin{table}[htbp]
\centering
\caption{Human Annotators Characteristics and Survey Feedback}
\footnotesize
\begin{tabular}{p{6cm} p{8cm}}
\toprule
\textbf{Measure} & \textbf{Value} \\
\midrule
Number of annotators & 10 (4 MBA, 4 MS, 2 PhD) \\
Average age & 28 years \\
Gender distribution & 7 females, 3 males \\
English fluency & 9 fluent, 1 advanced \\
Marketing courses taken (avg.) & 8 \\
Clarity of survey instructions & Yes = 10; No = 0 \\
Difficulty of survey task & 3.3 (1 = extremely easy, 5 = extremely difficult) \\
Challenges in attribute/feature coding & Yes = 4; No = 6 \\
Reasonable survey time & Yes = 8; No = 2 \\
Missing attributes/features & Yes = 2; No = 8 \\
\bottomrule
\end{tabular}
\label{tab:annotator_feedback}
\end{table}

Table \ref{tab:annotator_feedback} provides demographic details for our human coders. In total, we recruited ten coders, all holding advanced graduate degrees with a background in marketing. The coders were fluent in English. They unanimously found the task clear. The difficulty level was rated as average, and most coders agreed that the coding task was not particularly challenging and that the time allocated to complete the survey was reasonable. Most importantly, 80\% of the coders believed that the list of attributes was consistent and that no important attributes or features were omitted.

\begin{figure}[H]
    \centering
    \includegraphics[width=0.75\linewidth]{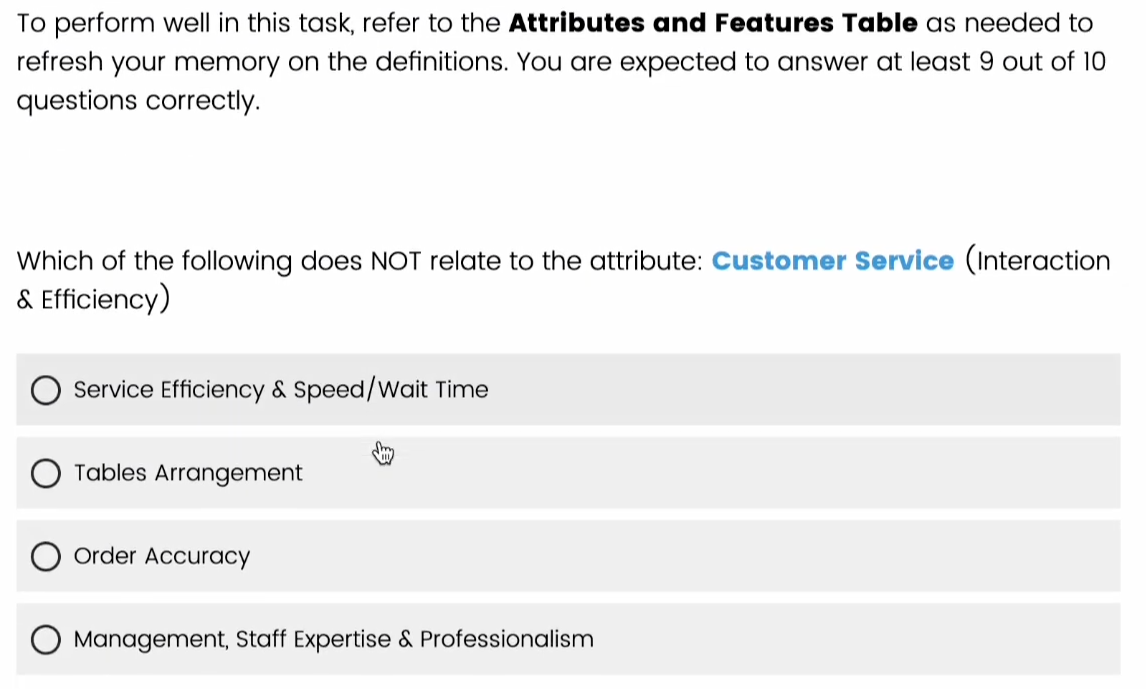}
    \caption{Example of Attention Quiz}
    \label{fig:quiz}
\end{figure}

Figure \ref{fig:quiz} shows an example attention check question. Human annotators were presented with ten such quizzes designed to test their familiarity with the attributes and their associated features. Each quiz was a multiple-choice question asking them to identify the feature that was not associated with the given attribute. Annotators were required to achieve a minimum score of 9 out of 10 before proceeding with the attribute and feature extraction tasks.

\begin{figure}[H]
    \centering
    \includegraphics[width=0.75\linewidth]{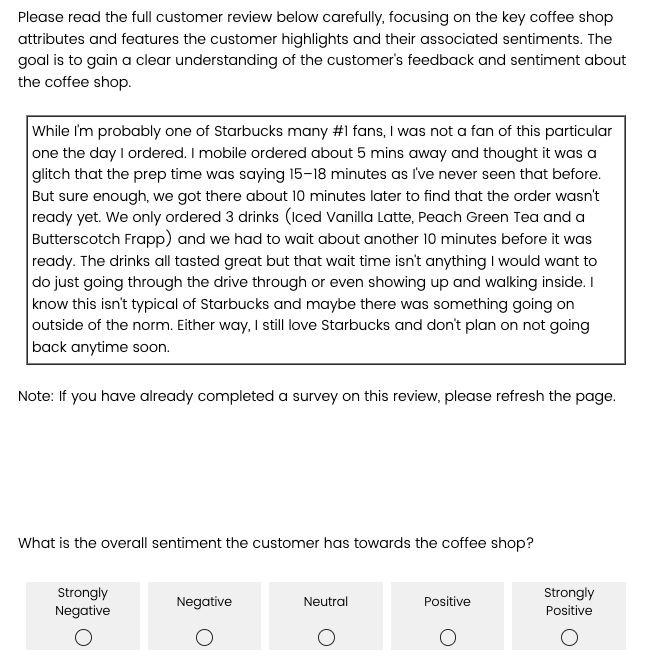}
    \caption{Review Sentiment Assessment }
    \label{fig:human_survey2}
\end{figure}

Figure~\ref{fig:human_survey2} presents an example of review-level sentiment assessment. In this task, human annotators were shown a random customer review and asked to evaluate the overall sentiment conveyed by the review on a 5-point scale, spanning strongly negative, negative, neutral, positive, and strongly positive.

\begin{sidewaysfigure}
    \centering
    \includegraphics[width=0.9\linewidth]{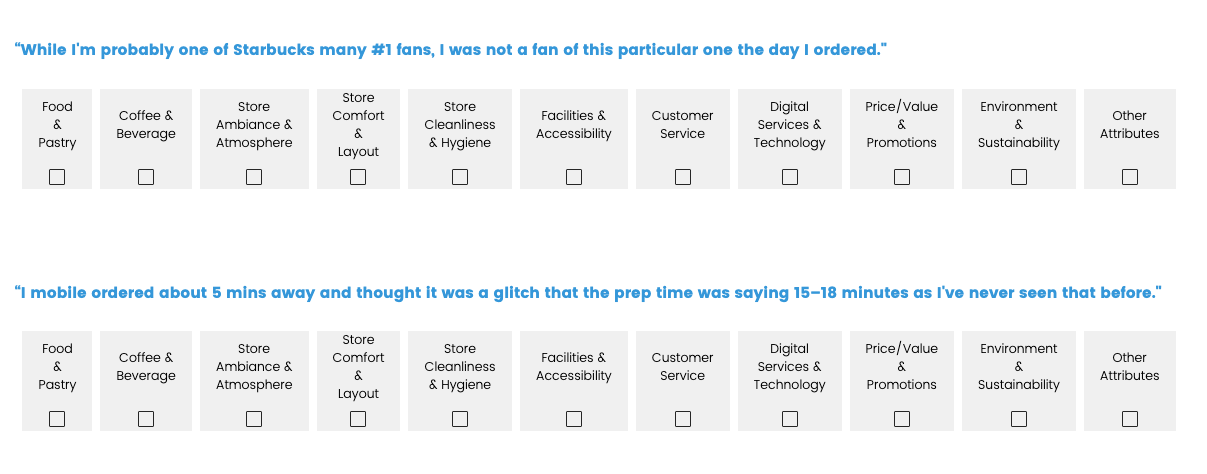}
    \caption{Sentence Allocation to Attributes}
    \label{fig:human_survey4}
\end{sidewaysfigure}

Figure~\ref{fig:human_survey4} illustrates an example of sentence-level allocation to attributes. Each review was segmented into individual sentences. Annotators were then asked to assign each sentence to one of the ten attributes identified in Step 1 of our approach. If a sentence did not pertain to any of the ten predefined attributes (See Table~\ref{tab:attribtues_features} of the main paper), annotators had the option to select the alternative ``Other Attributes.''

\begin{figure}[H]
    \centering
    \includegraphics[width=0.9\linewidth]{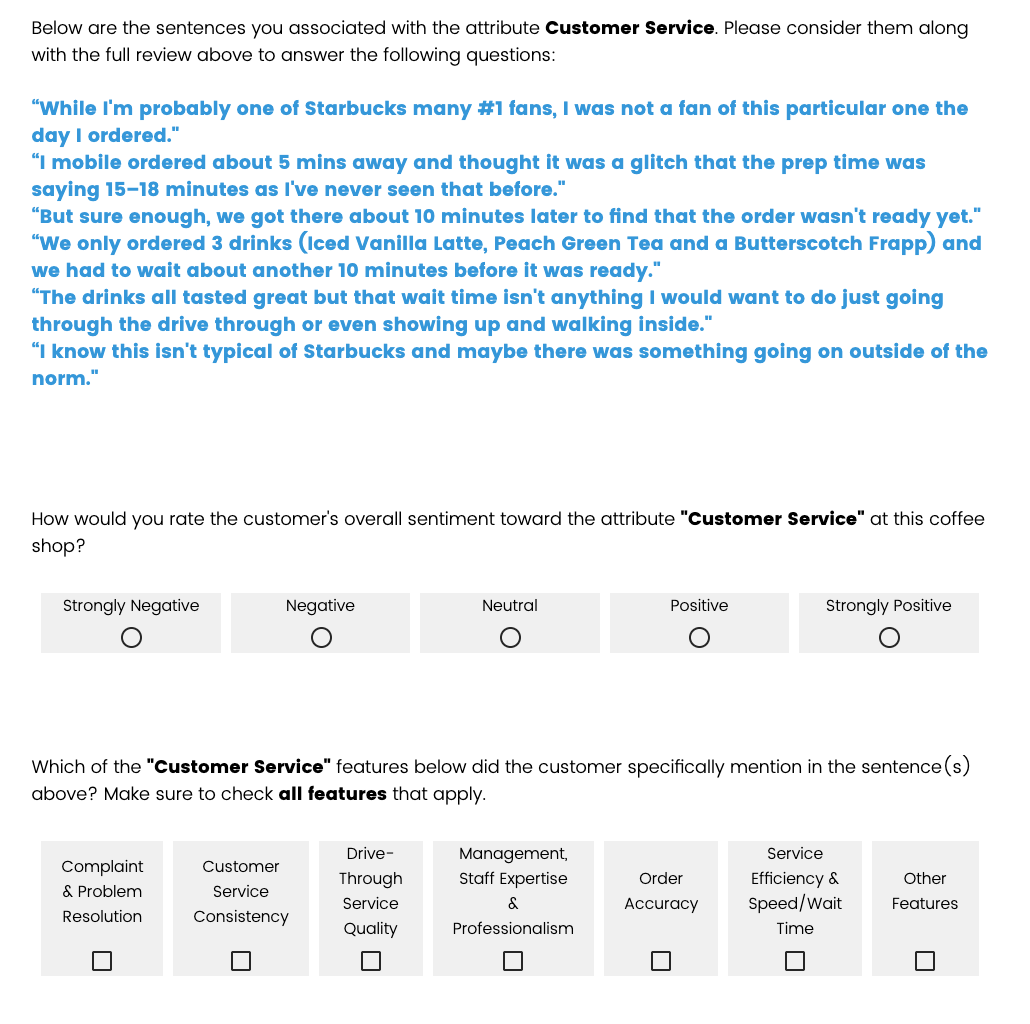}
    \caption{Attribute Sentiment Assessment and Feature Allocation}
    \label{fig:human_survey5}
\end{figure}

Figure~\ref{fig:human_survey5} shows an example of attribute-level sentiment assessment and feature identification. All sentences associated with a given attribute were grouped together. Annotators were first asked to evaluate the overall sentiment expressed toward the attribute, based on the content of the grouped sentences. Then, they were asked to identify the specific features discussed, selecting one or more from the predefined list of features associated with that attribute. They also had the option to select ``Other Features'' if the sentences mentioned features not included in our list of attribute features provided in Table~\ref{tab:attribtues_features} of the main paper.

\begin{figure}[H]
    \centering
    \includegraphics[width=0.9\linewidth]{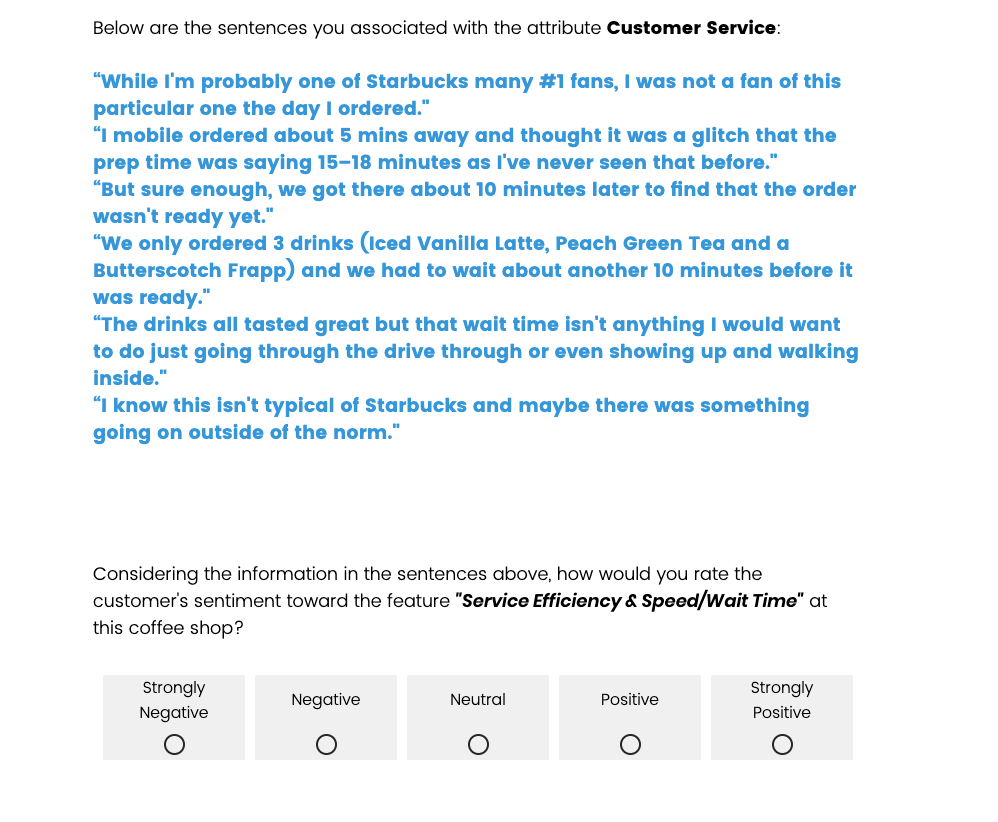}
    \caption{Feature Sentiment Assessment}
    \label{fig:human_survey6}
\end{figure}

Finally, Figure~\ref{fig:human_survey6} displays an example of feature-level sentiment assessment. All sentences related to the same feature were grouped, and annotators were asked to assess the sentiment toward that feature based on the grouped sentences.

\clearpage
\setcounter{table}{0}
\renewcommand{\thetable}{\ref{sec:human_regression}\arabic{table}}
\setcounter{figure}{0}
\renewcommand{\thefigure}{\ref{sec:human_regression}\arabic{figure}}

\section{Web Appendix \ref{sec:human_regression}: Predictive Performance of Attribute Sentiment}
\label{sec:human_regression}

We assess how well attribute-level sentiments predict actual consumer ratings. To this end, we estimate a regression model where the dependent variable is the rating provided by a consumer, and the independent variables are the extracted attributes and their associated sentiments from the review corresponding to the rating. Sentiment for each attribute is dummy-coded using four levels: {positive}, {neutral}, and {negative}, with an additional level, \textit{not mentioned}, to account for cases where an attribute is not present in the review.

We perform this analysis using both human-coded and LLM-coded attributes, and report the results of the two regression models in Table~\ref{tab:human_att_sent_regression}. In both models, the \textit{negative sentiment} level is used as the reference category for all attributes. First, we observe that both models achieve a high $R^2$ of .74, indicating that the attribute-level sentiments have strong explanatory power for predicting review ratings. Second, we find that all coefficients are positive, meaning that, relative to the negative sentiment baseline, either mentioning an attribute in a neutral or positive tone, or not mentioning it at all, is associated with higher ratings.

Finally, the coefficients obtained from the human-coded and LLM-coded models show strong alignment in both statistical significance and magnitude. This similarity is particularly notable for key attributes such as {Customer Service},  and Beverage Quality, and {Store Cleanliness and Hygiene}.

\begin{table}[htbp]
  \centering
  \caption{Regression Results of Attribute-Level Sentiments on Customer Ratings}
  \label{tab:human_att_sent_regression}
  \scriptsize
  \begin{tabular}{llcccccc}
    \toprule
    & & \multicolumn{3}{c}{\textbf{Human}} & \multicolumn{3}{c}{\textbf{GPT-4.1 mini}} \\
    \cmidrule(lr){3-5} \cmidrule(lr){6-8}
    \textbf{Attribute} & \textbf{Sentiment} & Coef. & SE & $p$-value & Coef. & SE & $p$-value \\
    \midrule
    Intercept & & -3.51 & .88 & $< .001$ & -3.12 & .87 & $< .001$ \\
    \midrule
    \multicolumn{8}{l}{\textbf{Customer Service}} \\
    & Neutral & 1.63 & .30 & $< .001$ & 1.06 & .69 & .126 \\
    & Positive & 2.22 & .14 & $< .001$ & 2.06 & .13 & $< .001$ \\
    & Not Mentioned & 1.49 & .19 & $< .001$ & 1.29 & .18 & $< .001$ \\
    \midrule
    \multicolumn{8}{l}{\textbf{Coffee \& Beverage}} \\
    & Neutral & .61 & .19 & .002 & .75 & .31 & .017 \\
    & Positive & .70 & .19 & $< .001$ & .89 & .19 & $< .001$ \\
    & Not Mentioned & .40 & .15 & .009 & .50 & .15 & .001 \\
    \midrule
    \multicolumn{8}{l}{\textbf{Facilities \& Accessibility}} \\
    & Neutral & .33 & .27 & .235 & .34 & .40 & .389 \\
    & Positive & .33 & .20 & .091 & .57 & .19 & .003 \\
    & Not Mentioned & .34 & .16 & .036 & .51 & .16 & .002 \\
    \midrule
    \multicolumn{8}{l}{\textbf{Store Ambiance \& Atmosphere}} \\
    & Neutral & .45 & .55 & .409 & -.08 & .50 & .866 \\
    & Positive & .37 & .25 & .136 & .39 & .27 & .148 \\
    & Not Mentioned & .27 & .21 & .201 & .21 & .24 & .381 \\
    \midrule
    \multicolumn{8}{l}{\textbf{Store Comfort \& Layout}} \\
    & Neutral & .10 & .43 & .811 & .49 & .41 & .229 \\
    & Positive & .46 & .28 & .103 & .63 & .27 & .019 \\
    & Not Mentioned & .13 & .24 & .596 & .32 & .22 & .141 \\
    \midrule
    \multicolumn{8}{l}{\textbf{Store Cleanliness \& Hygiene}} \\
    & Positive & 1.50 & .28 & $< .001$ & 1.31 & .28 & $< .001$ \\
    & Not Mentioned & 1.15 & .21 & $< .001$ & 1.09 & .19 & $< .001$ \\
    \midrule
    \multicolumn{8}{l}{\textbf{Food \& Pastry}} \\
    & Neutral & .59 & .44 & .179 & .78 & .96 & .416 \\
    & Positive & .44 & .39 & .261 & .72 & .37 & .053 \\
    & Not Mentioned & .32 & .32 & .327 & .50 & .30 & .096 \\
    \midrule
    \multicolumn{8}{l}{\textbf{Digital Services \& Technology}} \\
    & Neutral & .88 & .38 & .020 & .94 & .56 & .095 \\
    & Positive & .75 & .38 & .047 & .75 & .33 & .026 \\
    & Not Mentioned & .64 & .30 & .035 & .39 & .27 & .155 \\
    \midrule
    \multicolumn{8}{l}{\textbf{Price/Value \& Promotions}} \\
    & Neutral & .31 & .47 & .513 & .53 & .58 & .355 \\
    & Positive & .39 & .34 & .241 & .61 & .36 & .085 \\
    & Not Mentioned & .39 & .19 & .038 & .38 & .22 & .085 \\
    \midrule
    \multicolumn{8}{l}{\textbf{Environment \& Sustainability}} \\
    & Neutral$^*$ & --- & --- & --- & --- & --- & --- \\
    & Positive & 2.49 & .88 & .005 & 1.48 & 1.10 & .180 \\
    & Not Mentioned & 1.82 & .63 & .004 & 1.27 & .61 & .039 \\
    \midrule
    Missing Attributes & & \multicolumn{3}{c}{Yes} & \multicolumn{3}{c}{Yes} \\
    Nb. Obs. & & \multicolumn{3}{c}{300} & \multicolumn{3}{c}{300} \\
    $R^2$ & & \multicolumn{3}{c}{.74} & \multicolumn{3}{c}{.74} \\
    \bottomrule
    \multicolumn{8}{l}{\textit{Note.} $^*$Neutral level not included for Environment \& Sustainability, as it was not }\\
    \multicolumn{8}{l}{present in the data.}
  \end{tabular}
\end{table}

\clearpage
\setcounter{table}{0}
\renewcommand{\thetable}{\ref{sec:feature_validation}\arabic{table}}
\setcounter{figure}{0}
\renewcommand{\thefigure}{\ref{sec:feature_validation}\arabic{figure}}

\section{Web Appendix \ref{sec:feature_validation}:  Feature-Level Mention and Sentiment Distributions by Humans and LLM}
\label{sec:feature_validation}

Table~\ref{tab:features_human_llm} compares the distributions of feature-level mentions and sentiment between GPT-4.1 mini and human coders using sentence-level analysis with reasoning on the 3-point sentiment scale (negative, positive, and neutral). The two sets of distributions are highly congruent, mirroring our attribute-level results provided in Table~\ref{tab:comp_human_gpt_distr_ment} of the main paper. For example, human coders indicate that 70\% of reviews mention the feature Management, Friendliness, and Expertise (42\% positively, 26\% negatively, and the remainder neutrally), while the LLM produces a very similar figure with 72\% mentions (45\% positive, 26\% negative).

\begin{sidewaystable}
  \centering
  \caption{Feature-Level Mention and Sentiment Distributions by Humans and LLM}
  \tiny
    \begin{tabular}{p{2.5cm} p{4.25cm}lllllll}
        \toprule
    \multirow{2}[4]{*}{Attribute} & \multirow{2}[4]{*}{Feature} & \multicolumn{3}{c}{Human} &       & \multicolumn{3}{c}{GPT-4.1 mini} \\
\cmidrule{3-5}\cmidrule{7-9}          &       & Mention & Positive & Negative &       & Mention & Positive & Negative \\
    \midrule
        Customer Service & Management, Friendliness, Expertise & \barCell{70}{70}{blue} & \barCell{42}{42}{green} & \barCell{26}{26}{red} &       & \barCell{72}{72}{blue} & \barCell{45}{45}{green} & \barCell{26}{26}{red} \\
          & Service Efficiency \& Speed/Wait Time & \barCell{39}{39}{blue} & \barCell{16}{16}{green} & \barCell{22}{22}{red} &       & \barCell{45}{45}{blue} & \barCell{18}{18}{green} & \barCell{26}{26}{red} \\
          & Customer Service Consistency & \barCell{29}{29}{blue} & \barCell{20}{20}{green} & \barCell{8}{8}{red} &       & \barCell{11}{11}{blue} & \barCell{4}{4}{green} & \barCell{6}{6}{red} \\
          & Order Accuracy & \barCell{22}{22}{blue} & \barCell{6}{6}{green} & \barCell{15}{15}{red} &       & \barCell{25}{25}{blue} & \barCell{7}{7}{green} & \barCell{17}{17}{red} \\
          & Complaints \& Conflict Resolution & \barCell{9}{9}{blue} & \barCell{3}{3}{green} & \barCell{5}{5}{red} &       & \barCell{12}{12}{blue} & \barCell{4}{4}{green} & \barCell{8}{8}{red} \\
          & Drive-Through Service Quality & \barCell{9}{9}{blue} & \barCell{4}{4}{green} & \barCell{4}{4}{red} &       & \barCell{10}{10}{blue} & \barCell{4}{4}{green} & \barCell{6}{6}{red} \\
         \midrule
        Coffee \& Beverage &  Taste & \barCell{29}{29}{blue} & \barCell{19}{19}{green} & \barCell{8}{8}{red} &       & \barCell{29}{29}{blue} & \barCell{19}{19}{green} & \barCell{9}{9}{red} \\
          & Preparation \& Brewing Quality & \barCell{11}{11}{blue} & \barCell{5}{5}{green} & \barCell{6}{6}{red} &       & \barCell{18}{18}{blue} & \barCell{7}{7}{green} & \barCell{10}{10}{red} \\
          &  Selection & \barCell{5}{5}{blue} & \barCell{3}{3}{green} & \barCell{1}{1}{red} &       & \barCell{14}{14}{blue} & \barCell{6}{6}{green} & \barCell{3}{3}{red} \\
          &  Customization \& Personalization & \barCell{6}{6}{blue} & \barCell{2}{2}{green} & \barCell{3}{3}{red} &       & \barCell{10}{10}{blue} & \barCell{4}{4}{green} & \barCell{6}{6}{red} \\
          &  Flavor Consistency & \barCell{7}{7}{blue} & \barCell{5}{5}{green} & \barCell{2}{2}{red} &       & \barCell{9}{9}{blue} & \barCell{5}{5}{green} & \barCell{3}{3}{red} \\
          \midrule
        \multirow{2}{*}{\parbox[c]{\linewidth}{\raggedright Facilities \& Accessibility}} & Store Location Convenience & \barCell{16}{16}{blue} & \barCell{10}{10}{green} & \barCell{1}{1}{red} &       & \barCell{19}{19}{blue} & \barCell{13}{13}{green} & \barCell{4}{4}{red} \\
          & Drive-Through Availability \& Quality & \barCell{15}{15}{blue} & \barCell{4}{4}{green} & \barCell{8}{8}{red} &       & \barCell{12}{12}{blue} & \barCell{4}{4}{green} & \barCell{7}{7}{red} \\
          & Parking Accessibility & \barCell{8}{8}{blue} & \barCell{2}{2}{green} & \barCell{5}{5}{red} &       & \barCell{8}{8}{blue} & \barCell{2}{2}{green} & \barCell{6}{6}{red} \\
          \midrule
    \multirow{2}{*}{\parbox[c]{\linewidth}{\raggedright Store Ambiance \& Atmosphere}}  & Sense of Community/Inclusivity & \barCell{6}{6}{blue} & \barCell{3}{3}{green} & \barCell{3}{3}{red} &       & \barCell{9}{9}{blue} & \barCell{7}{7}{green} & \barCell{2}{2}{red} \\
          & Interior Design \& Décor & \barCell{5}{5}{blue} & \barCell{4}{4}{green} & \barCell{1}{1}{red} &       & \barCell{6}{6}{blue} & \barCell{5}{5}{green} & \barCell{0}{0}{red} \\
          & Music, Lighting, Noise & \barCell{5}{5}{blue} & \barCell{3}{3}{green} & \barCell{2}{2}{red} &       & \barCell{5}{5}{blue} & \barCell{4}{4}{green} & \barCell{1}{1}{red} \\
          \midrule
        \multirow{2}{*}{\parbox[c]{\linewidth}{\raggedright Store Comfort \& Layout}} & Seating Availability \& Comfort & \barCell{15}{15}{blue} & \barCell{8}{8}{green} & \barCell{6}{6}{red} &       & \barCell{15}{15}{blue} & \barCell{8}{8}{green} & \barCell{5}{5}{red} \\
          & Indoor/Outdoor Seating & \barCell{5}{5}{blue} & \barCell{3}{3}{green} & \barCell{1}{1}{red} &       & \barCell{7}{7}{blue} & \barCell{5}{5}{green} & \barCell{1}{1}{red} \\
          & Tables Arrangement & \barCell{5}{5}{blue} & \barCell{4}{4}{green} & \barCell{1}{1}{red} &       & \barCell{5}{5}{blue} & \barCell{3}{3}{green} & \barCell{1}{1}{red} \\
          & Workspace Quality & \barCell{6}{6}{blue} & \barCell{5}{5}{green} & \barCell{1}{1}{red} &       & \barCell{6}{6}{blue} & \barCell{5}{5}{green} & \barCell{1}{1}{red} \\
          \midrule
    \multirow{2}{*}{\parbox[c]{\linewidth}{\raggedright
Store Cleanliness \& Hygiene}}   & Store Cleanliness/Trash Disposal & \barCell{10}{10}{blue} & \barCell{6}{6}{green} & \barCell{5}{5}{red} &       & \barCell{11}{11}{blue} & \barCell{6}{6}{green} & \barCell{5}{5}{red} \\
          \\
          \midrule
    Food \& Pastry & Food \& Pastry Taste & \barCell{4}{4}{blue} & \barCell{3}{3}{green} & \barCell{0}{0}{red} &       & \barCell{4}{4}{blue} & \barCell{3}{3}{green} & \barCell{0}{0}{red} \\
          \midrule
    \multirow{2}{*}{\parbox[c]{\linewidth}{\raggedright
Digital Services \& Technology}}   & Mobile \& Online Ordering & \barCell{8}{8}{blue} & \barCell{3}{3}{green} & \barCell{1}{1}{red} &       & \barCell{8}{8}{blue} & \barCell{4}{4}{green} & \barCell{3}{3}{red} \\
          & Wifi Connectivity \& Power Outlets & \barCell{5}{5}{blue} & \barCell{3}{3}{green} & \barCell{1}{1}{red} &       & \barCell{4}{4}{blue} & \barCell{3}{3}{green} & \barCell{1}{1}{red} \\
          \midrule
\multirow{2}{*}{\parbox[c]{\linewidth}{\raggedright Price/Value \& Promotions}}  & Value for Money & \barCell{10}{10}{blue} & \barCell{1}{1}{green} & \barCell{8}{8}{red} &       & \barCell{8}{8}{blue} & \barCell{1}{1}{green} & \barCell{7}{7}{red} \\
   \\
    \bottomrule
    \multicolumn{6}{l}{Note. Features with less than 3\% mentions are not reported.}

    \end{tabular}%
  \label{tab:features_human_llm}%
\end{sidewaystable}%

\clearpage
\setcounter{table}{0}
\renewcommand{\thetable}{\ref{sec:comp}\arabic{table}}
\setcounter{figure}{0}
\renewcommand{\thefigure}{\ref{sec:comp}\arabic{figure}}

\section{Web Appendix \ref{sec:comp}: Comparison with Extant Methods}
\label{sec:comp}
We compare the predictive validity of our LLM approach against 25 state-of-the art methods. These 
 span three broad categories of NLP approaches: bag-of-words models, deep neural networks, and transformer-based models. Bag-of-words models emphasize interpretability by capturing word frequency and latent topics, and we implement four variants: TF-IDF \citep{ramos2003using}, which weighs words by their relative frequency across the corpus; LDA \citep{blei2003latent} and its nonparametric extension, the Hierarchical Dirichlet Process (HDP) \citep{boughanmi2021dynamics}, which model reviews as mixtures of topics; and Non-negative Matrix Factorization (NMF) \citep{lee2000algorithms}, which decomposes the term–document matrix into interpretable additive factors. Neural network models shift from interpretability to predictive performance by embedding reviews in a latent space: Word2Vec \citep{mikolov2013efficient} produces word embeddings that we aggregate at the review level, while Doc2Vec \citep{le2014distributed} directly learns fixed-length review embeddings. Finally, transformer-based models generate contextualized embeddings that capture nuanced semantics; we employ three Sentence-BERT (SBERT) variants—\textit{all-MiniLM-L6-v2}, \textit{paraphrase-MiniLM-L6-v2}, and \textit{paraphrase-MiniLM-L12-v2} \citep{reimers2019sentence}—to produce sentence embeddings used for rating prediction. Together, these methods provide a comprehensive set of benchmarks to evaluate the predictive validity of our proposed LLM approach.

We validate predictive performance using the full corpus of 12,682 reviews. The data are randomly split into training and test sets, with 90\% used for model training and performance assessed on the remaining 10\% holdout set. Table~\ref{tab:predictive} compares the predictive performance of the benchmark methods and our proposed approach. To ensure fairness, we estimate multiple variants of the benchmark models that vary in the number of latent features (topics or embeddings), allowing comparison across different model modalities.

Our proposed LLM-based model consistently outperforms all alternatives across multiple metrics, including Root Mean Squared Error (RMSE = .86), Mean Absolute Error (MAE = .70), and Mean Absolute Percentage Error (MAPE = 35.71), while maintaining strong correlation with ground truth (Pearson $r$ = .84, Spearman = .82). Importantly, it achieves these results with relatively few parameters (207) and offers full interpretability by design.

In contrast, traditional bag-of-words models such as TF-IDF and topic models (LDA, HDP, NMF) deliver substantially weaker performance, even as the number of topics or latent dimensions increases. Neural embedding models such as Word2Vec and Doc2Vec improve upon the bag-of-words baselines but still fall short in both accuracy and interpretability. Transformer-based SBERT models yield competitive correlation scores but similarly lack interpretability.

In sum, our LLM-based approach outperforms benchmark models in predictive accuracy while preserving interpretability, an essential condition for actionable insights.

\begin{sidewaystable}
  \centering
  \caption{ Predictive Performance of the Proposed Approach and Benchmark Models}
  \scriptsize
    \begin{tabular}{rlrcccccccc}
    \toprule
    \multicolumn{1}{l}{Model Type} & Model Name & \multicolumn{1}{c}{Variant} & Nb. Parameters & RMSE  & MAE   & MAPE  & R    & Pearson r & Spearman  & Interpretability \\
    \midrule
    \multicolumn{1}{l}{LLM} & Proposed &       & 207   & .86   & .70   & 35.71 & .70   & .84   & .82   & \multirow{13}[10]{*}{Yes} \\
\cmidrule{1-10}    \multicolumn{1}{l}{\multirow{12}[8]{*}{Bag-of-words}} & TF-IDF &       & 10,000 & .92   & .73   & 36.61 & .66   & .81   & .82   &  \\
\cmidrule{2-10}          & \multirow{5}[2]{*}{LDA} & 25 Topics & 25    & 1.16  & .95   & 49.78 & .45   & .68   & .67   &  \\
          &       & 50 Topics & 50    & 1.12  & .92   & 47.89 & .50   & .71   & .71   &  \\
          &       & 75 Topics & 75    & 1.09  & .89   & 45.69 & .52   & .73   & .73   &  \\
          &       & 100 Topics & 100   & 1.13  & .94   & 48.81 & .48   & .70   & .70   &  \\
          &       & 150 Topics & 150   & 1.12  & .91   & 47.42 & .49   & .70   & .71   &  \\
\cmidrule{2-10}          & HDP   & 150 Topics & 150   & 1.52  & 1.36  & 73.87 & .07   & .26   & .24   &  \\
\cmidrule{2-10}          & \multirow{5}[2]{*}{NMF} & 25 Dimensions & 26    & 1.33  & 1.14  & 61.20 & .28   & .53   & .58   &  \\
          &       & 50 Dimensions & 51    & 1.26  & 1.07  & 57.01 & .36   & .60   & .64   &  \\
          &       & 75 Dimensions & 76    & 1.23  & 1.04  & 55.44 & .39   & .62   & .67   &  \\
          &       & 100 Dimensions & 101   & 1.21  & 1.01  & 53.65 & .41   & .64   & .69   &  \\
          &       & 150 Dimensions & 151   & 1.20  & 1.02  & 54.15 & .42   & .65   & .70   &  \\
    \midrule
    \multicolumn{1}{r}{\multirow{10}[4]{*}{Deep Neural Networks}} & \multirow{5}[2]{*}{Word2Vec} & 25 Dimensions & 26    & 1.17  & .95   & 48.39 & .45   & .67   & .72   & \multirow{13}[6]{*}{No} \\
          &       & 50 Dimensions & 51    & 1.11  & .91   & 46.41 & .51   & .71   & .76   &  \\
          &       & 75 Dimensions & 76    & 1.08  & .89   & 44.89 & .53   & .73   & .77   &  \\
          &       & 100 Dimensions & 101   & 1.06  & .87   & 44.02 & .54   & .74   & .78   &  \\
          &       & 150 Dimensions & 151   & 1.05  & .86   & 43.05 & .55   & .74   & .78   &  \\
\cmidrule{2-10}          & \multirow{5}[2]{*}{Doc2Vec} & 25 Dimensions & 26    & 1.17  & .97   & 50.17 & .45   & .67   & .69   &  \\
          &       & 50 Dimensions & 51    & 1.16  & .96   & 49.48 & .45   & .67   & .70   &  \\
          &       & 75 Dimensions & 76    & 1.14  & .94   & 48.15 & .47   & .69   & .71   &  \\
          &       & 100 Dimensions & 101   & 1.15  & .94   & 47.95 & .47   & .69   & .71   &  \\
          &       & 150 Dimensions & 151   & 1.17  & .95   & 48.91 & .45   & .68   & .71   &  \\
\cmidrule{1-10}    \multicolumn{1}{r}{\multirow{3}[2]{*}{Transformers}} & \multirow{3}[2]{*}{SBERT} & all-MiniLM-L6-v2 & 384   & .93   & .73   & 36.04 & .65   & .81   & .80   &  \\
          &       & paraphrase-MiniLM-L6-v2 & 384   & .96   & .77   & 38.55 & .63   & .79   & .81   &  \\
          &       & paraphrase-MiniLM-L12-v2 & 384   & .91   & .72   & 35.45 & .67   & .82   & .80   &  \\
    \bottomrule
    \end{tabular}%
  \label{tab:predictive}%
\end{sidewaystable}%

    \clearpage
    \singlespacing

\end{document}